%% file: main.tex
\documentclass[journal]{IEEEtran}

\usepackage{amsmath,amsfonts}
\usepackage{array}

\usepackage{textcomp}
\usepackage{stfloats}
\usepackage{url}
\usepackage{verbatim}
\usepackage{graphicx}
\usepackage{cite}
\usepackage{bm}
\usepackage{booktabs}
\usepackage{multirow}

\ifCLASSOPTIONcompsoc
  \usepackage[caption=false,font=normalsize,
              labelfont=sf,textfont=sf]{subfig}
\else
  \usepackage[caption=false,font=footnotesize]{subfig}
\fi

\usepackage{xcolor}
\definecolor{ForestGreen}{RGB}{34,139,34}
\usepackage{placeins}

\usepackage[hidelinks]{hyperref}
\usepackage{cleveref}

\crefname{figure}{Fig.}{Figs.}
\crefname{equation}{Eq.}{Eqs.}
\crefname{table}{Table}{Tables}
\crefname{section}{Section}{Sections}
\crefname{chapter}{Chap.}{Chaps.}
\crefname{part}{Part}{Parts}

\begin{document}
\title{DUViN: Diffusion-Based Underwater Visual Navigation via Knowledge-Transferred Depth Features}
\author{Jinghe~Yang$^{1}$, Minh-Quan~Le$^{1}$, Mingming~Gong$^{2}$, and Ye~Pu$^{1}$

\thanks{$^{1}$Jinghe Yang (jinghey@student.unimelb.edu.au), Minh-Quan~Le, and Ye Pu are with the Department of Electrical and Electronic Engineering, The University of Melbourne, Parkville, VIC 3010, Australia.
$^{2}$Mingming Gong is with the School of Mathematics and Statistics, The University of Melbourne, Parkville, VIC 3010, Australia.}
\thanks{This article has been accepted for publication in \textit{IEEE Transactions on Robotics}. \textcopyright{} 2026 IEEE. Personal use of this material is permitted. Permission from IEEE must be obtained for all other uses, in any current or future media, including reprinting/republishing this material for advertising or promotional purposes, creating new collective works, for resale or redistribution to servers or lists, or reuse of any copyrighted component of this work in other works.}%
}
\maketitle

\begin{abstract}
Autonomous underwater navigation remains a challenging problem due to limited sensing capabilities and the difficulty of constructing accurate maps in underwater environments. In this paper, we propose a \textbf{D}iffusion-based \textbf{U}nderwater \textbf{Vi}sual \textbf{N}avigation policy via knowledge-transferred depth features, named \textbf{DUViN}, which enables vision-based end-to-end 4-DoF motion control for underwater vehicles in unknown environments. \textbf{DUViN} guides the vehicle to avoid obstacles and maintain a safe and perception-aware altitude relative to the terrain without relying on pre-built maps. To address the difficulty of collecting large-scale underwater navigation datasets, we propose a method that enables transfer under domain shifts from simulation with clear images to underwater environments by leveraging depth features and introducing a novel model transfer strategy. Specifically, our training framework consists of two phases: we first train the diffusion-based visual navigation policy on a clear image dataset collected in simulation using a pre-trained depth feature extractor. Secondly, we retrain the extractor on an underwater depth estimation task and integrate the adapted extractor into the trained navigation policy from the first phase. Experiments conducted in simulation and in an indoor water tank provide proof-of-concept validation of the effectiveness and sim-to-real transferability of our approach. The experimental videos are available in \url{https://www.youtube.com/playlist?list=PLqt2s-RyCf1gfXJgFzKjmwIqYhrP4I-7Y}.
\end{abstract}

\input{mainBody/1_Intro}
\input{mainBody/2_Related}
\input{mainBody/3_0_Methods}

\input{mainBody/3_1_Methods}

\input{mainBody/3_2_Methods}

\input{mainBody/4_1_Experiments}
\input{mainBody/4_2_Experiments}
\input{mainBody/5_Conclusion}

\bibliographystyle{IEEEtran}
\bibliography{ref}

\makeatletter
\renewcommand{\@IEEEBIOskipN}{1.5\baselineskip}
\makeatother

\end{document}

%% file: mainBody/1_Intro.tex
\section{Introduction}
\IEEEPARstart{T}{HE} rapid advancement of Autonomous Underwater Vehicles (AUVs) has propelled marine exploration into a new era. AUVs are increasingly deployed in tasks that reduce reliance on human divers in hazardous environments, including oil infrastructure inspection, marine biology research, and underwater equipment maintenance~\cite{sahoo2019advancements}. Among these tasks, navigating from a starting point to a designated destination remains a fundamental yet challenging problem, particularly in complex or unstructured underwater settings.

With the aid of underwater acoustic global positioning systems such as Ultra-Short Baseline (USBL), AUVs obtain global localization information during missions. However, safe and adaptive local motion planning, such as obstacle avoidance, relies on accurate onboard perception and real-time environmental awareness in unstructured underwater environments. Traditional approaches typically use acoustic sensors for environment mapping and obstacle detection \cite{paull2012sensor, zhang2017dynamic}. However, the acoustic sensors suffer from low resolution, slow scanning rates, and high operational costs~\cite{zhang2023autonomous}. Meanwhile, optical sensors like LiDAR and infrared, widely used in aerial and ground robotics, are ineffective underwater due to light absorption and scattering. By contrast, vision-based navigation has recently emerged as a promising solution and has been applied in several underwater navigation tasks \cite{manderson2018vision, karapetyan2021human}. 

Classical in-air vision-based planning methods typically involve accurate map construction using stereo vision or RGBD cameras, followed by trajectory optimization for obstacle avoidance~\cite{oleynikova2017voxblox, han2019fiesta}. These methods require dense geometric reconstruction, accurate model identification and rely on heavy online computation, making real-time deployment challenging~\cite{kulhanek2021visual}. To address these challenges, learning-based end-to-end visual navigation methods have emerged as a compelling alternative. These methods directly map raw visual input, typically RGB images, to action commands without requiring explicit map construction or localization. They are often trained using imitation learning~\cite{ross2011reduction,bojarski2016end} or reinforcement learning~\cite{haarnoja2018soft,lillicrap2015continuous}, and have demonstrated strong performance in complex and unconstrained environments. Some recent approaches also exhibit promising generalization to novel scenarios~\cite{shah2021rapid, shah2023vint, sridhar2024nomad}.

The underwater environment presents significant challenges for explicit map construction due to the limited availability of sensors~\cite{joshi2019experimental, qin2022survey}. As a result, the end-to-end learning-based method is a promising solution for AUVs. However, due to the significant domain gap between in-air and underwater environments, models trained in in-air scenarios suffer from performance degradation when directly applied to underwater settings~\cite{yang2024physics, ye2023underwater}. Moreover, collecting large-scale underwater datasets, particularly in unstructured environments, is expensive, further complicating the training of end-to-end policies. In particular, end-to-end navigation policies trained on RGB images are highly sensitive to appearance variations and often exhibit poor generalization across domains \cite{gervet2023navigating, chattopadhyay2021robustnav}. This challenge is pronounced in underwater environments, where variations in water types can lead to different visual degradations, which will affect the robustness of learned policies. One promising direction to address the generalization issue is to use abstract intermediate representations as the input for the decision model, instead of raw images, which has shown robustness under domain shift~\cite{gervet2023navigating}. To improve adaptation under domain shift in underwater settings, an end-to-end navigation policy was proposed in~\cite{lin2024uivnav}, where the authors leverage abstract image representations of Objects of Interest (OOIs), including oysters and rock reefs, instead of raw RGB images as input. The policy is then trained to navigate along these target objects while avoiding obstacles in the surrounding environment. However, this approach still faces notable limitations: the policy outputs only discrete yaw and pitch commands rather than continuous velocity commands, and it does not explicitly account for appearance variations under different levels of turbidity, which may limit its robustness in diverse underwater visual conditions.

Meanwhile, visual perception conditions should be considered in underwater vision-based navigation. The limited sensing range of cameras, caused by light absorption and scattering, can significantly degrade the performance of visual navigation methods. For example, the authors of~\cite{xanthidis2021aquavis} incorporate perception constraints into the planning process to ensure that visual landmarks remain within view along the planned trajectory. To ensure the visual perception in navigation, maintaining an appropriate and consistent altitude relative to the seabed is essential not only for collision avoidance, but also to preserve sufficient visual input for reliable navigation.  Unlike methods that rely on additional acoustic sensors for altitude regulation, our focus is on achieving this solely through vision-based perception. Compared with acoustic sensing, camera-based perception provides higher-resolution observations at lower cost, while also supporting monitoring-oriented underwater tasks.

\begin{figure*}[tp]
\centering
\includegraphics[width=0.99\linewidth]{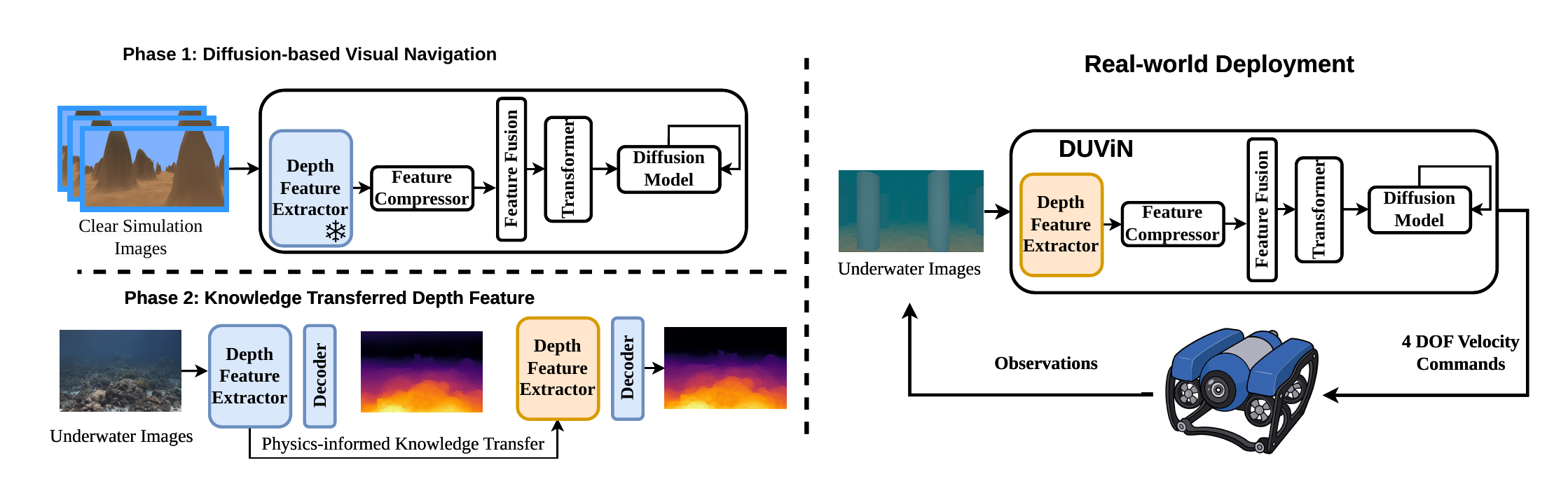}
\caption{\textbf{Overview of the DUViN framework:} \textbf{\textit{Phase 1}} – The navigation policy is trained using a clear image dataset collected in a simulator with underwater dynamics using monocular depth features extracted by a frozen, pre-trained Depth Anything encoder. These features are processed by the diffusion policy to generate smooth velocity commands. \quad 
\textbf{\textit{Phase 2}} – To enable cross-domain adaptation, the encoder is trained on a self-supervised, physics-informed underwater monocular depth estimation task using real underwater images, while the decoder remains frozen. \quad 
\textbf{\textit{Deployment}} – The adapted encoder is integrated into the trained navigation policy, replacing the original encoder. DUViN processes underwater observations to produce 4-DoF velocity commands for the AUV.}
\label{fig:phases}
\end{figure*}

To address the above challenges, particularly policy generalization and vision-based altitude control in underwater environments, we propose a \textbf{D}iffusion-based \textbf{U}nderwater \textbf{Vi}sual \textbf{N}avigation policy using knowledge-transferred monocular depth features, named \textbf{DUViN}.  Rather than relying on explicit map construction, such as VSLAM, which remains challenging in turbid underwater environments due to difficulties in finding stable visual correspondences, DUViN performs local navigation using only onboard image input, given the relative direction and distance to a known target destination. This enables the AUV to avoid obstacles, reach the goal, and maintain an appropriate altitude relative to the seabed terrain, thereby ensuring safety while preserving visual observability in challenging underwater environments.

The core idea of DUViN is to train the navigation policy on a clear image dataset collected in simulation using abstract depth features, and then transfer it to underwater environments by adapting the depth feature extraction module. The motivation for this design is that underwater navigation datasets are not only scarce, but also exhibit large appearance variations across different water conditions, turbidity levels, and lighting settings. As a result, a navigation policy trained or fine-tuned for a specific underwater condition may become sensitive to its visual characteristics, limiting its generalization to underwater environments with different visual conditions. To address this, we avoid using raw RGB observations and instead represent visual input using abstract depth features extracted from a pre-trained in-air monocular depth estimation model with strong generalization. Trained on large-scale and diverse in-air data, the model extracts depth features that are less sensitive to appearance changes than raw RGB inputs. Using these abstract depth features, rather than RGB inputs, further improves the policy's robustness to visual variations and supports transfer across different underwater visual conditions and previously unseen scene layouts.

However, due to the inherent domain gap between in-air and underwater imagery, directly applying in-air pre-trained visual models to underwater scenes leads to degraded performance. Underwater image restoration could be applied to mitigate underwater visual degradation~\cite{sea-thru,berman2020underwater}, but it would introduce an additional processing step and increase the computational burden for real-time deployment. To bridge this gap, we design DUViN as a two-phase cross-domain feature transfer framework rather than retraining the entire navigation policy with scarce underwater navigation data. Specifically, the framework has two phases: we first train a diffusion-based navigation policy on a clear image dataset in a simulator with underwater dynamics (see \cref{sec:Diffusion}), then adapt only the depth feature extraction module to underwater conditions through a downstream monocular depth estimation task (see \cref{sec:Monocular}). The adapted visual encoder is subsequently integrated into the navigation policy, enabling robust underwater visual navigation without requiring large-scale underwater navigation datasets. In this way, DUViN decouples policy learning from underwater visual adaptation, which is the key design that allows the proposed policy to transfer from clear image simulation training to underwater deployment.

The overall framework is illustrated in \cref{fig:phases}. Our main contributions are summarized as follows:
\begin{itemize}
    \item We propose \textbf{DUViN}, a diffusion-based underwater visual navigation policy that enables goal-directed three-dimensional motion with obstacle avoidance and terrain-following altitude maintenance without relying on pre-built maps, while improving transferability across different underwater visual conditions through transferred abstract depth features.
    \item We propose a two-phase knowledge-transfer framework to address the scarcity of underwater navigation training data across diverse underwater conditions. In the first phase, the navigation policy is trained in simulation using a clear image dataset, enabling it to learn navigation behavior from reliable depth features. In the second phase, the depth feature extractor is adapted to underwater imagery through a downstream monocular depth estimation task, thereby reducing the domain gap during policy transfer. This decoupling of policy learning from underwater visual adaptation is the central technical contribution of DUViN.
    \item We validate our approach in simulation, demonstrating its effectiveness and transferability, and further provide proof-of-concept validation on a physical AUV in an indoor water tank.
\end{itemize}

%% file: mainBody/2_Related.tex
\section{Related Works}
\subsection{Vision-Based End-to-End Navigation Policy}
Classical navigation approaches are often built upon accurate geometric map construction and state estimation \cite{malone2017hybrid,lu2024fapp}, which can be challenging in unstructured or sensor-limited environments. In contrast, recent end-to-end vision-based methods bypass the explicit mapping step and have demonstrated promising performance in different settings. Among them, reinforcement learning (RL) has emerged as a popular framework for learning navigation policies directly from visual inputs \cite{kahn2018self, devo2020towards}. Due to the collision risk when training directly in the real world, RL-based navigation is typically trained in a simulation environment, and applies domain generalization methods to transfer the policy to the real world \cite{tobin2017domain}. Imitation learning (IL) offers an alternative approach by learning navigation behaviors from expert demonstrations, which enables safer real-world deployment and improved sample efficiency \cite{codevilla2018end}. 

Meanwhile, recent methods such as ViNT~\cite{shah2023vint} and NoMaD~\cite{sridhar2024nomad} go beyond traditional IL by performing offline training on large-scale real-world robot operation data. These approaches learn generalized goal-conditioned policies without relying on expert supervision, and have demonstrated strong generalization and zero-shot performance across diverse environments and robot embodiments. Meanwhile, diffusion-based policies have recently emerged as a promising solution for robot trajectory planning and navigation, due to their ability to generate multi-modal action or trajectory distributions and high-dimensional sequential actions \cite{chi2023diffusion, carvalho2025motion, saha2024edmp}. In particular, NoMaD employs an observation-conditioned diffusion model to generate future action sequences, enabling expressive collision-free undirected exploration and goal-reaching navigation. Compared to ViNT, NoMaD achieves a higher success rate with fewer parameters,  making it more suitable for deployment on the onboard platforms. More recently, NaviDiffusor~\cite{zeng2025navidiffusor} further explored diffusion-based visual navigation by combining a diffusion model with designed cost guidance at the inference stage to achieve more robust navigation performance. Compared with these in-air methods, underwater local navigation faces much more severe appearance degradation and more limited training data.

Despite the scarcity of large-scale underwater datasets, several works have explored vision-based end-to-end navigation for underwater scenarios. In~\cite{manderson2018vision}, an end-to-end policy is trained using manually labelled actions to perform coral reef surveying while avoiding collisions. Building upon this,~\cite{manderson2020vision} introduces a hindsight relabelling strategy to enable goal-conditioned navigation. However, these methods are designed for specific coral reef inspection tasks and lack generalization to different underwater settings. To overcome this limitation,~\cite{lin2024uivnav} proposes an information-driven policy that leverages abstract intermediate representations (IR) of the Object of Interest (OOI), including estimated depth and semantic segmentation, as inputs to the behavior policy, thereby achieving domain-invariant navigation. However, it is trained using manually labelled data with discrete yaw and pitch commands only, which may limit its ability to adapt to complex 3D structures or terrain variations. Beyond these vision-based policies, a recent work, UnderwaterVLA~\cite{wang2025underwatervla}, introduces a dual-brain vision-language-action framework for autonomous underwater navigation, combining structured high-level reasoning with hydrodynamics-informed model predictive control for interpretable decision-making. Overall, compared with these prior methods, our work focuses on underwater visual navigation under limited data and across diverse underwater environments with different visual conditions.

\subsection{Monocular Depth Estimation}
Recent in-air monocular depth estimation methods trained on large-scale datasets have demonstrated strong generalization performance across diverse scenes~\cite{MiDaS, DPT, yang2024depthv1, yang2024depth}. Many of these models, such as MiDaS~\cite{MiDaS} and Depth Anything~\cite{yang2024depthv1}, adopt encoder-decoder architectures with either transformer-based or fully convolutional backbones. In these frameworks, visual features are first extracted by the encoder and subsequently combined and refined by the decoder to produce dense depth predictions. These models exhibit impressive zero-shot performance in real-world implementations. 

Compared to in-air settings, monocular depth estimation in underwater environments poses greater challenges due to the complex environment and the lack of large-scale labelled datasets. To mitigate these issues, several works have explored the use of underwater imaging formation models to assist training \cite{yu2023udepth} or adopted self-supervised learning or training on synthetic data~\cite{gupta2019unsupervised, hambarde2021uw, amitai2023self}. Although these approaches have shown promise, their performance remains limited. Another direction is to transfer pre-trained in-air models to the underwater domain. Despite the robustness in in-air settings, in-air models still suffer from noticeable performance degradation when directly applied to underwater images, primarily due to the domain gap between in-air and underwater environments, caused by the attenuation and scattering. In~\cite{yang2024physics}, the authors experimentally demonstrate that in-air models tend to overestimate depth in distant regions and produce unclear estimations in background areas when applied to underwater scenes. To address these issues, they propose a self-supervised, physics-informed framework that estimates underwater parameters from initial depth estimates and uses the underwater image formation model to correct estimation errors. This approach significantly reduces domain-induced errors and improves underwater depth estimation.

Meanwhile, monocular depth estimation has also demonstrated its feasibility for supporting downstream underwater navigation tasks \cite{lin2024uivnav}. Motivated by this, our work further leverages transferred depth features, rather than raw RGB observations, as the policy input to improve robustness across different underwater visual conditions.

%% file: mainBody/3_0_Methods.tex
\section{Method Overview}
In this work, we propose a vision-based, end-to-end navigation policy training method for underwater AUVs, based on a diffusion model with a transfer strategy. The proposed policy takes as input a sequence of image observations, along with an initial image that serves as an altitude reference, and is guided by the goal direction and distance. The policy outputs 4-DoF velocity commands for the following steps, corresponding to surge, sway, heave, and yaw in the robot frame. Due to the scarcity of underwater datasets, the entire navigation policy, encompassing all modules, is initially trained on a clear image dataset collected in simulation. To ensure the robustness to various underwater visual degradation conditions, we first utilize an in-air pre-trained depth estimation feature extractor, pre-trained on a large-scale in-air dataset, to provide visual features that condition the training of the diffusion-based decision model. Subsequently, to mitigate the performance degradation caused by domain shift when applying the navigation policy with the in-air feature extractor to underwater scenarios, we further adapt the feature extractor by integrating a downstream depth estimation model that incorporates underwater imaging physics. In the final real-world water tank deployment, the adapted extractor, fine-tuned for the underwater domain, replaces the original pre-trained extractor and guides the navigation of AUVs. This formulation is motivated by practical underwater robotic systems, where goal-related information can be provided by external acoustic positioning systems, such as USBL. DUViN focuses on local navigation using onboard image observations, while its velocity-based command interface is consistent with underwater platforms that commonly rely on onboard sensors such as DVL. In a full mission pipeline, given known destination information, DUViN performs local obstacle avoidance and altitude-aware motion generation, while the low-level controller tracks the generated velocity commands. As shown in \cref{fig:phases}, the training framework of DUViN has two phases.

In \textit{Phase 1}, we train a diffusion-based visual navigation policy using a clear image dataset without underwater visual effects collected in simulation, enabling the agent to avoid obstacles, maintain terrain-relative altitude, and reach the goal. To construct the training data, we collect sequences of image observations, goal direction and distance, and the corresponding velocity actions. These velocity sequences are generated in a simulator using a model-based method that accounts for the underwater dynamic constraints. To ensure altitude maintenance, the velocity supervision along the Z-axis is further adjusted to follow the terrain while preserving altitude relative to the initial observation height. To reduce sensitivity to appearance variations across underwater scenes, we utilize abstract visual features. Specifically, rather than using raw RGB images directly, intermediate features from a pre-trained depth estimation model are used. These extracted features are further compressed and fused with destination information, including the direction and distance to the destination, to match the input dimension of the diffusion-based navigation model.  Finally, the diffusion model is trained using the processed features and the generated velocity sequences.

\begin{figure*}[tp]
\centering
\includegraphics[width=0.9\linewidth]{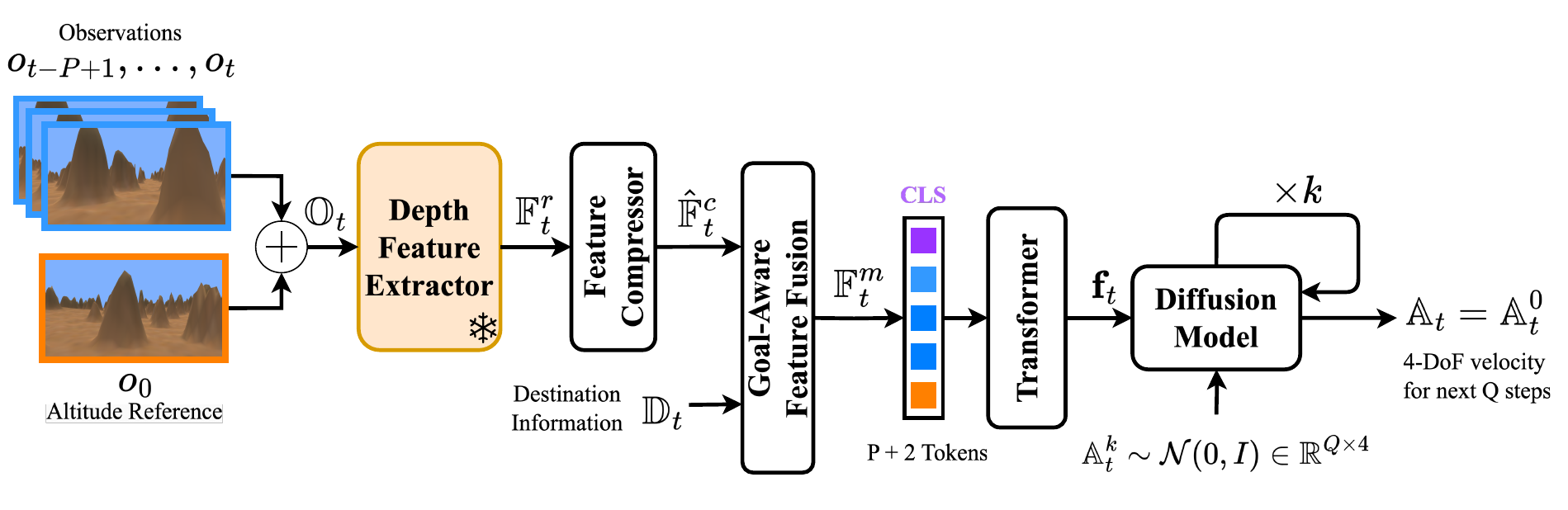}
\caption{\textbf{\textbf{DUViN} model structure:} DUViN takes a sequence of observations as input, with the initial frame serving as an altitude reference. Depth features are extracted, compressed, and fused with destination information. A transformer encodes the fused representation, which is then passed to a diffusion model to generate 4-DoF velocity commands.}
\label{fig:framework}
\end{figure*}

Benefiting from the strong generalization capability of the in-air pre-trained model trained on a large-scale dataset, and the use of abstract depth features, the navigation policy trained in \textit{Phase 1} demonstrates transferability across different environments. However, directly applying the navigation policy equipped with an in-air pre-trained feature extractor to underwater scenarios leads to noticeable performance degradation due to the domain gap between in-air and underwater imagery.

In \textit{Phase 2},  we adapt the depth feature extraction to the underwater setting to enable the navigation model to operate effectively underwater. Directly transferring a pre-trained depth estimation model results in performance degradation in underwater environments. Building on the transfer method of the underwater monocular depth estimation model described in prior work \cite{yang2024physics}, we further experimentally demonstrate that the primary cause of this performance drop is the degradation of the feature extractor. To address this, we freeze the decoder of the monocular depth estimation model and, guided by the underwater imaging formation model, retrain the feature extractor using real underwater images. The retrained feature extractor then replaces the original in-air extractor used in \textit{Phase 1} to extract observation features, which are subsequently processed and used as input to the trained diffusion-based navigation model.

Finally, during deployment, DUViN, equipped with the retrained feature extractor from \textit{Phase 2}, directly generates 4-DoF velocity commands from image observations. These commands are then tracked by the AUV’s low-level controllers to perform obstacle avoidance and terrain-following altitude maintenance.

%% file: mainBody/3_1_Methods.tex
\section{Phase 1: Diffusion-Based Visual Navigation via Monocular Depth Features}
\label{sec:Diffusion}
This section presents the overall method for training a diffusion-based visual navigation model that leverages depth features extracted from an in-air pre-trained depth estimation model. Given the known destination information, the navigation model generates 4-DoF velocity commands to navigate toward the destination while performing obstacle avoidance and altitude control following the seabed terrain change.

\subsection{Overall Model Structure}
The overall framework of our proposed model, DUViN, is illustrated in \cref{fig:framework}. DUViN consists of several components. A depth feature extractor first extracts abstract depth information from RGB observations. These features are then compressed to a lower-dimensional representation by a feature compressor. Subsequently, a goal-aware feature fusion block combines the observation features with the current goal information. A transformer module is then employed to summarize information from the sequence of observations and the initial reference image into a single feature representation. This feature representation serves as the condition for the diffusion-based navigation model, which denoises Gaussian noise into velocity commands. At each time step, the navigation policy takes as input a sequence of RGB observation images within a timeframe, together with the initial observation serving as an altitude reference to guide the underwater robot’s altitude maintenance during navigation. We denote the set of observation images at the current time step $t$ as:
\begin{equation}
\mathbb{O}_t = \left\{o_i \mid o_i \in \mathbb{R}^{3 \times H \times W}, \quad i \in \{0, t-P+1, ..., t\} \right\},
\end{equation}
where $H$ and $W$ denote the image height and width, respectively, and $P$ is the length of the observation sequence. The initial observation $o_0$ serves as the altitude reference. Unlike a scalar altitude target, the initial observation provides a purely visual reference for relative altitude maintenance without assuming explicit altitude measurement. Meanwhile, the sequence of observations $o_i$ for $i \in \{t-P+1, ..., t\}$, captured at time index $i$, provides the robot with information for obstacle avoidance and the change of terrain. When \( t \leq P \), we use the first observation \( o_1 \) for padding to ensure consistent input dimensions. The model also receives destination information, including the destination direction in the robot's body frame, $d_i \in [-\pi, \pi]$, and the distance progress ratio $s_i \in [0,1]$, defined as $s_i = 1 - \frac{\text{rest distance}}{\text{total distance}}$, as well as the initial values $d_0$ and $s_0$. Specifically, $s_0 = 0$ represents the starting position, while $s_i = 1$ indicates arrival at the destination. In our method, we represent the direction information in a two-dimensional form using its sine and cosine components and concatenate it with the distance information, forming a three-dimensional representation:
\begin{equation}
\mathbb{D}_t = \left\{ D_i = \begin{bmatrix} \sin d_i \\ \cos d_i \\ s_i \end{bmatrix} \in \mathbb{R}^{3} \,\middle|\, i \in \{0, t-P+1, ..., t\} \right\},
\end{equation}
where $\mathbb{D}_t$ represents the direction information for the observation set $\mathbb{O}_t$ at time $t$. Finally, the generated velocity actions for the next $Q$ steps, conditioned on the current observation $\mathbb{O}_t$ and goal $\mathbb{D}_t$, are denoted as:
\begin{equation}
 \mathbb{A}_t = \left\{\mathbf{a}_i \mid \mathbf{a}_i \in \mathbb{R}^{4},\ i \in \{t, \ldots, t+Q-1\} \right\}.
\end{equation}

\subsubsection{Depth Feature Extraction}
Without using the measured depth images as input to the navigation policy, our approach addresses the limitations of depth sensors in underwater environments by employing learning-based depth estimation. Meanwhile, to reduce computational overhead and simplify the navigation architecture, we do not explicitly estimate depth as intermediate outputs. Instead, we directly use the encoder’s intermediate depth features, which preserve essential geometric cues for obstacle avoidance and goal-directed motion while avoiding an additional decoding-and-re-encoding process, thereby reducing inference overhead and latency for real-time underwater navigation.

Depth serves as an abstract yet geometrically consistent representation across in-air and underwater domains, making it suitable for domain adaptation. Recent advances in in-air monocular depth estimation have produced models with strong generalization capabilities, trained on large-scale datasets. These models typically adopt an encoder-decoder structure, where the encoder extracts depth features and the decoder reconstructs refined depth estimates.

In our method, we adopt the latest work, Depth Anything V2~\cite{yang2024depth}, as the depth feature extractor. Depth Anything V2 is a monocular depth estimation model that employs DINOv2~\cite{oquab2023dinov2} as the encoder and DPT~\cite{DPT} as the depth decoder. Specifically, the RGB observations are first processed by the encoder to extract depth features and then passed through the transformer layers for feature reassembly. We directly extract the reassembled feature representations from the deepest transformer layer, resulting in a spatial feature map that captures global depth information with a compact spatial resolution. To optimize computational efficiency, we adopt the ViT-S encoder from Depth Anything V2 with a patch size of 14, producing feature representations with 384 channels. ViT-S is selected as a practical trade-off between feature quality and computational efficiency, so as to reduce inference latency for real-time navigation and maintain smooth command execution. To align with the input size requirements of the subsequent diffusion model, the input images are resized to $H = W = 448$ before feature extraction. This resizing ensures that the reassembled spatial feature map has a size of $16 \times 16$, resulting in 256 features. The raw extracted depth feature set from observations $\mathbb{O}_t$ at time $t$ is defined as:
\begin{equation}
\mathbb{F}_t^{r} = \left\{ \mathbf{f}^{r}_i \mid \mathbf{f}^{r}_i \in \mathbb{R}^{384 \times 16 \times 16}, \, i \in \{0, t-P+1, ..., t\} \right\},
\end{equation}
where $\mathbf{f}^{r}_i$ represents high-dimensional feature maps with 384 channels and a spatial resolution of $16 \times 16$.

\subsubsection{Feature Compression}
To further reduce the dimensionality of the features for compatibility with the navigation policy input, a compression module is introduced. It applies a single $1 \times 1$ convolution to project the multi-channel feature maps into a one-channel representation, which is subsequently flattened into a 1D vector. The set of compressed features at time $t$ is denoted as:
\begin{equation}
\mathbb{F}_t^{c} = \left\{ \mathbf{f}^{c}_i \mid \mathbf{f}^{c}_i \in \mathbb{R}^{256}, \quad i \in \{0, t-P+1, ..., t\} \right\}.
\end{equation}

Underwater image features extracted under turbid conditions may exhibit distribution shifts compared to those seen during training. To reduce feature-space shifts caused by changes in the observed scenes and water conditions, we draw inspiration from the domain adaptation technique Adaptive Instance Normalization (AdaIN)~\cite{huang2017arbitrary} and apply feature normalization. During training, we buffer the global mean and standard deviation of the compressed features across the altitude reference image dataset, denoted as $\mu_{\text{train}}^c$ and $\sigma_{\text{train}}^c$. At test time, we normalize each compressed feature $\mathbf{f}_i^c$ based on the statistics of the initial frame $\mathbf{f}_0^c$, and then rescale it using the global training statistics:
\begin{equation}
\hat{\mathbf{f}}^{c}_i = \frac{\mathbf{f}^{c}_i - \mu_0^c}{\sigma_0^c} \cdot \sigma_{\text{train}}^c + \mu_{\text{train}}^c,
\label{fn-1}
\end{equation}
where $\mu_0^c$ and $\sigma_0^c$ denote the mean and standard deviation of the initial compressed feature $\mathbf{f}^{c}_0$. After normalization, the resulting set of the normalized features is denoted as:
\begin{equation}
\hat{\mathbb{F}}_t^{c} = \left\{ \hat{\mathbf{f}}^{c}_i \mid \hat{\mathbf{f}}^{c}_i \in \mathbb{R}^{256}, \quad i \in \{0, t-P+1, ..., t\} \right\}.
\label{fn-2}
\end{equation}

This feature normalization helps maintain consistency in the extracted feature space across different scenes and water conditions.

\subsubsection{Goal-Aware Feature Fusion}
Since the goal of our navigation model is to reach the destination location, we assume that the destination information, including direction $d_i$ and distance $s_i$ in the robot body frame, can be acquired by the robot, using underwater global localization sensors such as USBL, together with each image observation. 

To enable goal-directed navigation, it is necessary to fuse the destination information with the extracted depth features. Specifically, we integrate the destination information into the depth feature representations extracted in the previous steps. This fusion provides the diffusion model with the representations that encode both the surrounding environment and the navigation goal.

Then, we use FiLM (Feature-wise Linear Modulation) \cite{perez2018film} to fuse the goal information with the compressed monocular depth feature. Specifically, the goal encoding $D_i \in \mathbb{D}_t$ is first passed through two separate single-layer fully connected networks to obtain the modulation parameters:
\begin{equation}
\gamma_i = f_{\gamma}(D_i), \quad  \beta_i = f_{\beta}(D_i),
\end{equation}
where $f_{\gamma}(\cdot)$ and $f_{\beta}(\cdot)$ are independent single-layer fully connected transformations. The outputs $\gamma_i, \beta_i \in \mathbb{R}^{256}$ match the dimensionality of the compressed depth feature $\hat{\mathbf{f}}^{c}_i \in \hat{\mathbb{F}}^{c}$. Next, FiLM is applied to modulate the compressed depth features:
\begin{equation}
\mathbf{f}^{m}_i = \gamma_i \odot \hat{\mathbf{f}}^{c}_i + \beta_i, \quad \text{for } \hat{\mathbf{f}}^{c}_i \in \hat{\mathbb{F}}^{c}.
\end{equation}

Here, $\odot$ denotes element-wise multiplication, where each dimension of $\hat{\mathbf{f}}^{c}_i$ is scaled and shifted element-wise by $\gamma_i$ and $\beta_i$. This modulation fuses the depth features with the destination information, dynamically guiding the model to adapt its representation to reach the target destination. The set of modulated features at time $t$ is denoted as:
\begin{equation}
\mathbb{F}_t^{m} = \left\{\mathbf{f}^{m}_i \mid \mathbf{f}^{m}_i \in \mathbb{R}^{256}, \ i \in \{0, t-P+1, ..., t\} \right\}.
\end{equation}

\subsubsection{Transformer and Diffusion Model}
Instead of generating waypoints followed by separate waypoint tracking, our diffusion model directly outputs smooth 4-DoF velocity commands for the next \( Q \) action steps, with a time interval of \( \Delta t \). These velocity commands can be directly tracked by the low-level controllers of the underwater robot. This design eliminates the need for an additional localization estimation and trajectory planning module, thereby simplifying the system architecture, reducing control latency and the need for map-based localization. Moreover, underwater robots are typically equipped with onboard sensors like DVL (Doppler Velocity Log) and IMU that can reliably measure velocity and acceleration, making velocity-based control more practical and robust compared to position-based trajectory tracking.

Before feeding into the diffusion model, the modulated features in $\mathbb{F}_t^{m}$, corresponding to several consecutive observations, are treated as tokens and combined into a unified feature representation. Transformers are effective at modeling relationships among multiple inputs and capturing long-range dependencies, making them well-suited for this aggregation task. Therefore, we adopt a transformer encoder, following a similar architecture to \cite{sridhar2024nomad}, but modified to include a CLS (classification) token for aggregating global contextual information. The input to the encoder consists of $P+2$ tokens, corresponding to the CLS token, the token of the initial observation, and the tokens of the sequence of subsequent observations. The encoder comprises four layers with four attention heads. The CLS token, appended to the input sequence, captures the overall scene context, and its representation extracted from the final transformer layer serves as the final input to the navigation diffusion model, denoted as $\mathbf{f}_{\text{t}}$.
\begin{equation}
\mathbf{f}_t = \text{Transformer}(\mathbb{F}_t^{m})_{\text{CLS}}, \quad \mathbf{f}_t \in \mathbb{R}^{256}.
\end{equation}

Given the observation set $\mathbb{O}_t$ at time step $t$, we extract the observation content feature $\mathbf{f}_t$, which encodes the scene representation for guiding the diffusion model. We use a 1D CNN-based diffusion model \cite{chi2023diffusion}, leveraging the Denoising Diffusion Probabilistic Model (DDPM) \cite{ho2020denoising} to approximate the distribution $p(\mathbb{A}_t \mid \mathbf{f}_t)$. The model generates velocity commands $\mathbb{A}_t$ by iterative denoising a sampled action sequence, conditioned on the content feature $\mathbf{f}_t$ and the diffusion step $k$, following the update rule:
\begin{equation}
\mathbb{A}_{t}^{k-1} = \alpha_k \cdot \left( \mathbb{A}_{t}^{k} - \gamma_k \, \epsilon_{\theta}(\mathbf{f}_t, \mathbb{A}_{t}^{k}, k) \right) + \sigma_k \mathcal{N}(0, I),
\end{equation}
where $\alpha_k$, $\gamma_k$, and $\sigma_k$ are functions of the noise schedule step $k$, and $\mathcal{N}(0, I)$ is a standard Gaussian noise term added in each iteration. The denoising process begins with an initial noisy action sample $\mathbb{A}_t^k \sim \mathcal{N}(0, I)$ and iteratively refine to the final command $\mathbb{A}_t = \mathbb{A}_t^0$. During training, we randomly sample a diffusion step $k$ and perturb the groundtruth velocity command $\mathbb{A}_t$ by adding Gaussian noise $\epsilon^k \sim \mathcal{N}(0, I)$. The noise predictor $\epsilon_{\theta}$ is trained to estimate the added noise by minimizing the following objective:
\begin{equation}
\begin{aligned}
\mathcal{L}
&=\mathbb{E}_{k,\,\boldsymbol{\epsilon}^k}\Biggl[
\sum_{\tau=t}^{t+Q-1}\kappa^{\tau-t}
\left\|\epsilon^k_{\tau}
-\epsilon_{\theta}\left(
\mathbf{f}_{\tau},\,\mathbb{A}_{\tau}+\epsilon^k_{\tau},\,k
\right)\right\|_2^2
\Biggr],
\end{aligned}
\end{equation}
where $\kappa \in (0, 1]$ is a discount factor that is to emphasize accurate prediction of near-term actions over distant ones.

\subsection{Training Data Generation}
\subsubsection{Data Collection Overview}
Collecting a large-scale underwater training dataset for AUV navigation is challenging due to the complexity of underwater environments and the limitations of available sensors. Although underwater simulation environments offer an alternative, they still suffer from domain gaps, and the diversity of real underwater visual conditions, such as varying water types and turbid conditions, further complicates generalization. In contrast, our method enables training on clear image datasets collected in simulation and transferring the trained navigation policy to underwater settings, improving the feasibility and reducing the practical burden of data collection.

Our training data is collected in a simulated Unity environment with clear images featuring randomly generated terrains with height variations and randomly placed sparse hill-like obstacles. The robot, equipped with an onboard camera, is placed in the environment and tasked with reaching a randomly selected target location while avoiding collisions. The collected dataset consists of sequences of RGB image observations and corresponding 4-DoF smooth velocity actions, generated by considering the dynamics model, through the following two steps:

\begin{itemize}
\item \textbf{Motion Planning for Goal-Reaching Obstacle Avoidance}: For a given environment and destination, a waypoint trajectory is first generated to guide the robot toward the destination while avoiding obstacles. During trajectory generation, we also enforce altitude maintenance relative to the terrain and ensure that the robot's facing direction provides sufficient visual information for navigation. After that, to generate smooth and dynamically feasible velocity actions suitable for AUV control in practice, we employ a model-based motion planning method that accounts for the dynamic constraints of the AUV (see \cref{sec:3-1}). 

 \item \textbf{Altitude Maintenance Z-Axis Motion Planning:}  Additionally, to enable the robot to maintain a consistent altitude relative to the terrain while referring to the altitude from the initial observation, we randomize the initial observation with different altitudes. The velocity commands along the $z$-axis are then adjusted to simultaneously perform terrain following and altitude maintenance relative to the initial altitude (see \cref{sec:3-2}).
\end{itemize}

The details of the dataset generation and collection in our implementation are introduced in the following sections.
\subsubsection{Motion Planning for Goal-Reaching Obstacle Avoidance}
\label{sec:3-1}
For a given environment, a set of waypoints that satisfy both safety and perception requirements is first generated using trajectory planning methods, such as A*, RRT. We define the set of planned waypoints \( \mathbb{W} \) in a 4-DoF as:
\begin{equation}
\mathbb{W} = \{ w_W^n = (x_W^n, y_W^n, z_W^n, \psi_W^n) \mid n = 1, ..., N \},
\end{equation}
where the \(n\)-th waypoint \(w_W^n\) consists of the position and the yaw angle in the world frame \(W\). \(N\) is the total number of waypoints. To generate consistent velocity commands across diverse terrains, we use Model Predictive Control (MPC), which produces smooth trajectories that respect dynamics constraints and exhibit uniform behavior across different randomly generated environments. The initial trajectory planning along the Z-axis follows the changes in terrain height, starting from a randomly assigned initial altitude. The smooth velocity command set \(\mathbb{V}\) in the body frame, used to track all waypoints in the set \(\mathbb{W}\), is defined as:
\begin{equation}
\mathbb{V} = \{ v_B^m = (v_{x,B}^m, v_{y,B}^m, v_{z,B}^m, v_{\psi,B}^m) \mid m = 1, \dots, T_t \},
\label{velocity_body}
\end{equation}
where \( v_B^m \) denotes the 4-DoF velocity command in the body frame \( B \) at time step \( m \), with a constant time interval \( \Delta t \). \( T_t \) is the total number of actions to track the whole trajectory. The full sequence $\mathbb{V}$ is constructed by repeatedly solving an MPC problem over a fixed prediction horizon \(T_p\). Each MPC iteration optimizes a local sequence of velocity commands: $\{ v_B^{\tau}, v_B^{\tau+1}, \dots, v_B^{\tau+T_p-1} \}$, starting from the current state, applies the first control \(v_B^{\tau}\), then shifts the horizon forward. This process is rolled out until the final waypoint is reached. The optimization problem at time step \( \tau \) is formulated as:
\begin{equation}
\begin{aligned}
\min_{\{v_B^m\}}
&\!\!\sum_{m=\tau}^{\tau+T_p-1}
\Big(
    \left\|p^{m+1}-w_W^n\right\|^2
    + \lambda_0
    \left\|v_B^m-v_B^{m-1}\right\|^2
\Big) \\
\text{s.t.} \quad
& p^{m+1}
  = p^m + \Delta t \cdot J(v_B^m), \\
& -V_{\max} \leq v_B^m \leq V_{\max}, \\
& \forall m \in \{\tau,\ldots,\tau+T_p-1\}, \\
& p^\tau \text{ and } v_B^{\tau-1} \text{ are given}.
\end{aligned}
\label{MPC1}
\end{equation}

The predicted position and yaw angle in the world frame are denoted as \( p^m = (x_W^m, y_W^m, z_W^m, \psi_W^m) \). The system dynamics are represented by the function \( J(\cdot) \) ~\cite[Eq.~(7.191)]{fossen2011handbook}, which maps the 4-DoF body-frame velocity command \( v_B^m \) to the corresponding displacement in the world frame. The tracked waypoint index \( n \) is updated outside the optimization loop and is incremented when the current position satisfies the proximity condition \( \| p^{\tau} - w_W^n \| < \epsilon \). Velocity constraints for each DoF are defined as \( V_{\max} = (v_{x,\max}, v_{y,\max}, v_{z,\max}, v_{\psi,\max}) \). The weighting factor \( \lambda_0 \) balances tracking accuracy and control smoothness.

\subsubsection{Altitude Maintenance Z-Axis Motion Planning}
\label{sec:3-2}
The initial Z-axis velocity commands in \( \mathbb{V} \) consider only terrain adaptation, without maintaining the initial altitude observed at the starting point \( o_0 \). To enable altitude control relative to the initial observation, we introduce an additional MPC controller that simultaneously refines the $Z$-axis velocity to adapt to terrain variations while maintaining an altitude relative to the initial observation.

Given an observation set \( \mathbb{O}_t \), we denote the sampled initial observation as \( o_0 \), collected at altitude \( z_0 \), and the current observation as \( o_t \), collected at altitude \( z_t \). Let \( \tau_t \) denote the corresponding time index in the velocity set \(\mathbb{V}\) associated with observation \( o_t \). Starting from index \( \tau_t \), we extract the next \( Q \) velocity commands in the Z-axis: $\{ v_{z,B}^{\tau_t}, v_{z,B}^{\tau_t+1}, \dots, v_{z,B}^{\tau_t+Q-1} \}$, which are then reindexed as a local velocity set $\mathbb{V}_{t,z}$:
\begin{equation}
\mathbb{V}_{t,z} = \{ v_{z,B}^{\tau_t+m} \mid m = 0, \dots, Q-1\},
\end{equation}
where \(Q\) corresponds to the number of future steps predicted for each observation set by the navigation policy. To compute the refined Z-axis velocity sequence \( \mathbb{U}_{t,z} = \{ u_{z,B}^{\tau_t+m} \mid m = 0, \dots, Q-1 \} \), we solve the following MPC problem:
\begin{align}
\min_{\mathbb{U}_{t,z}}\quad
& \sum_{m=0}^{Q-1}(z^m-z_0)^2
+\lambda_1\sum_{m=1}^{Q-1}
(u_{z,B}^{\tau_t+m}-u_{z,B}^{\tau_t+m-1})^2 \notag\\
& {}+\lambda_2\sum_{m=0}^{Q-1}
(u_{z,B}^{\tau_t+m}-v_{z,B}^{\tau_t+m})^2 \notag\\
\text{s.t.}\quad
& z^m=z^{m-1}+\Delta t\cdot u_{z,B}^{\tau_t+m},
\label{MPC2}\\
& z^0=z_t,\quad u_{z,B}^{\tau_t}=v_{z,B}^{\tau_t},
\notag\\
& -v_{z,\max}\leq u_{z,B}^{\tau_t+m}\leq v_{z,\max},
\,\forall m\in\{0,\dots,Q-1\}. \notag
\end{align}

The final groundtruth velocity command set used for supervision, aligned with the observation sequence \( \mathbb{O}_t \) at time \( t \), is defined as:
\begin{equation}
\begin{aligned}
\mathbb{V}_t=\Bigg\{\Big(
& v_{x,B}^{\tau_t+m},\;v_{y,B}^{\tau_t+m},\\
& u_{z,B}^{\tau_t+m},\;v_{\psi,B}^{\tau_t+m}
\Big)\;\Big|\;m\in\{0,\dots,Q-1\}\Bigg\},
\end{aligned}
\end{equation}

where \( \tau_t \) denotes the corresponding index in the global velocity sequence at time \( t \). During training, \( \mathbb{V}_t \) serves as the groundtruth supervision for the diffusion model, which generates the actions \( \mathbb{A}_t \) conditioned on the observation \( \mathbb{O}_t \) and the goal information \( \mathbb{D}_t \).

In summary, to construct the velocity groundtruth for training, we first compute waypoints \( \mathbb{W} \) to reach the destination. Using Eq.~\eqref{MPC1}, we then generate the initial 4-DoF velocity command sequence \( \mathbb{V} \), ensuring consistency across varied terrain. At each time step \( t \), we sample the next \( Q \) velocity steps from \( \mathbb{V} \), starting at index \( \tau_t \), and collect the corresponding observation set \( \mathbb{O}_t \), which includes the RGB image sequence and a randomly sampled initial observation \( o_0 \) with altitude \( z_0 \). Finally, we apply Eq.~\eqref{MPC2} to refine the Z-axis velocity, enabling altitude maintenance relative to the initial reference. In the training loop, the extracted features from the image observations are fused with the goal information to condition the input of the diffusion model. The denoising steps are then performed following the standard diffusion process. The diffusion model is trained to iteratively denoise and recover the target velocity sequence under the supervision of the expert trajectory.

%% file: mainBody/3_2_Methods.tex
\begin{table*}[!bhp]
\centering
\caption{Depth Estimation Under Different Training Configurations. \textbf{\textcolor{ForestGreen}{Bold green}} indicates best, \textcolor{ForestGreen}{green} indicates second best}
\label{tab:freeze_ablation}
\renewcommand{\arraystretch}{1.1}  
\begin{tabular}{c|c|ccccc|ccccc}
\toprule
\multicolumn{2}{c|}{\textbf{Dataset}} & \multicolumn{5}{c|}{\textbf{Sea-thru}} & \multicolumn{5}{c}{\textbf{SQUID}} \\
\hline
\textbf{Model} & \textbf{Config} & $\boldsymbol{\delta\!<\!1.15^3} \uparrow$  & \textbf{AbsRel} $\downarrow$ & \textbf{SqRel} $\downarrow$ & \textbf{RMSE} $\downarrow$ & \textbf{SILog} $\downarrow$  & $\boldsymbol{\delta\!<\!1.15^3} \uparrow$  & \textbf{AbsRel} $\downarrow$ & \textbf{SqRel} $\downarrow$ & \textbf{RMSE} $\downarrow$ & \textbf{SILog} $\downarrow$\\
\hline
\multirow{4}{*}{DPT-H} 
  & Ori.     & 0.957 & 0.124 & 0.051 & 1.479 & 0.045 & 0.950 & 0.133 & 0.066 & 1.626 & 0.039\\    
  & E. F.    & 0.957 & 0.124 & 0.051 & 1.479 & 0.045 & 0.950 & 0.132 & 0.066 & 1.620 & 0.039\\
  & D. F.    & \textcolor{ForestGreen}{0.964} & \textcolor{ForestGreen}{0.108} & \textbf{\textcolor{ForestGreen}{0.038}} & \textbf{\textcolor{ForestGreen}{1.379}} & \textbf{\textcolor{ForestGreen}{0.039}} & \textcolor{ForestGreen}{0.954} & \textcolor{ForestGreen}{0.122} & \textcolor{ForestGreen}{0.049} & \textcolor{ForestGreen}{1.571} & \textcolor{ForestGreen}{0.036}\\  
  & w/o F.   & \textbf{\textcolor{ForestGreen}{0.966}} & \textbf{\textcolor{ForestGreen}{0.105}} & \textcolor{ForestGreen}{0.039} & \textcolor{ForestGreen}{1.391} & \textbf{\textcolor{ForestGreen}{0.039}} & \textbf{\textcolor{ForestGreen}{0.955}} & \textbf{\textcolor{ForestGreen}{0.115}} & \textbf{\textcolor{ForestGreen}{0.042}} & \textbf{\textcolor{ForestGreen}{1.550}} & \textbf{\textcolor{ForestGreen}{0.034}}\\
\hline
\multirow{4}{*}{DA V2} 
  & Ori.     & 0.967 & 0.095 & 0.030 & 1.403 & 0.036 & 0.964 & 0.092 & 0.024 & 1.534 & 0.027\\  
  & E. F.    & 0.967 & 0.095 & 0.030 & 1.405 & 0.036 & 0.964 & 0.092 & 0.024 & 1.538 & 0.027\\
  & D. F.    & \textbf{\textcolor{ForestGreen}{0.975}} & \textbf{\textcolor{ForestGreen}{0.087}} & \textbf{\textcolor{ForestGreen}{0.024}} & \textbf{\textcolor{ForestGreen}{1.364}} & \textbf{\textcolor{ForestGreen}{0.033}} & \textcolor{ForestGreen}{0.973} & \textbf{\textcolor{ForestGreen}{0.091}} & \textbf{\textcolor{ForestGreen}{0.020}} & \textcolor{ForestGreen}{1.409} & \textcolor{ForestGreen}{0.022}\\
  & w/o F.   & \textcolor{ForestGreen}{0.974} & \textcolor{ForestGreen}{0.092} & \textcolor{ForestGreen}{0.026} & \textcolor{ForestGreen}{1.383} & \textcolor{ForestGreen}{0.034} & \textbf{\textcolor{ForestGreen}{0.974}} & \textbf{\textcolor{ForestGreen}{0.091}} & \textbf{\textcolor{ForestGreen}{0.020}} & \textbf{\textcolor{ForestGreen}{1.397}} & \textbf{\textcolor{ForestGreen}{0.021}}\\
\bottomrule
\end{tabular}
\end{table*}

\section{Phase 2: Knowledge-Transferred Monocular Depth Features for Underwater Navigation}
\label{sec:Monocular}
Although abstract depth features are employed to train the diffusion-based navigation model on the clear image simulation dataset (see in \cref{sec:Diffusion}), a notable domain gap persists when the model is deployed in underwater environments. This performance degradation is primarily attributed to the domain shift encountered by monocular depth estimation models trained on clear images.

As observed in~\cite{yang2024physics}, in-air monocular depth estimation models tend to produce distorted depth estimations when applied to underwater images, due to light attenuation and backscattering. To address this issue, a self-supervised Physics-informed Underwater Depth Estimation (\textbf{PUDE}) is proposed to refine monocular depth estimations in underwater settings without requiring groundtruth depth, based on the following underwater image formation model~\cite{chiang2011underwater}:
\begin{equation}
\begin{aligned}
I_c(x)
&=J_c(x)t_c(x)
+B_c^\infty\left(1-t_c(x)\right),\\
t_c(x)
&=e^{-\beta_c z(x)},
\end{aligned}
\label{eq:underwater_image_formation}
\end{equation}
where \(I_c(x)\) and \( J_c(x) \) denote the pixel intensities of the observed underwater and the clear scene at pixel location \( x \) for color channel \( c \in \{R, G, B\} \). The medium transmission \(t_c(x)\) depends on the distance \(z(x)\) from the camera to the scene point and the attenuation coefficient \(\beta_c\), while \(B_c^\infty\) denotes the background ambient light caused by backscattering.

The initial depth estimates produced by the in-air model for underwater images, together with the input underwater images, are used to estimate the water parameters associated with Eq.~\eqref{eq:underwater_image_formation}. Based on the physical relationship among underwater image formation, water parameters, and scene depth, PUDE derives lower and upper bounds on \(t_c(x)\). Specifically, \(\hat{t}_c(x)\) denotes the reference transmission map derived from the initial depth estimation produced by the original in-air model, while \(\hat{B}^{\infty}_c\) denotes the estimated background ambient light. In contrast, \(\hat{t}^{P}_c(x)\) denotes the transmission map derived from the adapted depth estimation and is optimized through the bound losses during self-supervised training. Following PUDE, depth estimations that violate these physical bounds are penalized through the lower bound loss \(\mathcal{L}_{\mathrm{bl}}\) and the upper bound loss \(\mathcal{L}_{\mathrm{bu}}\):
\begin{align}
\mathcal{L}_{\mathrm{bl}}
&=\frac{1}{|\Omega|}
\sum_{\substack{c\in\{G,B\}\\x\in\Omega}}
\left[
-I_c(x)
+\left(1-\hat{t}^{P}_c(x)\right)\hat{B}^{\infty}_{c}
\right]_{+},
\label{eq:pude_bound_losses}\\
\mathcal{L}_{\mathrm{bu}}
&=\frac{1}{|\Omega_b|}
\sum_{\substack{c\in\{G,B\}\\x\in\Omega_b}}
\left[
-\left(1-\hat{t}^{P}_c(x)\right)\hat{B}^{\infty}_{c}
-\hat{t}_c(x)+I_c(x)
\right]_{+}, \notag
\end{align}
where \([\cdot]_{+}\) denotes the ReLU operator, \(\Omega\) is the image pixel set, and the background pixel set \(\Omega_b\subseteq\Omega\) is defined as $\Omega_b =
\left\{
x\in\Omega
\;\middle|\;
\frac{\hat{B}^{\infty}_c\left(1-\hat{t}_c(x)\right)}
{I_c(x)}
\geq \gamma,\;
\forall c\in\{G,B\}
\right\}
$, where \(\gamma\) is a threshold for selecting pixels that are likely to belong to background regions. The overall PUDE objective is:
\begin{equation}
\mathcal{L}_{\mathrm{PUDE}}
=
\omega_{\mathrm{sim}}\mathcal{L}_{\mathrm{sim}}
+
\omega_{\mathrm{bl}}\mathcal{L}_{\mathrm{bl}}
+
\omega_{\mathrm{bu}}\mathcal{L}_{\mathrm{bu}},
\label{eq:pude_objective}
\end{equation}
where \(\omega_{\mathrm{sim}}\), \(\omega_{\mathrm{bl}}\), and \(\omega_{\mathrm{bu}}\) are the corresponding loss weights. The similarity loss $\mathcal{L}_{\mathrm{sim}}$ is based on the median-normalized mean absolute difference between the initial in-air and adapted depth estimations, thereby preserving structural information from the initial in-air model. The lower bound loss $\mathcal{L}_{\mathrm{bl}}$ corrects depth overestimation in distant regions, while the upper bound loss \(\mathcal{L}_{\mathrm{bu}}\) refines depth underestimation in background regions. These losses provide physics-informed signals for self-supervised training.

To further investigate this domain gap, we first experimentally verify that the performance degradation primarily arises from the feature extraction stage. We then present the results of the transferred monocular depth features on the underwater depth estimation task, which are subsequently served as the feature input to the trained diffusion-based navigation model. It is worth noting that our experimental observations show that the improvement brought by physics-informed adaptation mainly comes from the encoder rather than the decoder. This observation underlies the design of DUViN, as it suggests that the underwater domain gap can be mitigated by adapting the feature extractor through a downstream task and then integrating the adapted encoder into the navigation policy.

\begin{figure*}[t]
    \centering
    \subfloat[RGB image]{%
        \begin{minipage}[b]{0.198\textwidth}
            \centering
            \includegraphics[width=\linewidth]{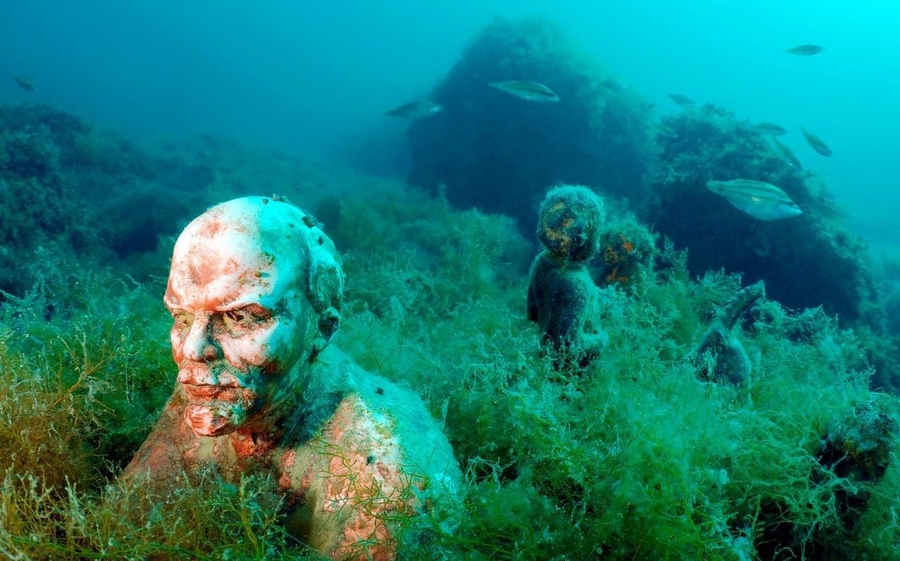}
            \par\nointerlineskip
            \includegraphics[width=\linewidth]{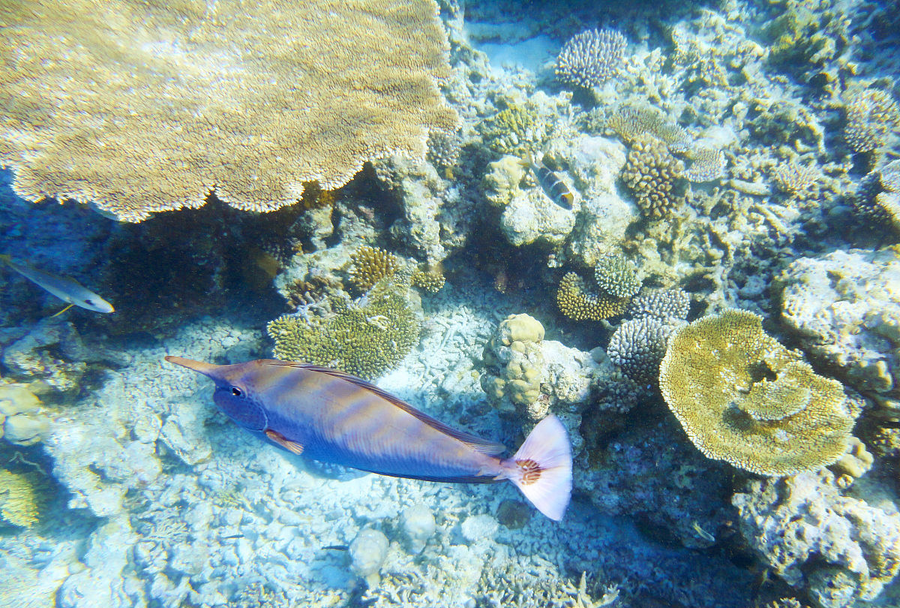}
        \end{minipage}%
    }
    \hfill
    \subfloat[DA V2: Ori]{%
        \begin{minipage}[b]{0.198\textwidth}
            \centering
            \includegraphics[width=\linewidth]{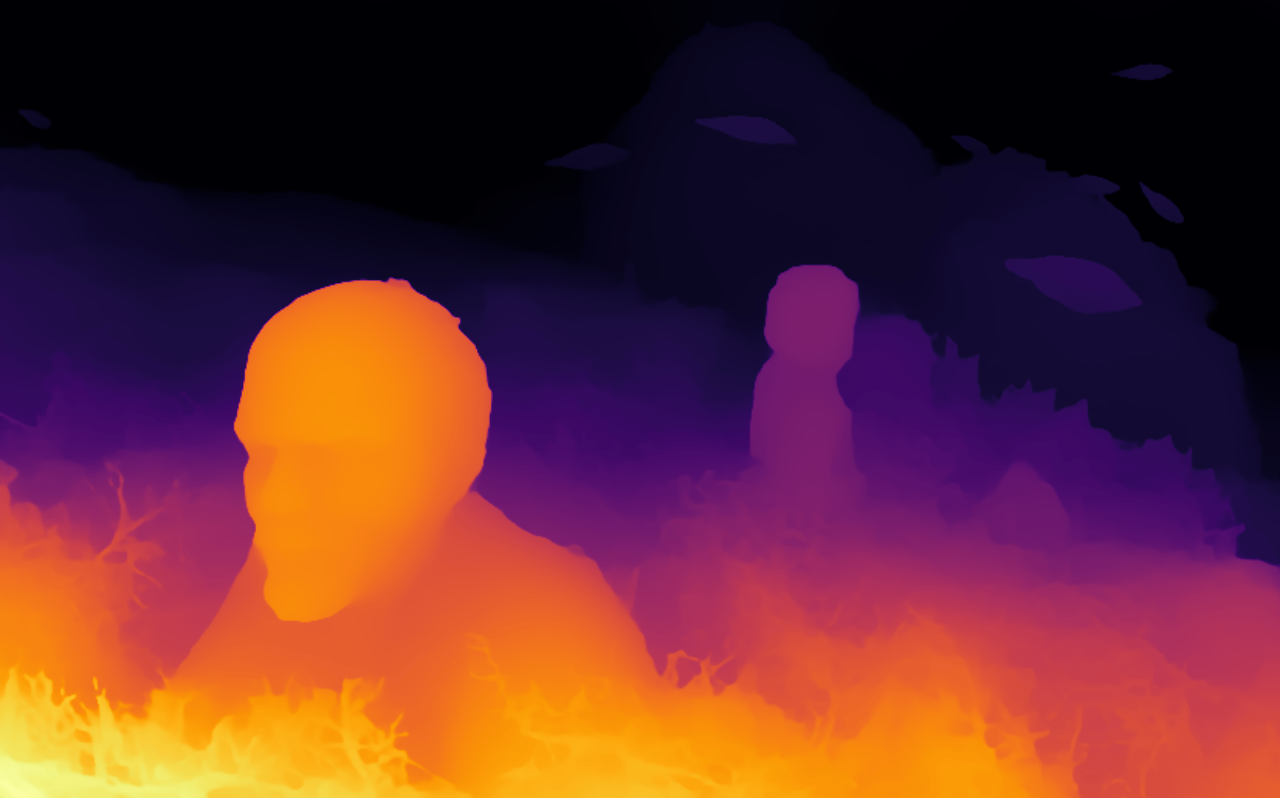}
            \par\nointerlineskip
            \includegraphics[width=\linewidth]{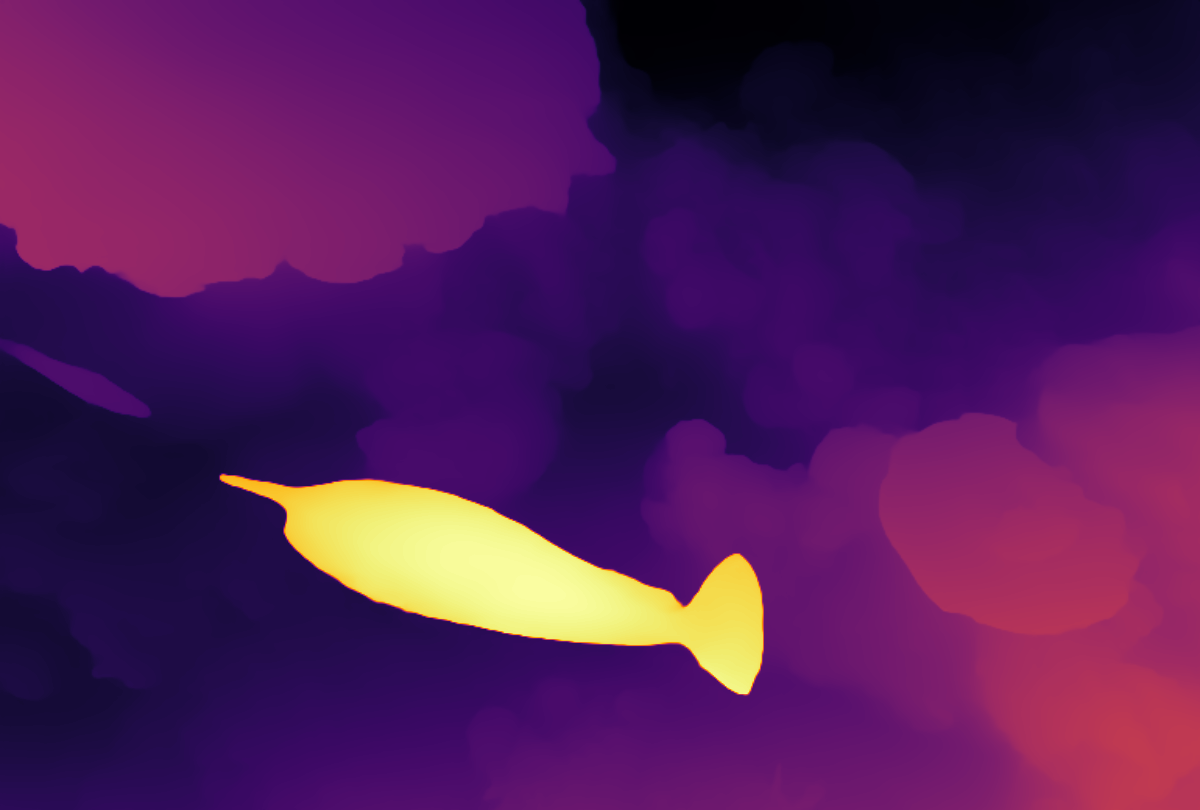}
        \end{minipage}%
    }
    \hfill
    \subfloat[DA V2: E. F.]{%
        \begin{minipage}[b]{0.198\textwidth}
            \centering
            \includegraphics[width=\linewidth]{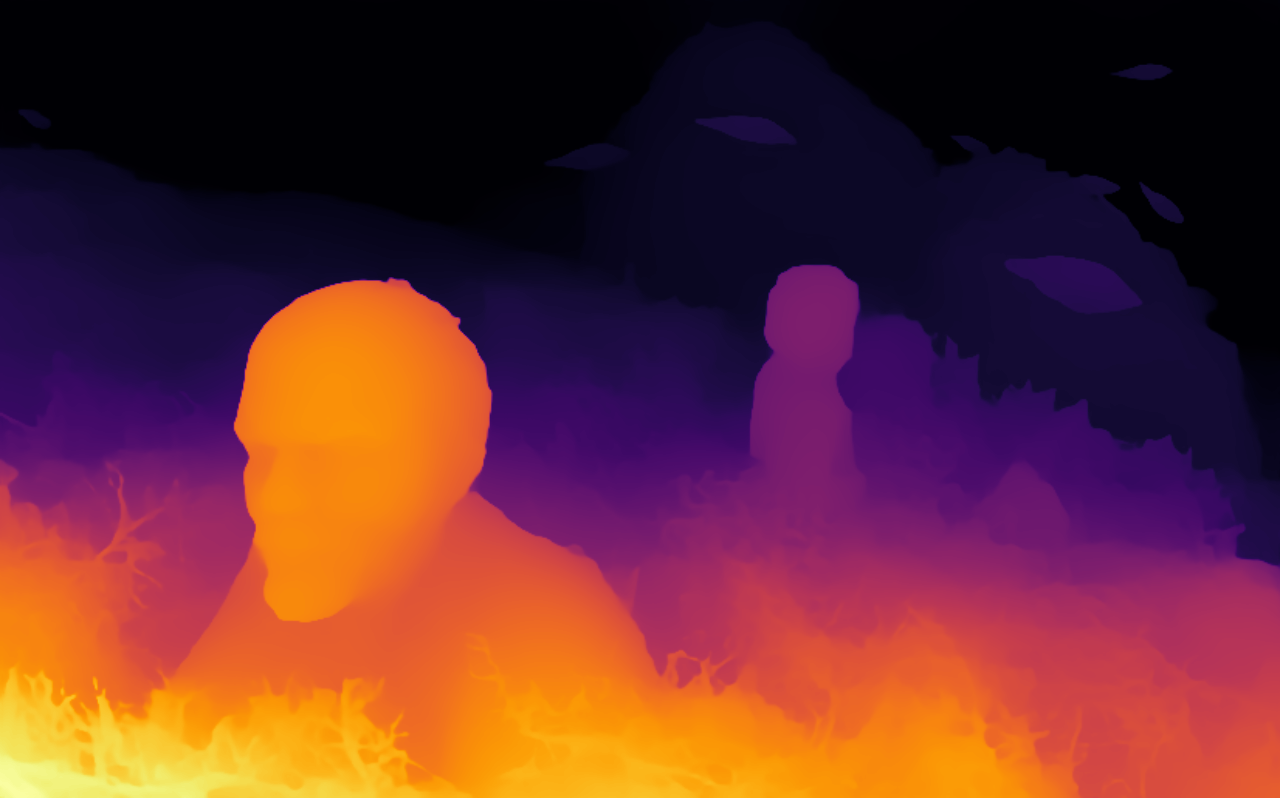}
            \par\nointerlineskip
            \includegraphics[width=\linewidth]{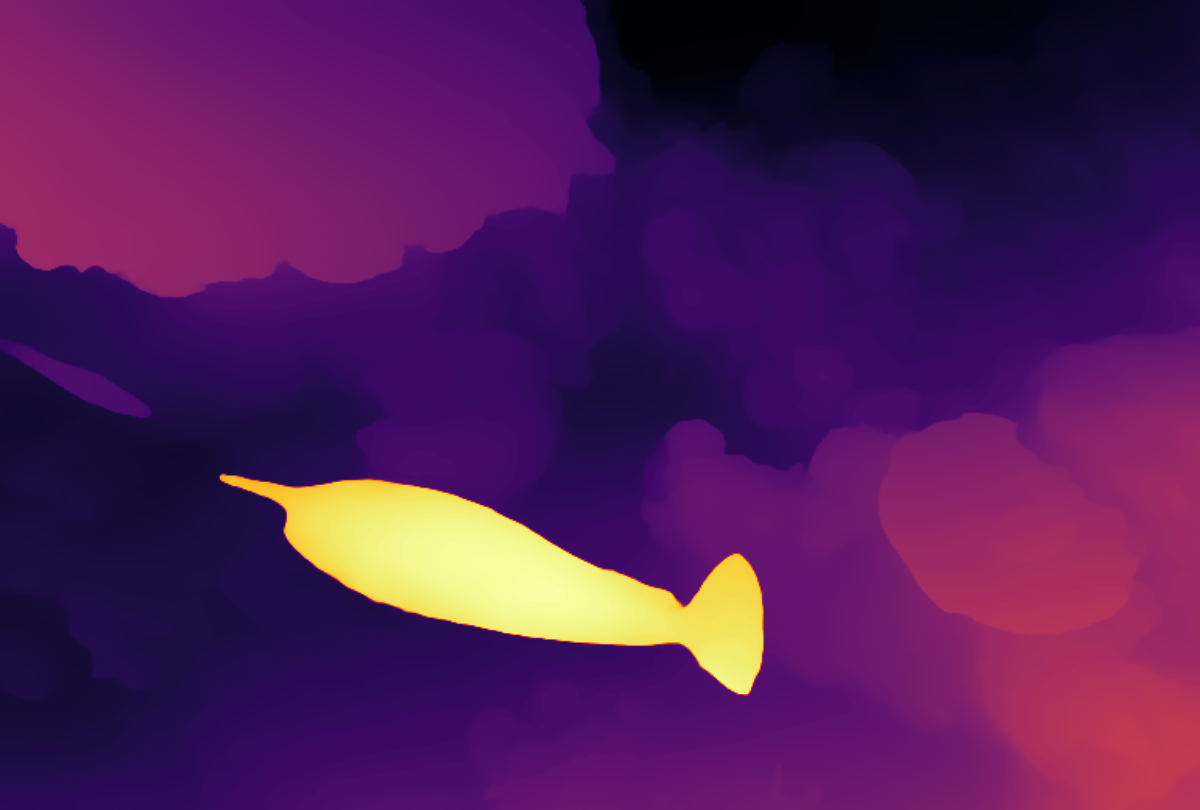}
        \end{minipage}%
    }
    \hfill
    \subfloat[DA V2: D. F.]{%
        \begin{minipage}[b]{0.198\textwidth}
            \centering
            \includegraphics[width=\linewidth]{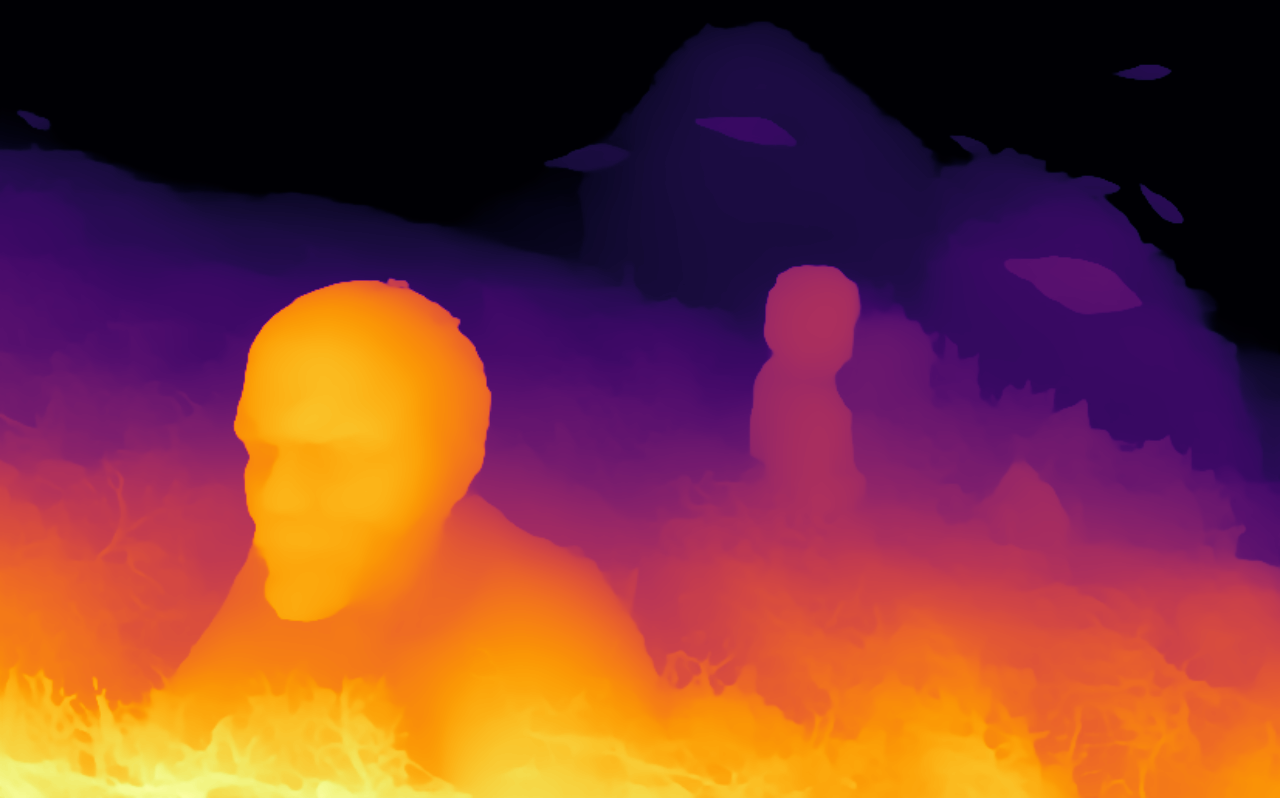}
            \par\nointerlineskip
            \includegraphics[width=\linewidth]{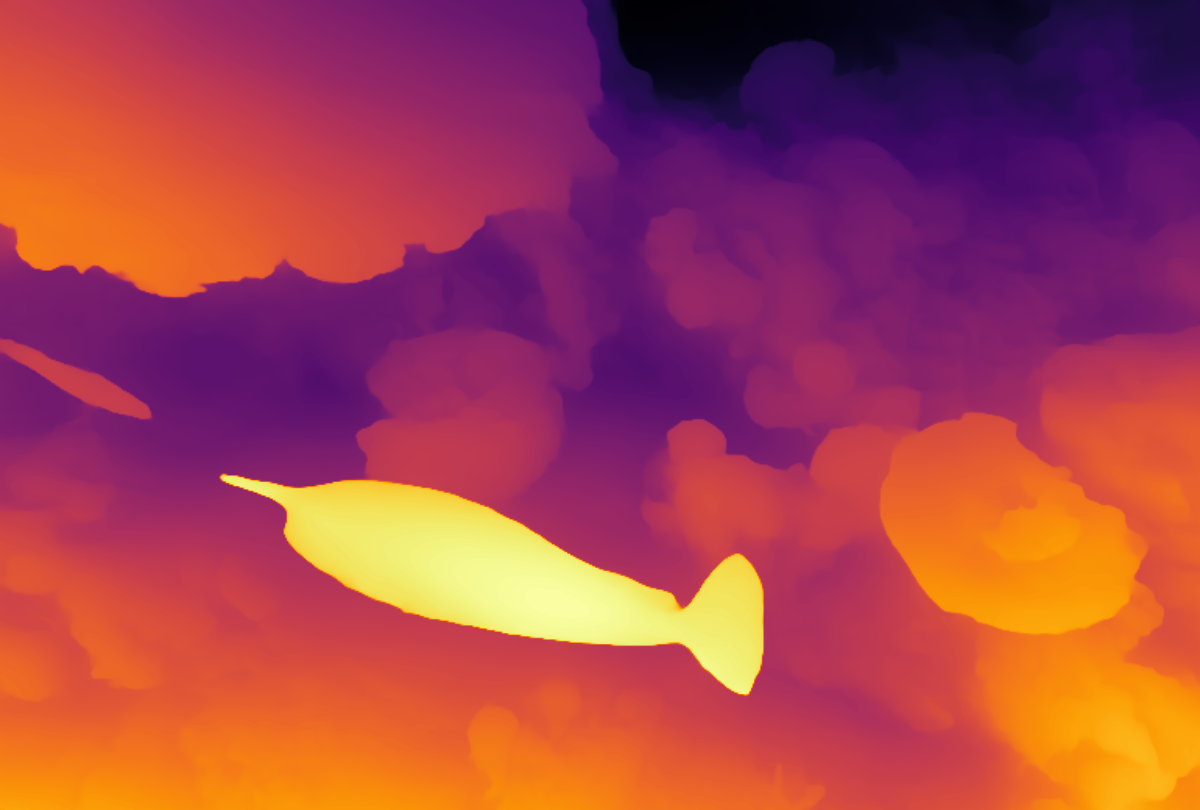}
        \end{minipage}%
    }
    \hfill
    \subfloat[DA V2: w/o F.]{%
        \begin{minipage}[b]{0.198\textwidth}
            \centering
            \includegraphics[width=\linewidth]{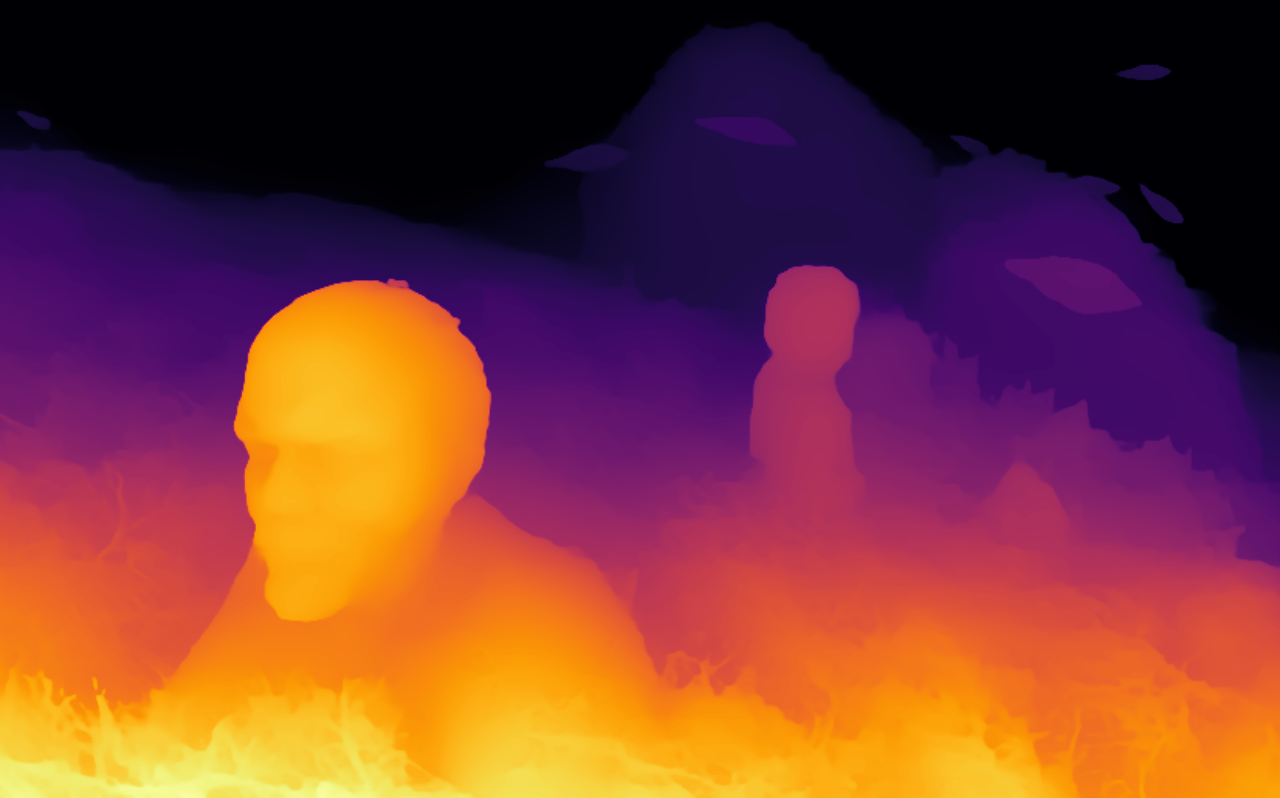}
            \par\nointerlineskip
            \includegraphics[width=\linewidth]{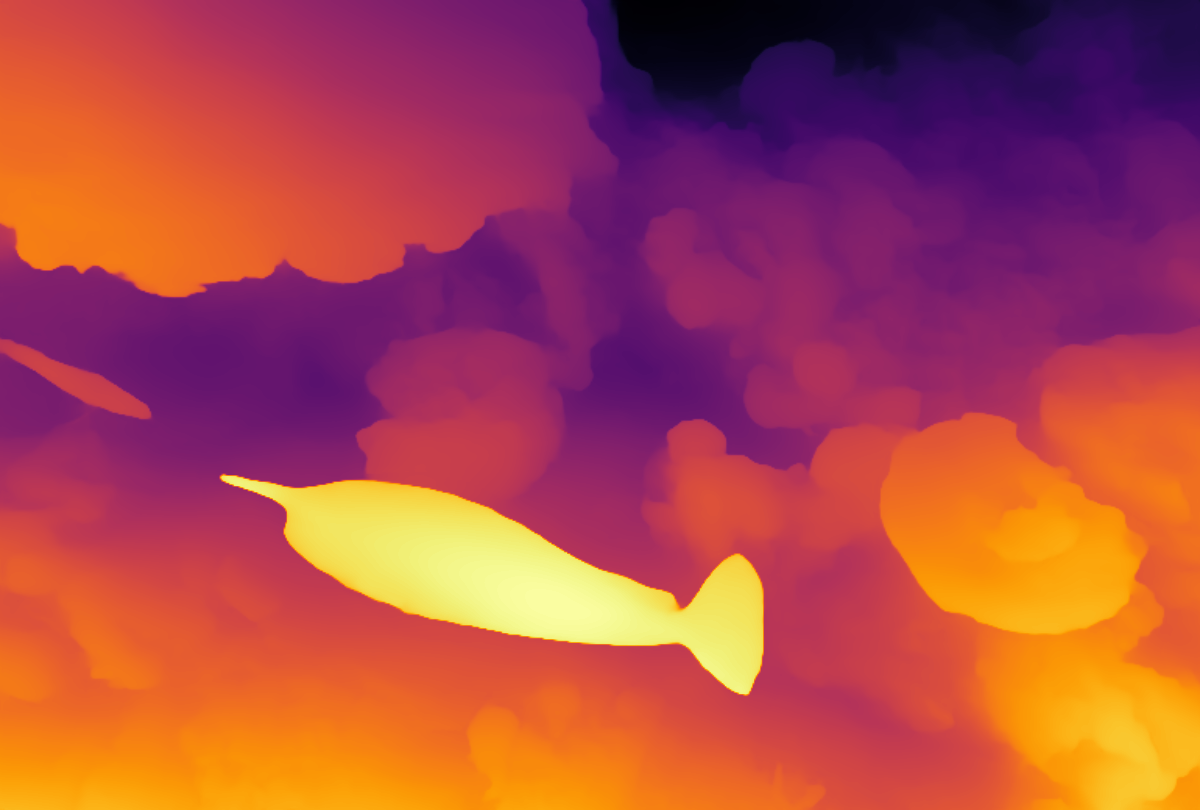}
        \end{minipage}%
    }

    \caption{\textbf{Qualitative depth estimation results of Depth Anything V2 (DA V2):} The original model struggles to detect distant regions clearly. The physics-informed knowledge-transferred models, with decoder frozen (D. F.) and without any freezing (w/o F.), show improved performance, whereas freezing the encoder (E. F.) results in minimal change.}
    \label{fig:depth_demo}
\end{figure*}

\subsection{Observations in Physics-Informed Knowledge Transfer}
Many state-of-the-art monocular depth estimation models, such as Depth Anything V2~\cite{yang2024depth}, DPT~\cite{DPT}, and MiDaS~\cite{MiDaS}, adopt an encoder–decoder architecture. These models typically employ either a transformer-based encoder (e.g., DINOv2~\cite{oquab2023dinov2}) or a convolutional encoder (e.g., ResNet) to extract image features, which are subsequently processed by a decoder to produce dense depth maps.

Then an interesting question arises: \textit{When applying PUDE, the self-supervised method that incorporates physics-based priors to refine depth predictions in underwater scenes, which component of the model contributes most to the observed improvement?}

In our implementation of PUDE on the monocular depth models, we observe that the primary performance gains occur in the feature extraction stage, suggesting that physics-informed refinement significantly enhances the encoder's ability to produce domain-adapted representations. To verify our observations, we conduct experiments on two models, Depth Anything V2 (DA V2) \cite{yang2024depth} and DPT-Hybrid (DPT-H) \cite{DPT}, with different encoder architectures. Following the same training pipeline described in PUDE, we train each model under four configurations by selectively freezing or unfreezing the encoder and decoder during training. The four settings are: the original model without training (Ori.), encoder frozen (E. F.), decoder frozen (D. F.), and full model training without freezing (w/o F.). The experiments are conducted on the SeaThru D5 \cite{sea-thru} and SQUID \cite{berman2020underwater} real underwater image dataset. We follow the evaluation metrics described in \cite{eigen2014depth} in evaluations: the mean absolute relative error (AbsRel), squared relative error (SqRel), root mean square error (RMSE) and Scale-invariant logarithmic error (SILog). The quantitative results are presented in \cref{tab:freeze_ablation}, and qualitative examples from the dataset \cite{li2019underwater} are presented in \cref{fig:depth_demo}.

With self-supervised training that incorporates underwater physics information, both Depth Anything V2 and DPT-Hybrid exhibit improved performance in underwater depth estimation. Interestingly, we observe that freezing the encoder and training only the decoder leads to negligible performance gains.  In contrast, freezing the decoder and fine-tuning only the encoder achieves performance comparable to, or even surpassing, that of full model training. This suggests that the effectiveness of physics-informed knowledge transfer primarily relies on adapting the encoder. As such, the encoder plays a crucial role in learning domain-specific representations during the transfer to underwater environments. Accordingly, we replace the depth feature encoder in the navigation model with the physics-informed, transferred encoder from the depth estimation task, see in \cref{fig:phases}. This enables the DUViN framework to transfer a navigation policy trained in clear image simulation to underwater scenarios with different visual conditions, despite the significant visual domain shift.

%% file: mainBody/4_1_Experiments.tex
\section{Experiments}
To evaluate the effectiveness and transferability of DUViN, we conduct experiments both in simulated underwater environments and in an indoor water tank. These experiments are designed to answer the following key questions:

\begin{figure}[b]
    \centering
    \subfloat[Hills]{%
        \includegraphics[width=0.241\textwidth]{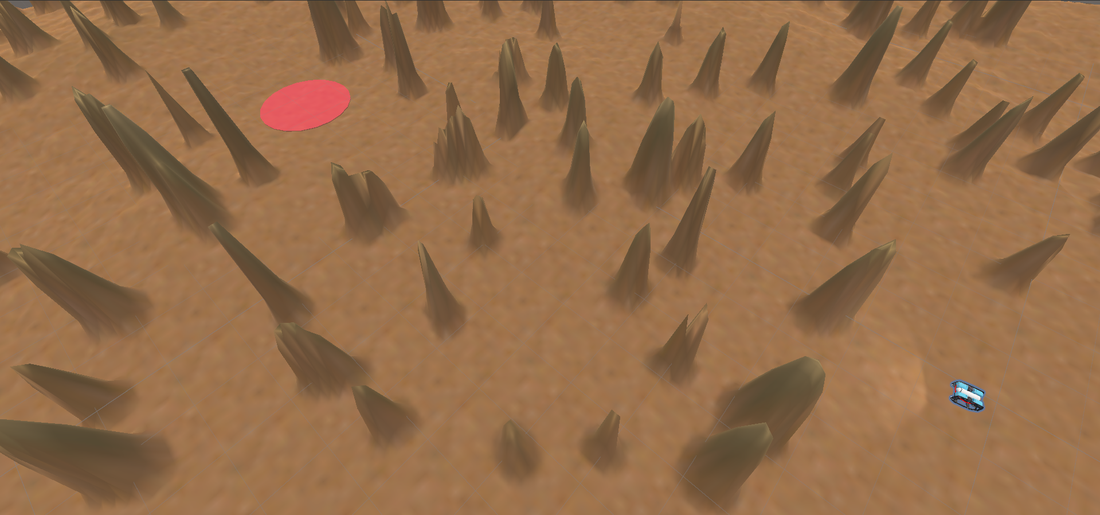}%
    }\hfill
    \subfloat[Pillars]{%
        \includegraphics[width=0.241\textwidth]{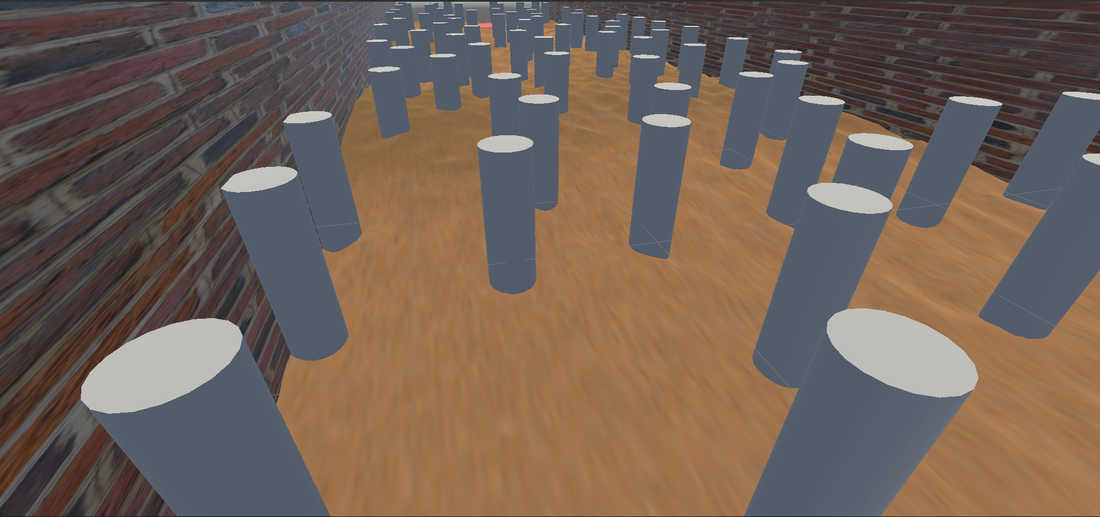}%
    }
    \caption{\textbf{Visual examples of the experimental environments.} The simulation evaluation environments \textbf{Pillars} and \textbf{Hills} are shown.}
    \label{fig:env_types}
\end{figure}

\begin{itemize}
\item Can DUViN maintain robust goal-reaching and obstacle avoidance under challenging underwater conditions with visual degradation? 
\item Is the physics-informed transfer of depth feature extractor beneficial for DUViN in underwater navigation?
\item Does the use of a reference altitude image in DUViN enable effective altitude maintenance?
\end{itemize}

In \cref{sec:simulate}, we evaluate DUViN in simulated underwater environments under three different turbidity levels, while considering the dynamics model of the AUV to assess its navigation performance. In \cref{sec:real}, we provide proof-of-concept validation of DUViN's sim-to-real transfer using a physical BlueROV2 platform in an indoor water tank.

\subsection{Implementation Details}
In our implementation, the observation time step $P$ is set to 3, with each step having a $0.5\,\mathrm{s}$ interval. The action time step $Q$ is set to $32$, with a $0.1\,\mathrm{s}$ interval per step. We generate a total of 120 random terrains with a total of $9924$ image observations, where the data from 100 terrains are used for training, and the remaining 20 are reserved for validation. We randomly sample $1000$ initial observations $o_0$ in terrains with random altitude as reference. The navigation diffusion model is trained using the AdamW optimizer with a learning rate of \( 1 \times 10^{-4} \) for 100 epochs and a batch size of 64. The denoising step \( k \) is set to 10. The discount factor $\kappa$ is set to 0.99. In velocity generation, we set \( \lambda_0 \), \( \lambda_1 \), and \( \lambda_2 \) to 0.1, 5, and 1, respectively. The distance threshold \( \epsilon \) is set to 0.2. The maximum velocity constraints $V_{max}$ are set to (0.6, 0.4, 0.2, 0.15).

\subsection{Simulation Experiments}
\label{sec:simulate}
\subsubsection{Simulation Setup}
We conduct our simulation experiments in Unity across two distinct scenarios: \textbf{Hills}, a terrain varying environment populated with randomly generated hills, and \textbf{Pillars}, a long, narrow tank densely filled with randomly placed pillars. Examples of the two environments are shown in \cref{fig:env_types}. Our underwater simulator incorporates the AUV dynamics model, hydrodynamic effects, terrain variations, and optional underwater visual effects or clear image rendering.

\begin{figure}[t]
    \centering
    \captionsetup[subfigure]{farskip=0pt}
    \subfloat[Clear]{%
        \includegraphics[width=0.24\textwidth]{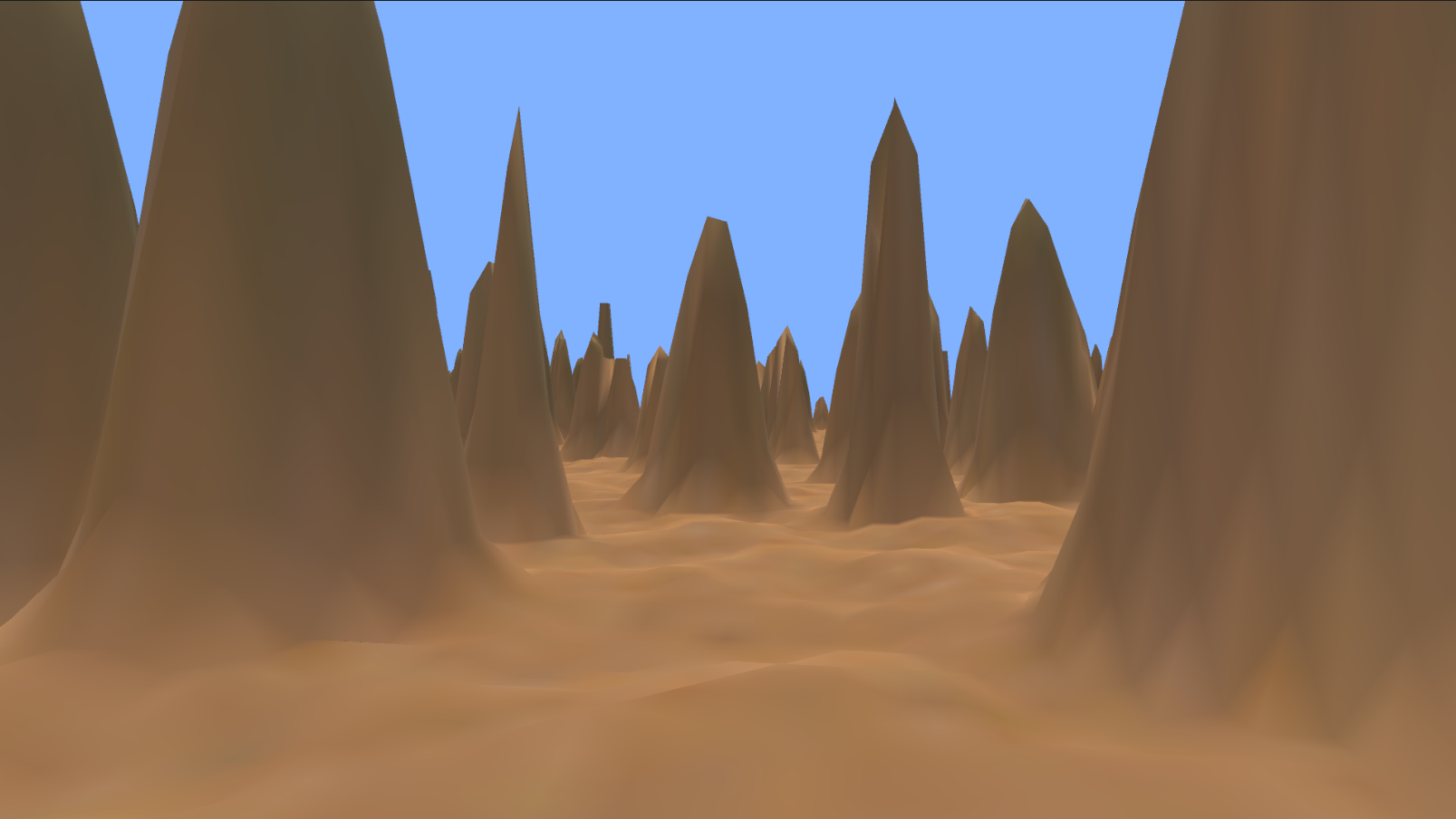}%
    }\hspace{0.5mm}%
    \subfloat[Type 1C]{%
        \includegraphics[width=0.24\textwidth]{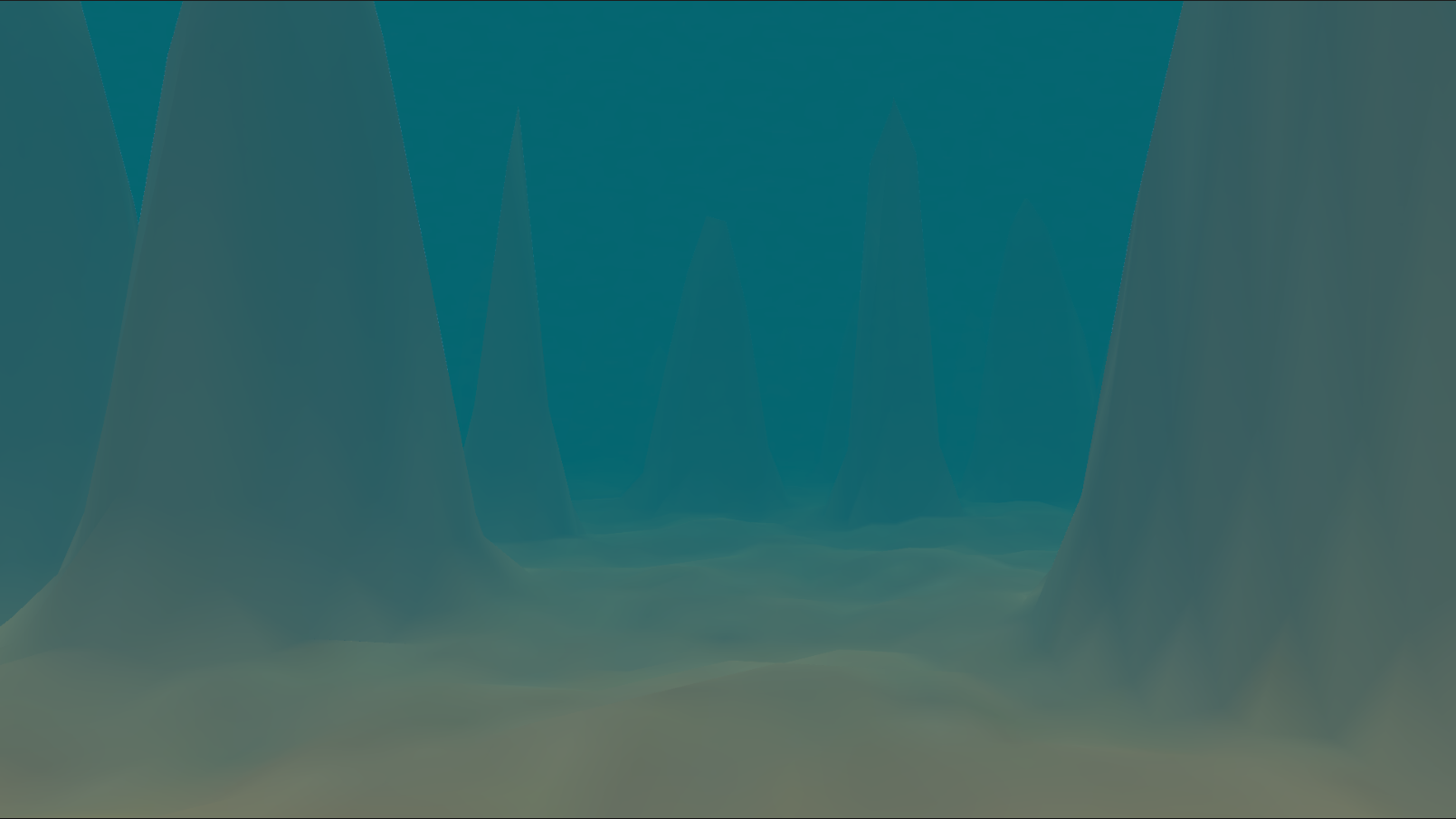}%
    }
    \par\vspace{1mm}
    \subfloat[Type 3C]{%
        \includegraphics[width=0.24\textwidth]{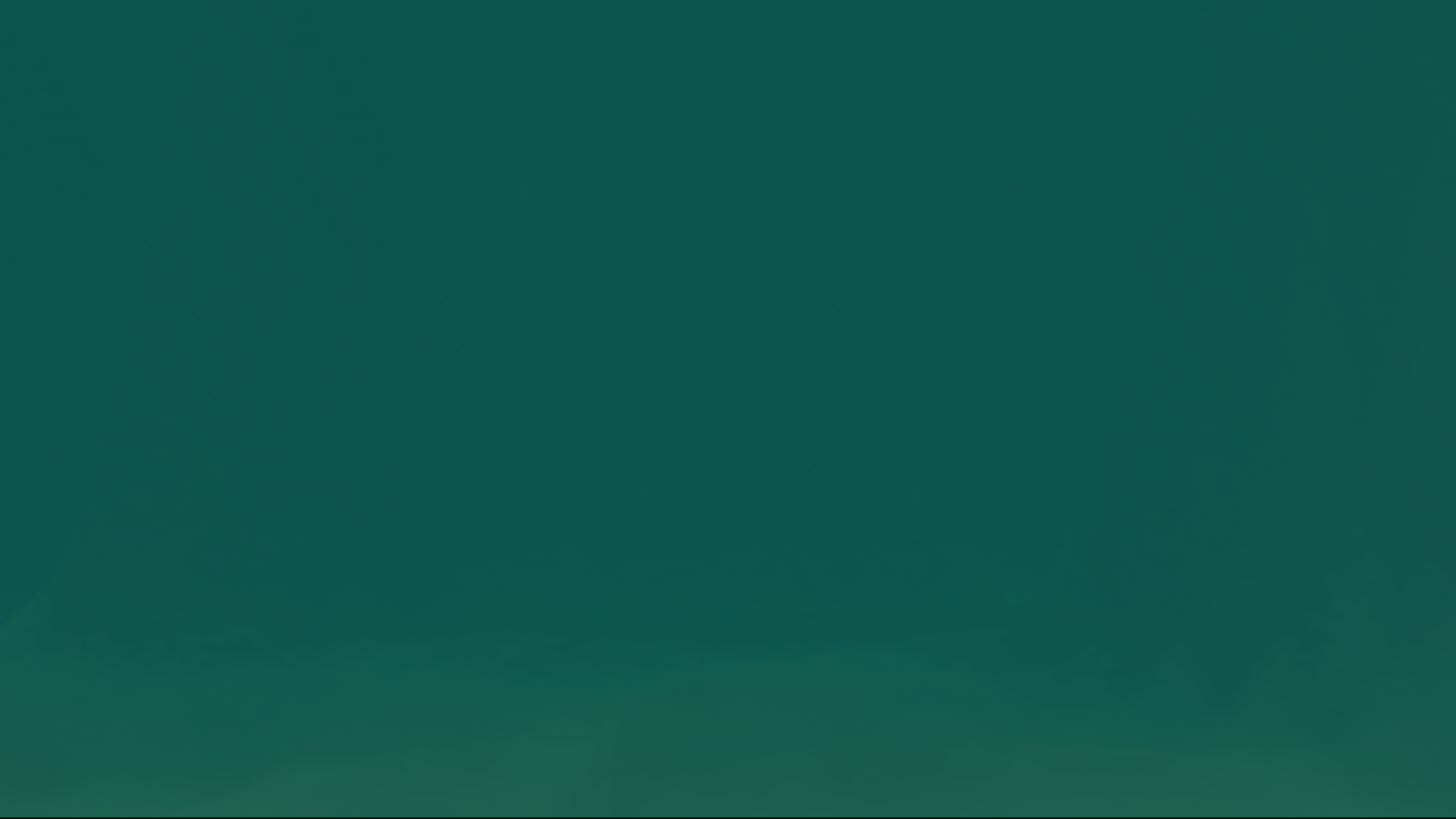}%
    }\hspace{0.5mm}%
    \subfloat[Type 7C]{%
        \includegraphics[width=0.24\textwidth]{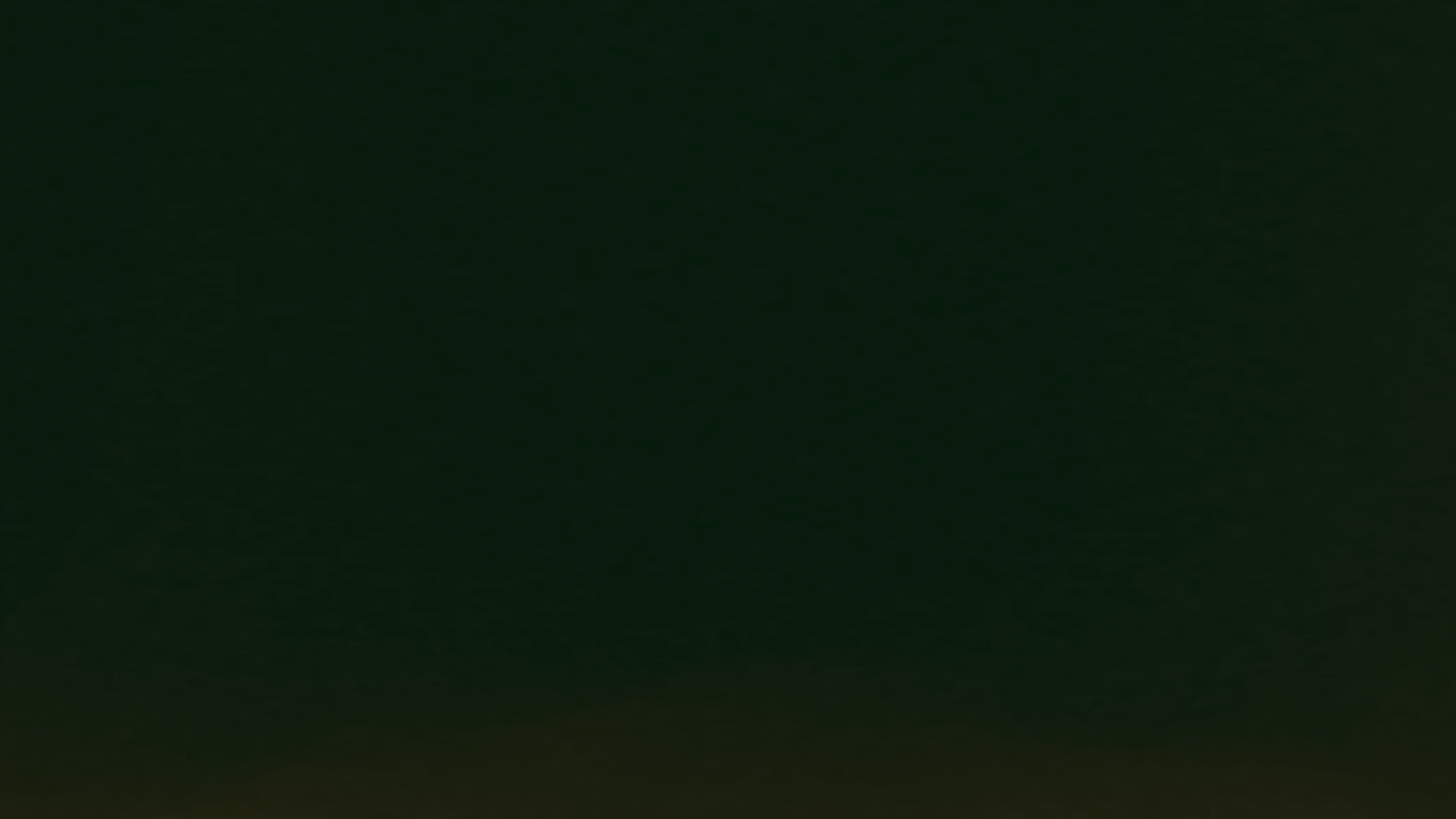}%
    }
    \caption{\textbf{Visual examples of water types.} Visual examples of three evaluation underwater environments with different levels of turbidity.}
    \label{fig:water_types}
\end{figure}

\begin{table}[b]
\centering
\caption{Water Type Parameters in Simulator. \textbf{V. R.} denotes the visibility range in meters}
\label{tab:underwater_parameters}
\begin{tabular}{|c|c|c|c|}
\noalign{\hrule height 1pt} 
\textbf{Para.} & \textbf{Type 1C} & \textbf{Type 3C} & \textbf{Type 7C} \\
\noalign{\hrule height 1pt} 
\(\beta_c\) & \{0.83, 0.44, 0.55\} & \{3.18, 1.14, 1.47\} & \{3.18, 3.55, 4.40\} \\
\hline
\(B_c^\infty\) & \{0.01, 0.40, 0.44\} & \{0.03, 0.32, 0.31\} & \{0.01, 0.09, 0.03\} \\
\hline
\(\sigma_h\) & 0.1 & 0.2 & 0.3 \\
\hline
\(k_M\) & 0.001 & 0.002 & 0.003 \\
\noalign{\hrule height 1pt} 
\textbf{V. R.}& 10.47 m & 4.04 m & 1.45 m \\
\noalign{\hrule height 1pt} 
\end{tabular}
\end{table}

\textbf{Underwater simulation environment:} To simulate diverse underwater visual conditions with varying turbidity, we extend the physics-based underwater image formation model in Eq.~\eqref{eq:underwater_image_formation}, following~\cite{yang2023knowledge,yang2025knowledge}, by additionally incorporating forward scattering and noise induced by marine snow. The extended model is formulated as follows:
\begin{equation}
\begin{aligned}
I_c(x)
&=(J_c\ast h_f)(x)\cdot e^{-\beta_c z(x)}\\
&\quad+B_c^\infty\bigl(1-e^{-\beta_c z(x)}\bigr)
+B_c^{\text{N}}(x),
\end{aligned}
\end{equation}
where \( J_c(x) \) denotes the pixel intensity of the clear scene at pixel location \( x \) for color channel \( c \in \{R, G, B\} \), and \( z(x) \) is the distance from the camera to the scene point. The term \( h_f \) is the depth-dependent point spread function (PSF), which models forward scattering as a Gaussian blur whose standard deviation is proportional to distance, specifically \( \sigma_h \cdot z(x) \). The terms \( B_c^\infty \) and \( B_c^{\text{N}}(x) \) represent the background ambient light and the image noise caused by the marine snow and suspended particles, respectively. \( B_c^{\text{N}}(x) \) is computed as follows:
\begin{equation}
B_c^{\text{N}}(x) = k_M M(x) \left( 1 - B_c^\infty \right) \left( 1 - e^{-\beta_c z(x)} \right),
\end{equation}
where \( M(x) \in [0, 1] \) is the Perlin noise map with a strength ratio denoted as \( k_M \). We simulate three water conditions (Type 1C, 3C, 7C) from clear to turbid based on the Jerlov water types \cite{Jerlov}. The corresponding parameters are summarized in \cref{tab:underwater_parameters}, the visual examples are shown in \cref{fig:water_types}.

\begin{figure}[t]
    \centering
    \includegraphics[width=0.56\linewidth]{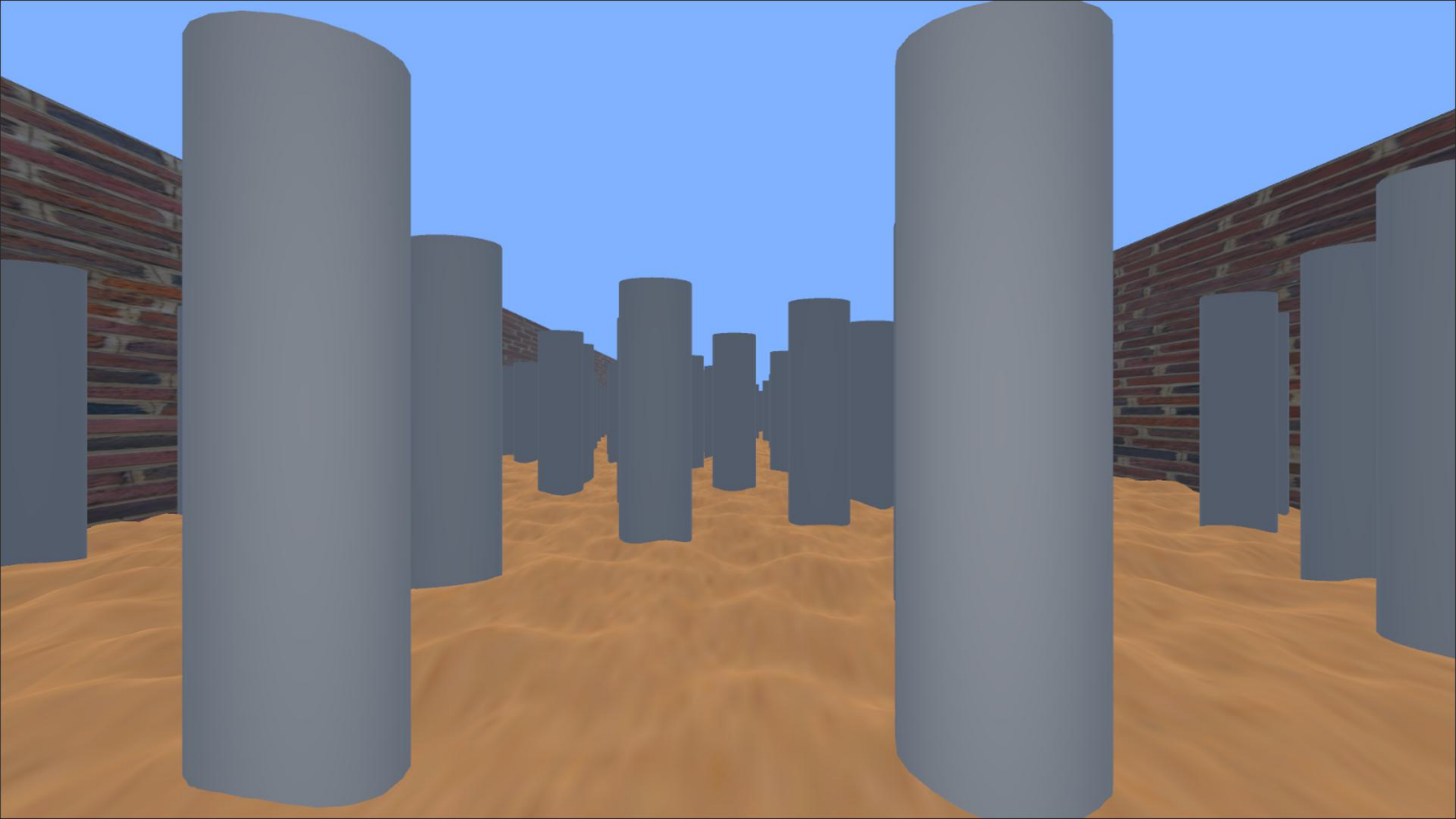}\hfill%
    \includegraphics[width=0.42\linewidth]{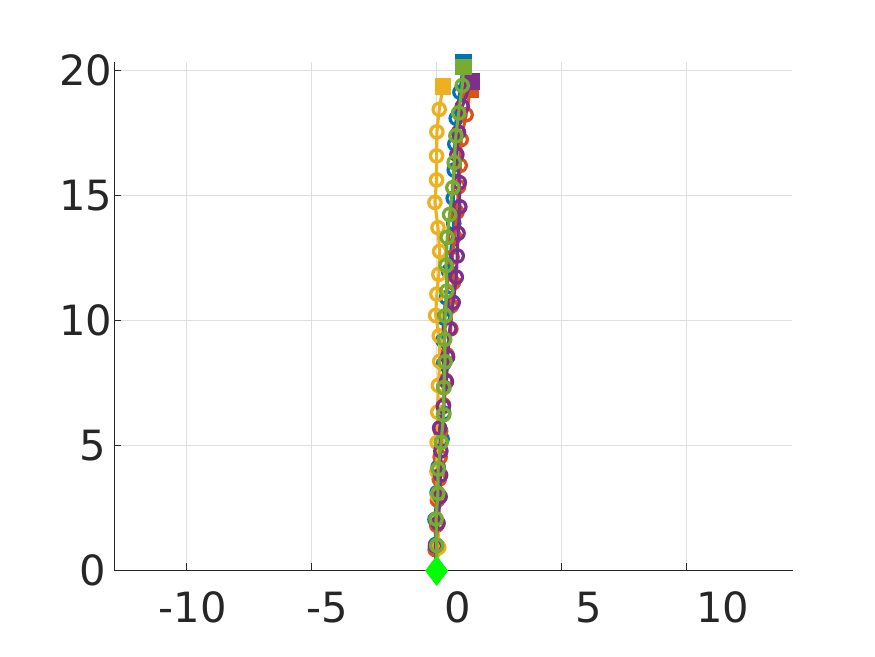}
    \par\vspace{0.5mm}
    {\footnotesize Clear image testing environment\par}
    \vspace{2mm}
    \includegraphics[width=0.56\linewidth]{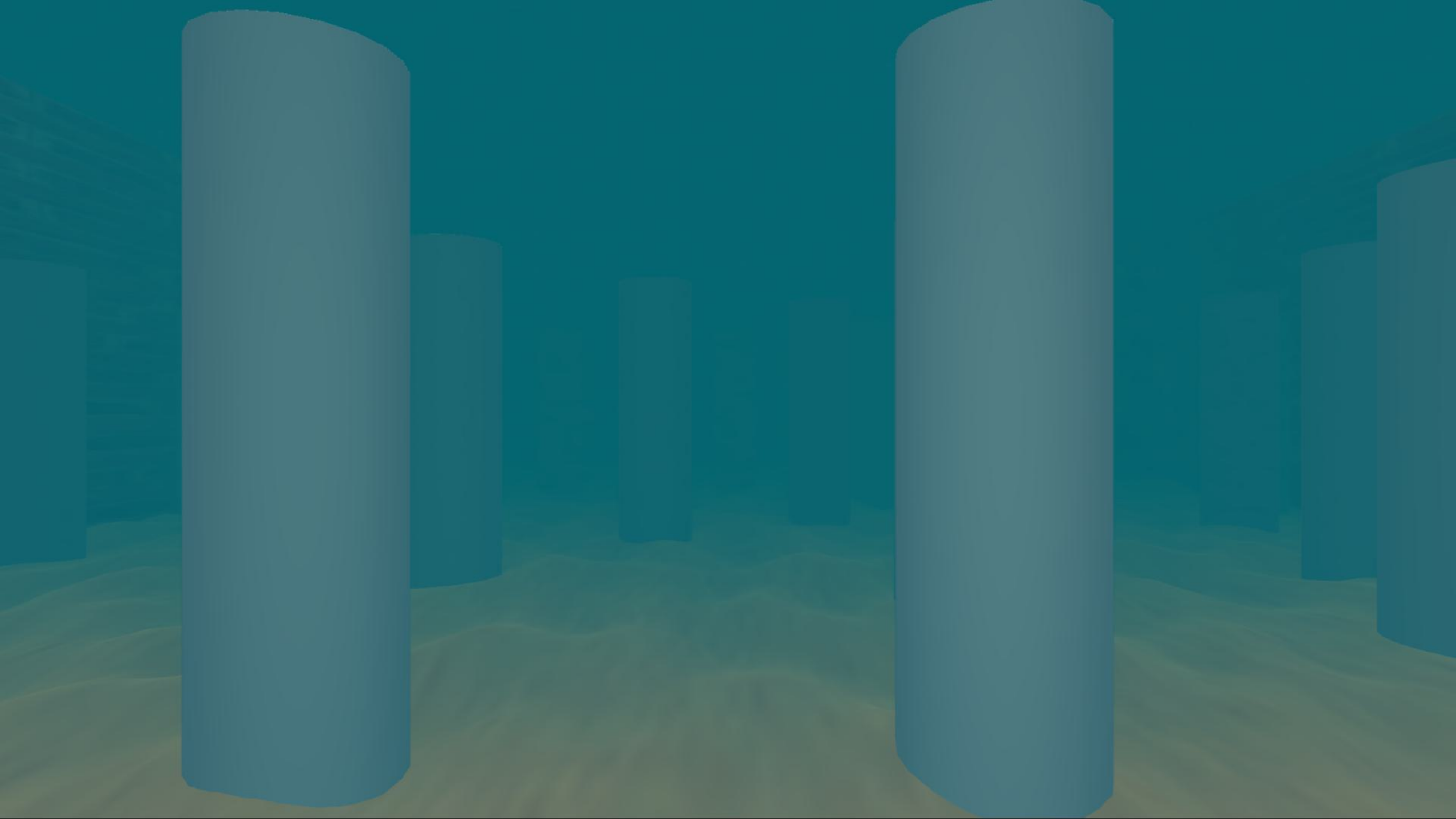}\hfill%
    \includegraphics[width=0.42\linewidth]{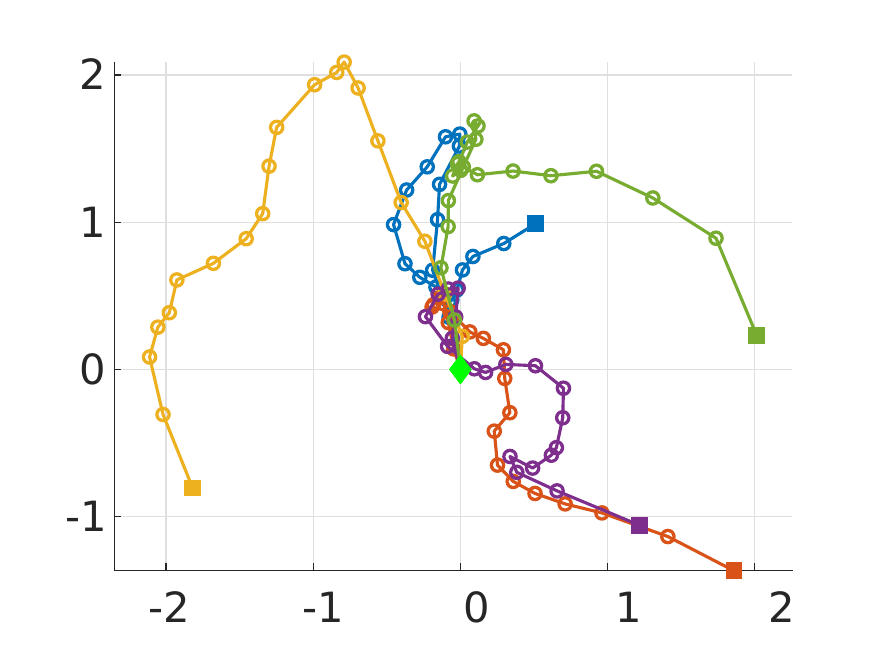}
    \par\vspace{0.5mm}
    {\footnotesize Type 1C testing environment\par}
    \caption{\textbf{NoMaD failure cases under clear image and underwater conditions.}
    Example observations across water types, highlighting the difficulty in degraded visibility.}
    \label{fig:simulate_nomad}
\end{figure}

\textbf{AUV dynamics model:} To simulate the realistic behaviors of the AUV, we employ the 6-DoF dynamic model described in~\cite{fossen2011handbook}. The dynamics are formulated as follows:
\begin{equation}
\mathcal{M}\dot{\nu} + \mathcal{C}(\nu)\nu + \mathcal{D}(\nu)\nu + g(\eta) = \tau_f,
\label{auv_dynamics}
\end{equation}
where \( \nu \in \mathbb{R}^6 \) represents the AUV's linear and angular velocities in the body frame. \( \mathcal{M} \in \mathbb{R}^{6\times 6} \) is the inertia matrix including both rigid-body and added mass effects, \( \mathcal{C}(\nu) \in \mathbb{R}^{6\times 6} \) captures Coriolis and centripetal forces, and \( \mathcal{D}(\nu) \in \mathbb{R}^{6\times 6} \) models hydrodynamic damping. The term \( g(\eta) \in \mathbb{R}^{6} \) represents restoring forces due to gravity and buoyancy. The torque input is denoted by \( \tau_f \in \mathbb{R}^{6}\). The definitions of the above dynamic terms are detailed in~\cite{fossen2011handbook}, and the parameter values of the AUV model are based on the BlueROV2 Heavy configuration, as referenced in~\cite{von2022open}.

Given the generated velocity commands \( \mathbb{A}_t \) in 4 DoF, the remaining velocities in pitch and roll are regulated to zero. We employ a simple PD controller for tracking, with the input torque defined as:
\begin{equation}
\tau_f = k_p e + k_d \dot{e},
\end{equation}
where \( e \in \mathbb{R}^{6} \) denotes the velocity error, and \( \dot{e}  \in \mathbb{R}^{6}\) represents its time derivative. The proportional gain \( k_p \) and the derivative gain \( k_d \) are both diagonal matrices in \( \mathbb{R}^{6 \times 6} \). 

\begin{table*}[t]
\centering
\caption{Obstacle Avoidance Performance Under Different Water Conditions at Fixed Altitude}
\resizebox{\textwidth}{!}{%
\begin{tabular}{c|ccc|ccc|ccc|ccc}
\toprule
\textbf{Water Types} & \multicolumn{3}{c|}{w/o Water}  & \multicolumn{3}{c|}{Type 1C} & \multicolumn{3}{c|}{Type 3C} & \multicolumn{3}{c}{Type 7C} \\
\midrule
\textbf{Metrics} & Succ.\%~$\uparrow$ & CF\%~$\uparrow$ & A.\ C.~$\downarrow$ & Succ.\%~$\uparrow$ & CF\%~$\uparrow$ & A.\ C.~$\downarrow$ & Succ.\%~$\uparrow$ & CF\%~$\uparrow$ & A.\ C.~$\downarrow$ & Succ.\%~$\uparrow$ & CF\%~$\uparrow$ & A.\ C.~$\downarrow$ \\
\midrule
NoMaD~\cite{sridhar2024nomad} & 85 & 15 & 2.05 & 0 & 0 & \text{N/A} & 0 & 0 & \text{N/A} & 0 & 0 & \text{N/A} \\
\midrule
\textbf{DUViN-air} & \textbf{100} & \textbf{65} & 0.45 &  \textbf{100} & 50 & 0.60 &  95 & 40 & 0.85 & 95 & 25 & 1.20 \\
\midrule
\textbf{DUViN} & \textbf{100} & \textbf{65} & \textbf{0.40} & \textbf{100} & \textbf{60} & \textbf{0.45} & \textbf{100} & \textbf{60} & \textbf{0.50} & \textbf{100} & \textbf{55} & \textbf{0.65} \\
\bottomrule
\end{tabular}
}
\label{simualte_inair}
\end{table*}

\begin{table*}[t]
\centering
\caption{Obstacle Avoidance Performance in 4-DoF in \textbf{Hills} and \textbf{Pillars} Scenarios With Varying Terrain}
\resizebox{\textwidth}{!}{%
\begin{tabular}{c|ccc|ccc|ccc|ccc}
\toprule
\textbf{Map} & \multicolumn{12}{c}{Hills} \\
\midrule
\textbf{Water Types} & \multicolumn{3}{c|}{w/o Water}  & \multicolumn{3}{c|}{Type 1C} & \multicolumn{3}{c|}{Type 3C} & \multicolumn{3}{c}{Type 7C} \\
\midrule
\textbf{Metrics} & Succ.\%~$\uparrow$ & CF\%~$\uparrow$ & A.\ C.~$\downarrow$ & Succ.\%~$\uparrow$ & CF\%~$\uparrow$ & A.\ C.~$\downarrow$ & Succ.\%~$\uparrow$ & CF\%~$\uparrow$ & A.\ C.~$\downarrow$ & Succ.\%~$\uparrow$ & CF\%~$\uparrow$ & A.\ C.~$\downarrow$ \\

\midrule
DUViN-air & \textbf{100} & \textbf{90} & \textbf{0.10} & 80 &  60 & 0.25 & 70 & 10 & 0.93 & 70 & 5 & 1.29 \\
\textbf{DUViN} &  95 & \textbf{90} & 0.11 & \textbf{85} &  \textbf{70} &  \textbf{0.18} & \textbf{90} &  \textbf{45} & \textbf{0.56} & \textbf{80} & \textbf{40} & \textbf{0.69} \\
\bottomrule
\end{tabular}
}

\vspace{1mm}

\resizebox{\textwidth}{!}{%
\begin{tabular}{c|ccc|ccc|ccc|ccc}
\toprule
\textbf{Map} & \multicolumn{12}{c}{Pillars} \\
\midrule
\textbf{Water Types} & \multicolumn{3}{c|}{w/o Water}  & \multicolumn{3}{c|}{Type 1C} & \multicolumn{3}{c|}{Type 3C} & \multicolumn{3}{c}{Type 7C} \\
\midrule
\textbf{Metrics} & Succ.\%~$\uparrow$ & CF\%~$\uparrow$ & A.\ C.~$\downarrow$ & Succ.\%~$\uparrow$ & CF\%~$\uparrow$ & A.\ C.~$\downarrow$ & Succ.\%~$\uparrow$ & CF\%~$\uparrow$ & A.\ C.~$\downarrow$ & Succ.\%~$\uparrow$ & CF\%~$\uparrow$ & A.\ C.~$\downarrow$ \\

\midrule
DUViN-air & \textbf{100} & \textbf{55} & \textbf{0.55} & \textbf{100} & 50 & 0.85 & 95 & 20 & 1.65 & 70 & 15 & 2.05 \\
\textbf{DUViN} & \textbf{100} & \textbf{55} & 0.60 & \textbf{100} & \textbf{60} & \textbf{0.55} & \textbf{100} & \textbf{40} & \textbf{1.05} & \textbf{85} & \textbf{40} & \textbf{1.21} \\
\bottomrule
\end{tabular}
}
\label{simualte_transfer}
\end{table*}

\subsubsection{Obstacle Avoidance Performance Comparisons}
We first compare the proposed DUViN navigation model with the state-of-the-art visual navigation method, NoMaD~\cite{sridhar2024nomad}, which is trained on RGB observations and directly extracts image features for policy inference. This comparison aims to highlight the advantages of our method in leveraging depth features for underwater navigation and to validate the effectiveness of the transferred feature extractor.

To this end, we evaluate NoMaD and two variants of DUViN: one employing the original in-air depth feature extractor without adaptation (DUViN-air), and the other using the transferred depth feature extractor trained via PUDE~\cite{yang2024physics}. Since NoMaD does not incorporate altitude control and adopts a different goal guidance strategy, we design the experiments within the randomly generated \textbf{Pillars} scenarios with a flat terrain surface, and the altitude of the AUV is fixed during the experiments at 0.4 meters. Due to the narrow width and the presence of side walls in the \textbf{Pillars} environment, NoMaD can guide the AUV to the end of the tank in an exploration mode without explicit goal input. We evaluate performance across three underwater conditions, as well as a baseline setting without underwater effects (w/o Water), to assess the models' performance under varying levels of visual degradation. We use the following evaluation metrics in our experiments:
\begin{itemize}
    \item \textbf{Success rate} (Succ.\%): The percentage of trials in which the AUV reaches the end of the tank (goal) with three or fewer collisions.
    \item \textbf{Collision-free rate} (CF\%): The percentage of trials in which the AUV reaches the end of the tank without any collision.
    \item \textbf{Average collisions} (A.\ C.): The mean number of collisions per reached trial.
\end{itemize}

Each method is evaluated over 20 independent trials. The results are summarized in \cref{simualte_inair}. The results demonstrate that NoMaD, which is trained on RGB images, struggles to adapt to underwater conditions. While NoMaD exhibits robust performance in unseen in-air environments, its navigation performance degrades significantly when transferred to underwater scenarios. Specifically, under clear image conditions (w/o Water), NoMaD can navigate towards the end of the tank but with a higher collision rate compared to both DUViN-air and DUViN. However, under all three underwater conditions, NoMaD fails to reach the end of the tank. It instead generates waypoints that exhibit roundabout or evasive behaviors, resembling those triggered by obstacle avoidance near walls. Representative examples are illustrated in \cref{fig:simulate_nomad}.

\begin{figure}[t]
    \centering
    \subfloat[Pillars testing environment\label{fig:duvin_failure_a}]{%
        \includegraphics[width=0.48\textwidth, trim=80 30 80 30, clip]{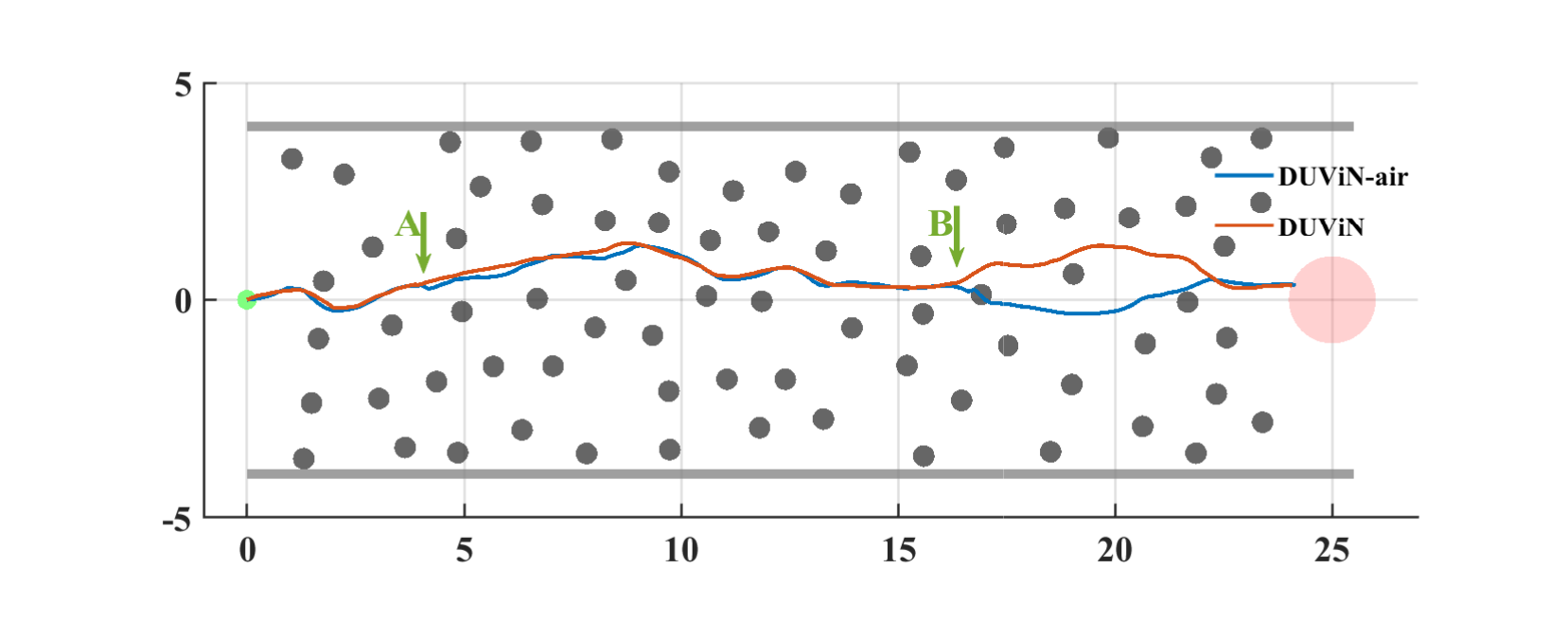}%
    }
    \par
    \subfloat[Feature-level analysis of Scene A\label{fig:scene_a_analysis}]{%
        \begin{minipage}{0.48\textwidth}
            \centering
            \begin{minipage}[b]{0.32\linewidth}
                \centering
                \includegraphics[width=\linewidth]{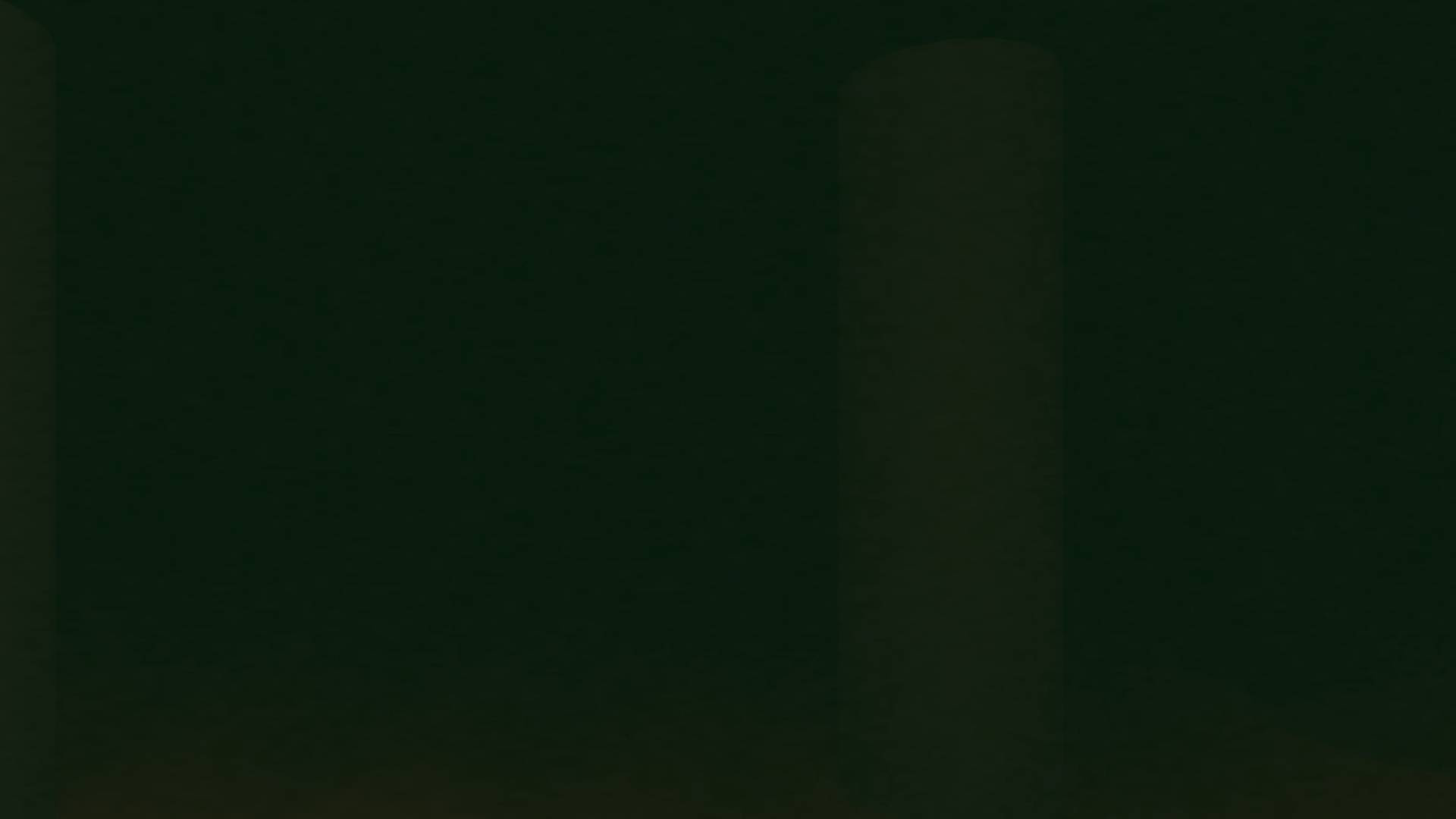}
                \par{\footnotesize Input Scene A\par}
            \end{minipage}\hspace{0.8mm}%
            \begin{minipage}[b]{0.32\linewidth}
                \centering
                \includegraphics[width=\linewidth]{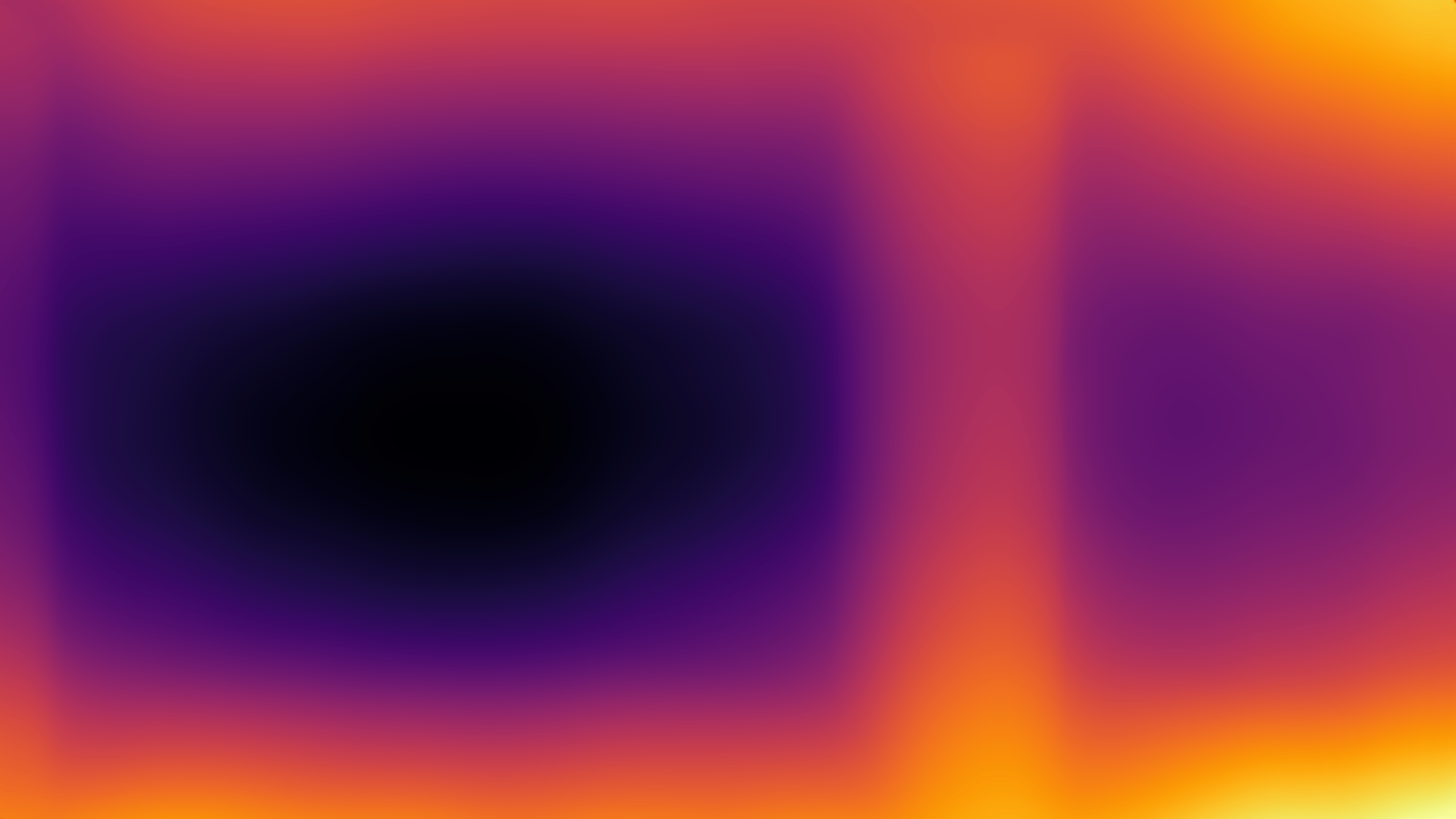}
                \par{\footnotesize DUViN-air\par}
            \end{minipage}\hspace{0.8mm}%
            \begin{minipage}[b]{0.32\linewidth}
                \centering
                \includegraphics[width=\linewidth]{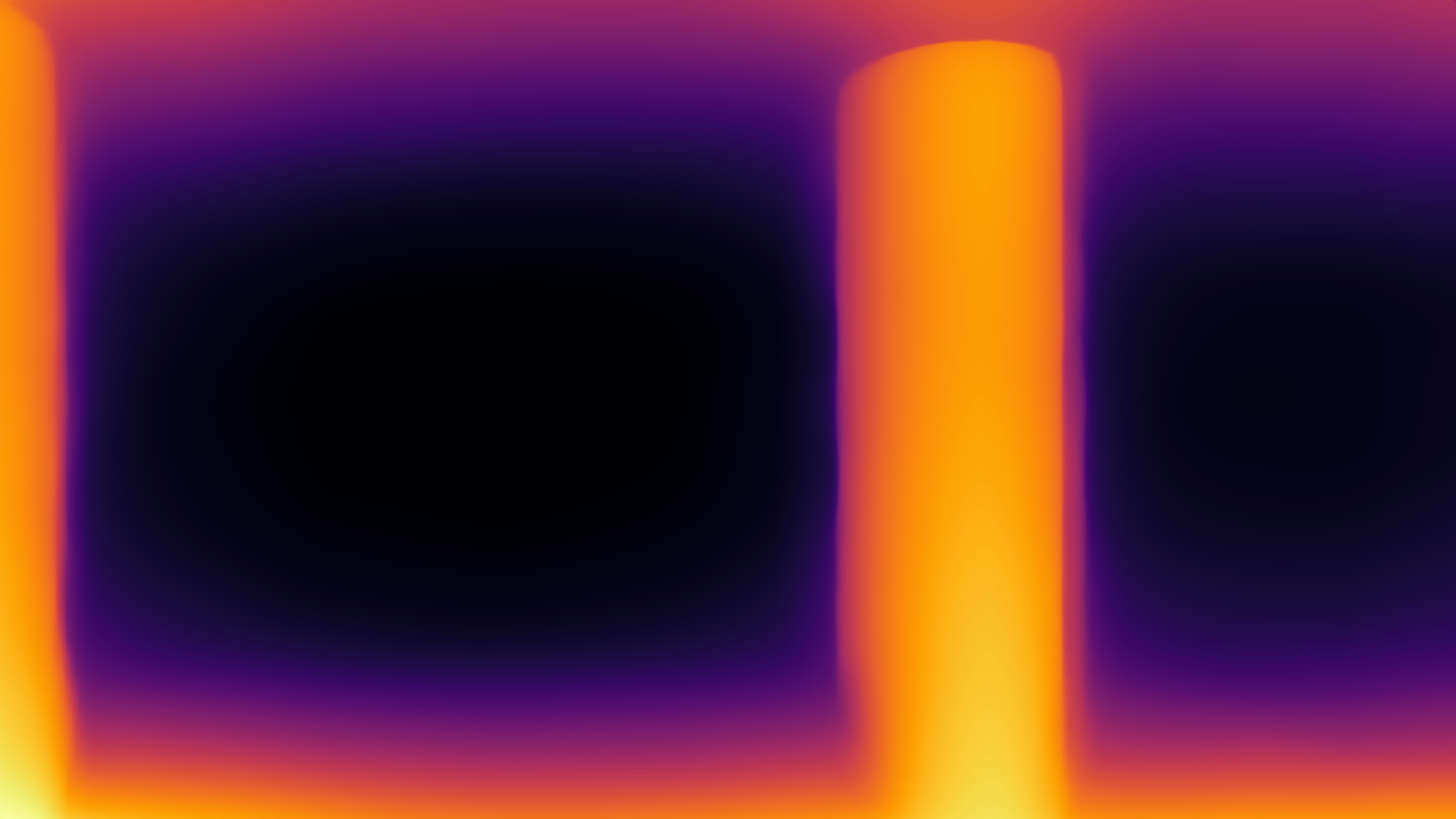}
                \par{\footnotesize DUViN\par}
            \end{minipage}\hspace{0.8mm}%
            \par
            \begin{minipage}[b]{0.21\linewidth}
                \centering
                \includegraphics[width=\linewidth]{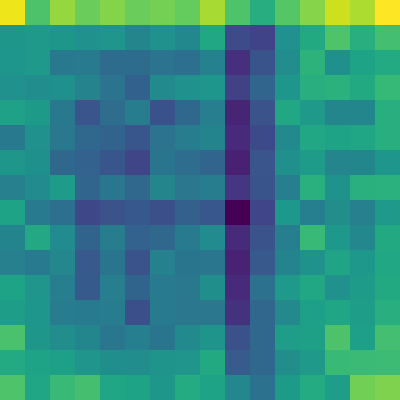}
                \par{\footnotesize DUViN(a) $\mathbb{F}_t^c$\par}
            \end{minipage}\hspace{0.8mm}%
            \begin{minipage}[b]{0.21\linewidth}
                \centering
                \includegraphics[width=\linewidth]{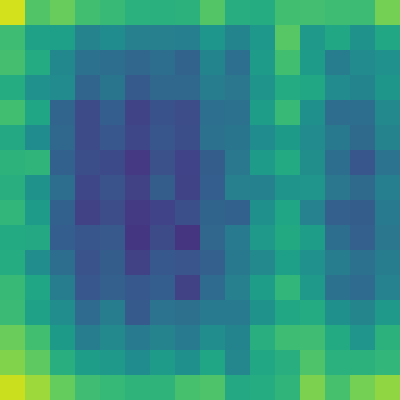}
                \par{\footnotesize DUViN $\mathbb{F}_t^{c}$\par}
            \end{minipage}\hspace{0.8mm}%
            \begin{minipage}[b]{0.21\linewidth}
                \centering
                \includegraphics[width=\linewidth]{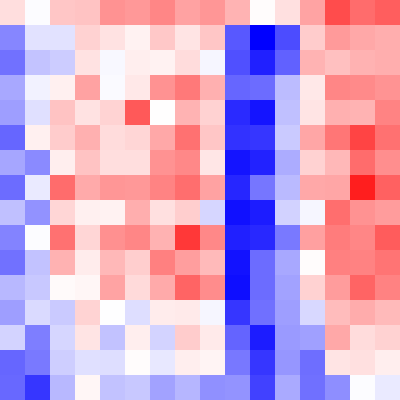}
                \par{\footnotesize Feature diff.\par}
            \end{minipage}\hspace{0.8mm}%
            \begin{minipage}[b]{0.32\linewidth}
                \centering
                \includegraphics[width=\linewidth]{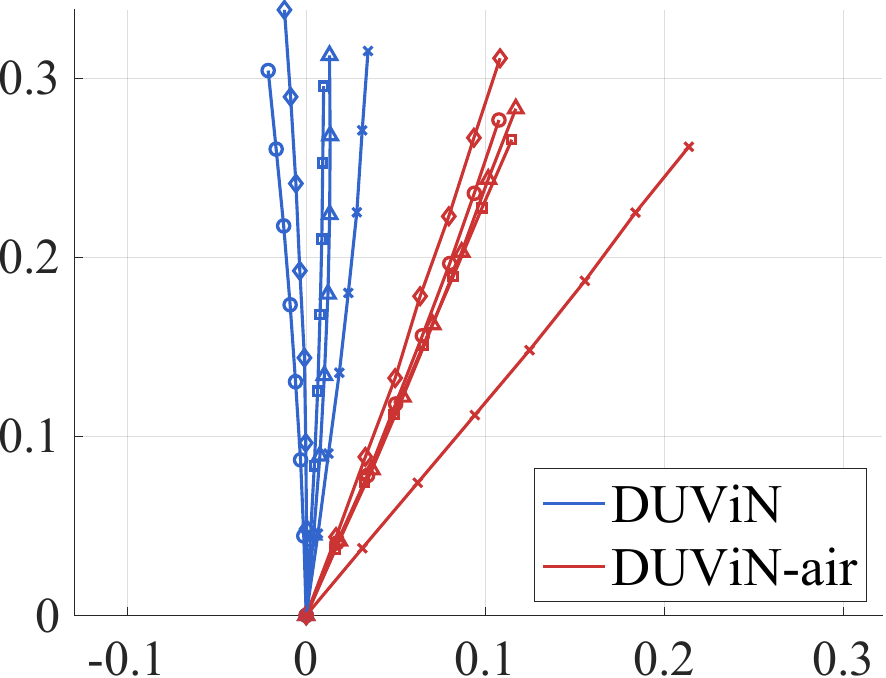}
                \par{\footnotesize Actions\par}
            \end{minipage}\hspace{0.8mm}%
        \end{minipage}%
    }
    \par
    \subfloat[Feature-level analysis of Scene B\label{fig:scene_b_analysis}]{%
        \begin{minipage}{0.48\textwidth}
            \centering
            \begin{minipage}[b]{0.32\linewidth}
                \centering
                \includegraphics[width=\linewidth]{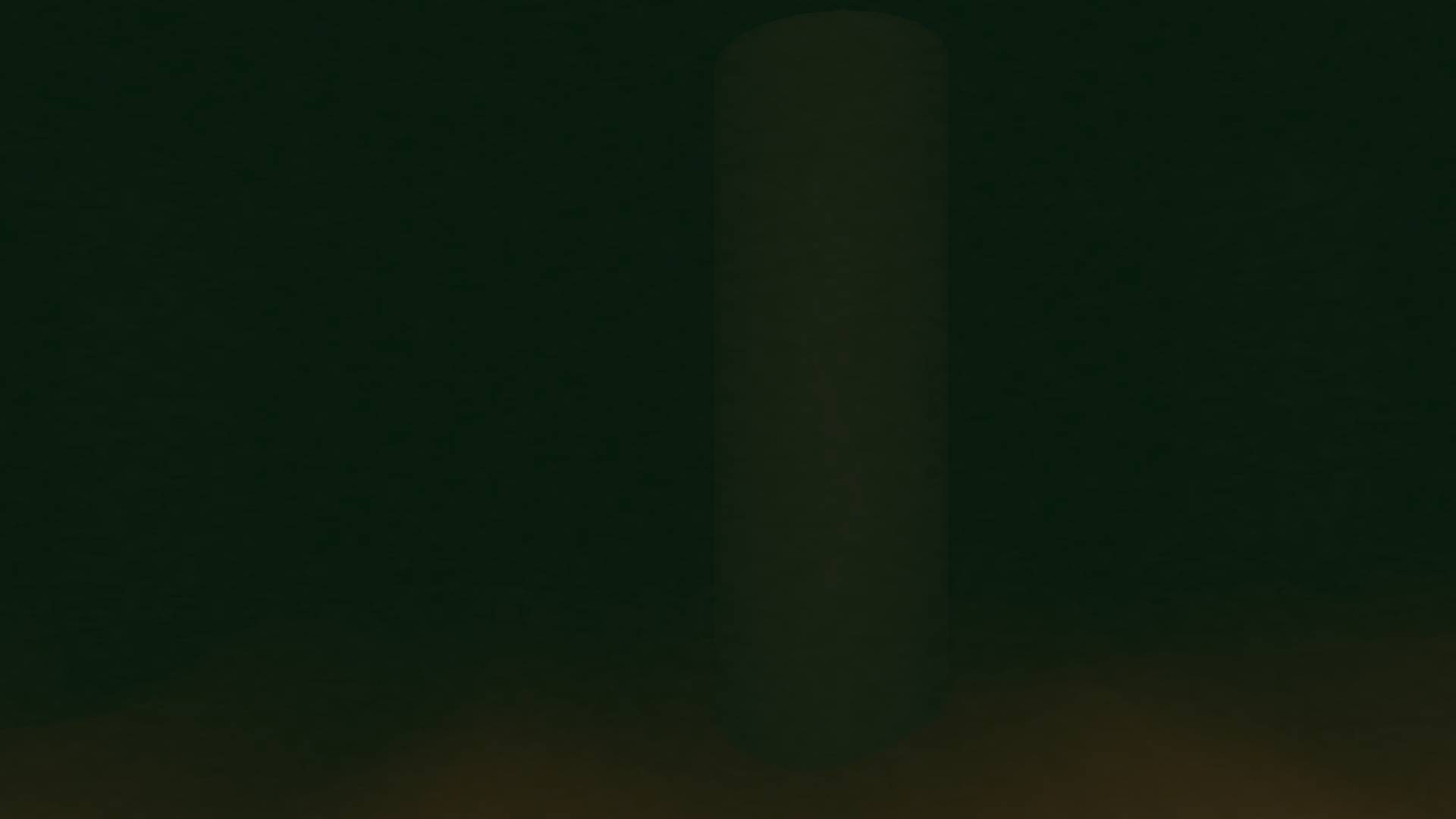}
                \par{\footnotesize Input Scene B\par}
            \end{minipage}\hspace{0.8mm}%
            \begin{minipage}[b]{0.32\linewidth}
                \centering
                \includegraphics[width=\linewidth]{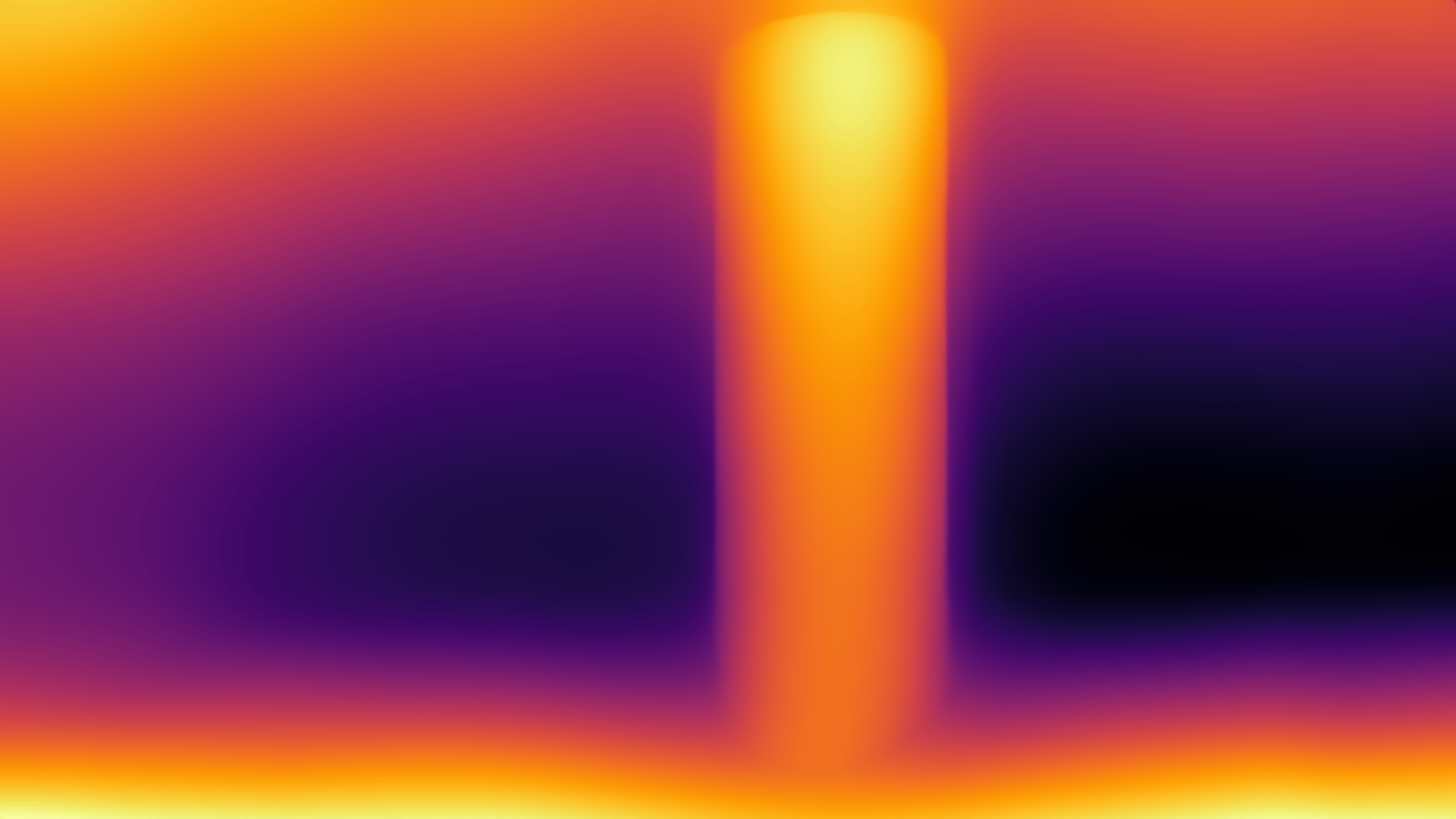}
                \par{\footnotesize DUViN-air\par}
            \end{minipage}\hspace{0.8mm}%
            \begin{minipage}[b]{0.32\linewidth}
                \centering
                \includegraphics[width=\linewidth]{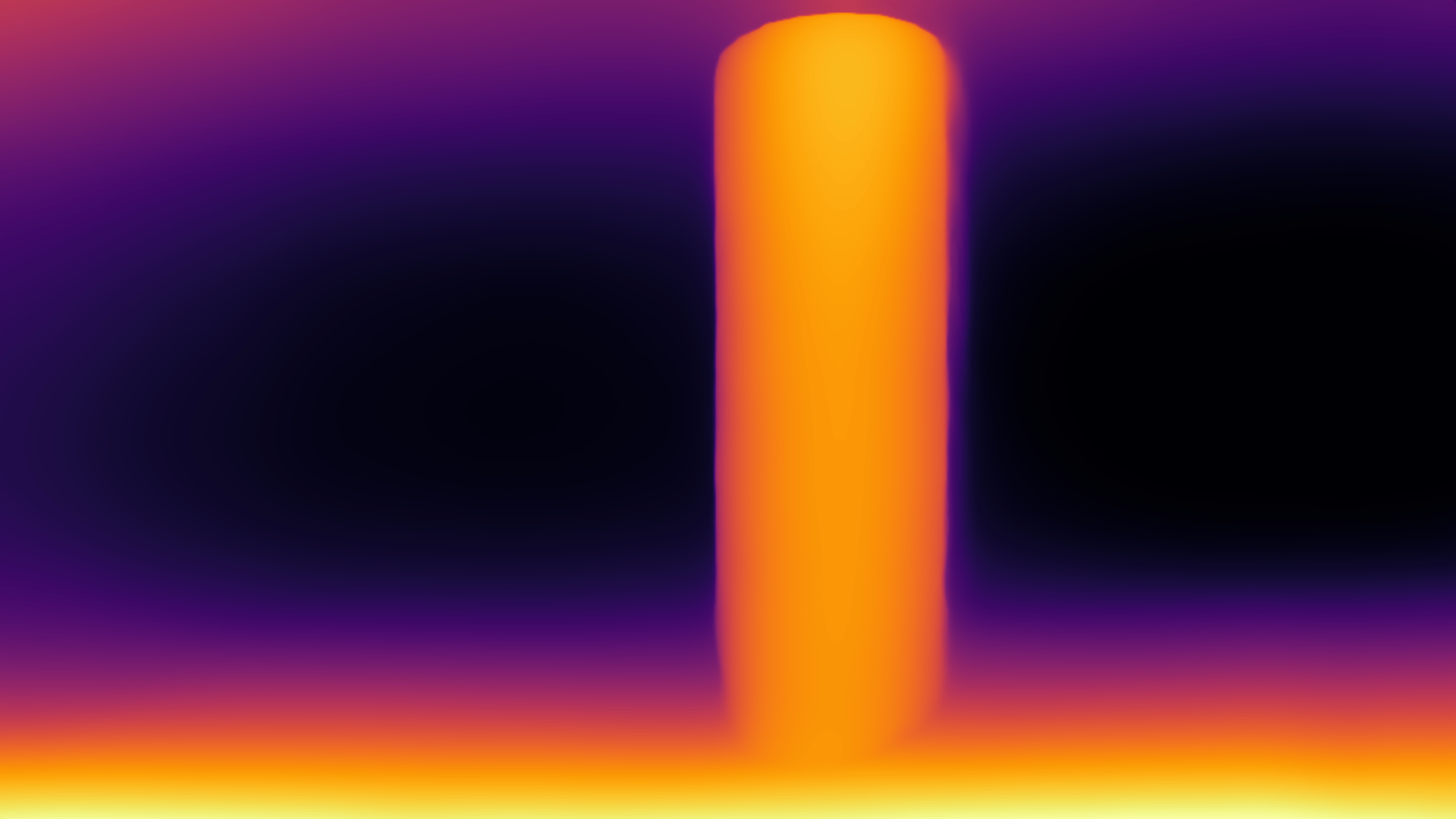}
                \par{\footnotesize DUViN\par}
            \end{minipage}\hspace{0.8mm}%
            \par
            \begin{minipage}[b]{0.21\linewidth}
                \centering
                \includegraphics[width=\linewidth]{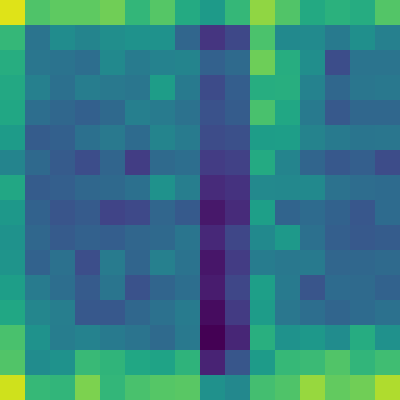}
                \par{\footnotesize DUViN(a) $\mathbb{F}_t^c$\par}
            \end{minipage}\hspace{0.8mm}%
            \begin{minipage}[b]{0.21\linewidth}
                \centering
                \includegraphics[width=\linewidth]{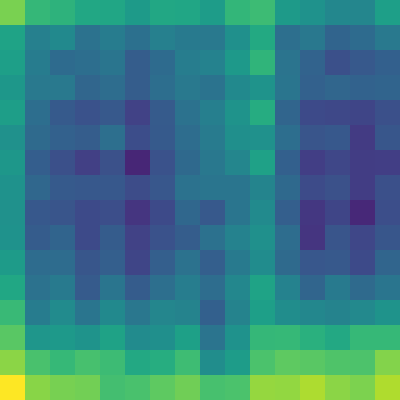}
                \par{\footnotesize DUViN $\mathbb{F}_t^{c}$\par}
            \end{minipage}\hspace{0.8mm}%
            \begin{minipage}[b]{0.21\linewidth}
                \centering
                \includegraphics[width=\linewidth]{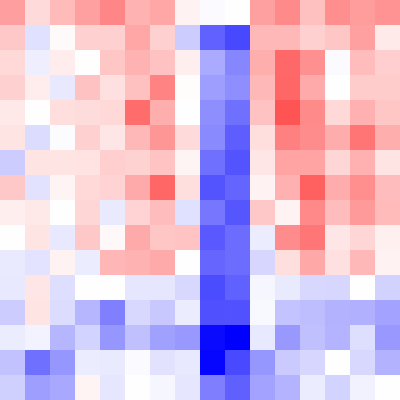}
                \par{\footnotesize Feature diff.\par}
            \end{minipage}\hspace{0.8mm}%
            \begin{minipage}[b]{0.32\linewidth}
                \centering
                \includegraphics[width=\linewidth]{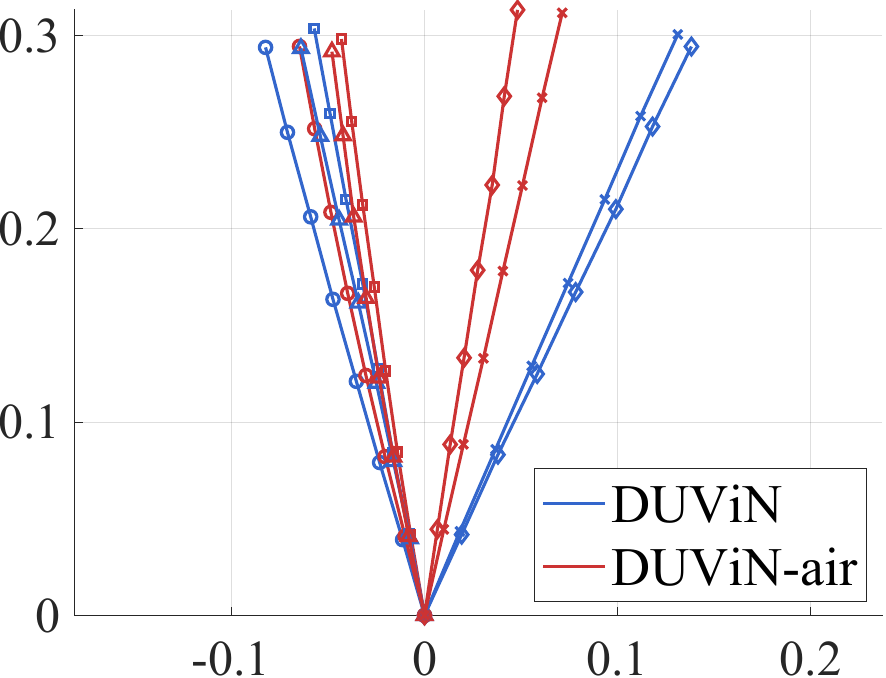}
                \par{\footnotesize Actions\par}
            \end{minipage}\hspace{0.8mm}%
        \end{minipage}%
    }
    \caption{\textbf{Feature-level comparison between DUViN and DUViN-air in two scenes (A and B).} Depth estimation results based on the DUViN depth extractor show improved performance in extremely turbid and dark underwater environments, resulting in more reliable feature extraction of $\mathbb{F}_t^{c}$ than DUViN-air, noted as DUViN(a). The feature difference map highlights differences in blue (closer estimation by DUViN) and red (further). This suggests DUViN provides a more timely and accurate response to approaching obstacles.}
    \label{fig:duvin_failure}
\end{figure}

In contrast, both DUViN-air and DUViN demonstrate significantly more robust performance in underwater environments, primarily due to the incorporation of abstract depth information. Notably, DUViN-air can successfully complete navigation tasks across all three water types, although it suffers more pronounced performance degradation compared to DUViN, which benefits from the transferred depth encoder. Furthermore, DUViN exhibits stronger robustness to increased turbidity, maintaining a 55\% collision-free rate in collision-free navigation even under severe visibility degradation. Nevertheless, the overall performance still degrades under more challenging water conditions. In particular, under Type 7C, the visibility range is reduced to only 1.45 m, which makes navigation extremely difficult in dense obstacle environments. Moreover, the simulator incorporates the AUV dynamics model described in \cref{auv_dynamics}, including hydrodynamic effects and actuation constraints, which further limit the vehicle’s ability to react quickly and accurately in such conditions, especially when the available reaction time is significantly reduced by severe turbidity.

Additionally, we further validate the effectiveness and necessity of the transferred depth feature extractor in DUViN by evaluating its performance in a 4-DoF terrain-following task with altitude maintenance. In this experiment, we demonstrate full 4-DoF control of the AUV in both randomly generated \textbf{Hills} and \textbf{Pillars} environments, where the terrain altitude varies continuously. We compare DUViN-air, which uses the original in-air depth feature extractor, with the full DUViN model incorporating the transferred depth encoder.

Beyond the obstacle collision evaluation presented in the previous section, we further assess the AUV's ability to maintain a safe and effective altitude relative to the terrain. Specifically, the acceptable altitude is limited to 0.5 meters above a predefined reference altitude, with an additional requirement that the clearance from the seabed remains above 0.1 meters. Any deviation beyond this range is counted as a collision. For goal-reaching evaluation, the AUV is considered successful only if it arrives within a 1-meter radius of the designated target point. Each method is evaluated over 20 independent trials in each environment. The results are summarized in \cref{simualte_transfer}. With varying terrain, the AUV encounters more collisions compared to flat environments where it navigates at a constant altitude. However, DUViN, equipped with the transferred encoder, demonstrates a more robust and safer navigation policy than DUViN-air in the two different simulation environments, which employs an in-air depth encoder. DUViN achieves a higher success rate, a higher collision-free rate, and fewer collision instances overall in the three different water types. These results highlight DUViN's transferability across different layouts and water conditions, enabled by the encoder's pretraining on a large-scale in-air dataset, the effective use of abstract depth information for navigation and the transferred depth feature extractor. As a result, DUViN with the transferred encoder performs more reliably under extreme turbidity in complex underwater conditions.

To intuitively illustrate the reasons behind the observed results and how the transferred depth encoder contributes to performance, we analyze representative test cases. Specifically, we compare the outputs of DUViN and DUViN-air in one of the \textbf{Pillars} environment under water type 7C (most challenging case), where DUViN completes the navigation successfully while DUViN-air results in a collision. The results are presented in \cref{fig:duvin_failure_a}. As shown, the robot equipped with DUViN not only avoids all obstacles but also follows a smoother trajectory with fewer orientation adjustments under identical environmental conditions. To further investigate the feature-level differences, we extract and analyze the images at scenes A and B, as shown in \cref{fig:scene_a_analysis,fig:scene_b_analysis}. Here we visualize the intermediate feature $\mathbb{F}_t^c$ after the feature compressor module, highlighting the representations extracted by the in-air pre-trained depth feature extractor and the transferred feature extractor adapted to underwater physics. The results clearly demonstrate that $\mathbb{F}_t^c$ from DUViN can successfully detect obstacles, whereas DUViN-air fails to do so. These regions appear black in the feature visualization and are shown in blue on the feature difference map, indicating that DUViN-air estimates obstacles as distant. Such incorrect feature extraction adversely affects subsequent navigation decisions. In Scene A, DUViN performs a smooth avoidance maneuver, while DUViN-air exhibits a jittery response due to inaccurate depth estimation in the turbid environment. Similarly, in Scene B, DUViN initiates a timely obstacle avoidance, indicating better obstacle awareness.

\subsubsection{Feature Normalization Ablation}

Changes in the observed scenes and water conditions can cause shifts in the extracted depth feature space. The feature normalization is introduced to reduce these variations and provide more consistent visual features to the diffusion navigation policy. To evaluate its necessity, we compare DUViN with \textbf{DUViN-wo-FN}, a variant retrained after removing the feature normalization in \cref{fn-1,fn-2}. In DUViN-wo-FN, the extracted depth features are directly passed to the subsequent feature-fusion module. Both models use the transferred encoder and are trained using the same settings. The evaluation is conducted at fixed altitude in the \textbf{Pillars} under the Type 3C water condition.

\begin{table}[b]
\centering
\caption{Ablation of Feature Normalization Under the Type 3C Water Condition in the Pillars Environment}
\label{tab:feature_norm_ablation}
\renewcommand{\arraystretch}{1.2}
\setlength{\tabcolsep}{12pt}
\begin{tabular}{l|ccc}
\toprule
\textbf{Model} &
\textbf{Succ.\%} $\uparrow$ &
\textbf{CF\%} $\uparrow$ &
\textbf{A.\ C.} $\downarrow$ \\
\midrule
DUViN-wo-FN &
85 &
10 &
2.40 \\
\textbf{DUViN} &
\textbf{100} &
\textbf{60} &
\textbf{0.50} \\
\bottomrule
\end{tabular}
\end{table}

As shown in \cref{tab:feature_norm_ablation}, removing feature normalization reduces the success rate from 100\% to 85\% and the collision-free rate from 60\% to 10\%, while increasing the average number of collisions from 0.5 to 2.4. These results demonstrate the contribution of feature normalization under the underwater domain shift and support its role in maintaining consistency in the extracted feature space across different scenes and water conditions.

\subsubsection{Altitude Maintenance Analysis}
To evaluate the effectiveness of altitude maintenance, we conduct two experiments: terrain following and altitude regulation under external disturbances. We compare two models: \textbf{DUViN}, which incorporates the transferred underwater encoder; and \textbf{DUViN-woA}, an ablated variant trained without the altitude maintenance Z-axis planning method described in \cref{MPC2}.

\begin{figure}[t]
    \centering
    \captionsetup[subfigure]{farskip=0pt}
    \subfloat[Terrain-following performance\label{fig:terrain_following_a}]{%
        \parbox{\linewidth}{%
            \centering
            \includegraphics[width=0.99\linewidth]{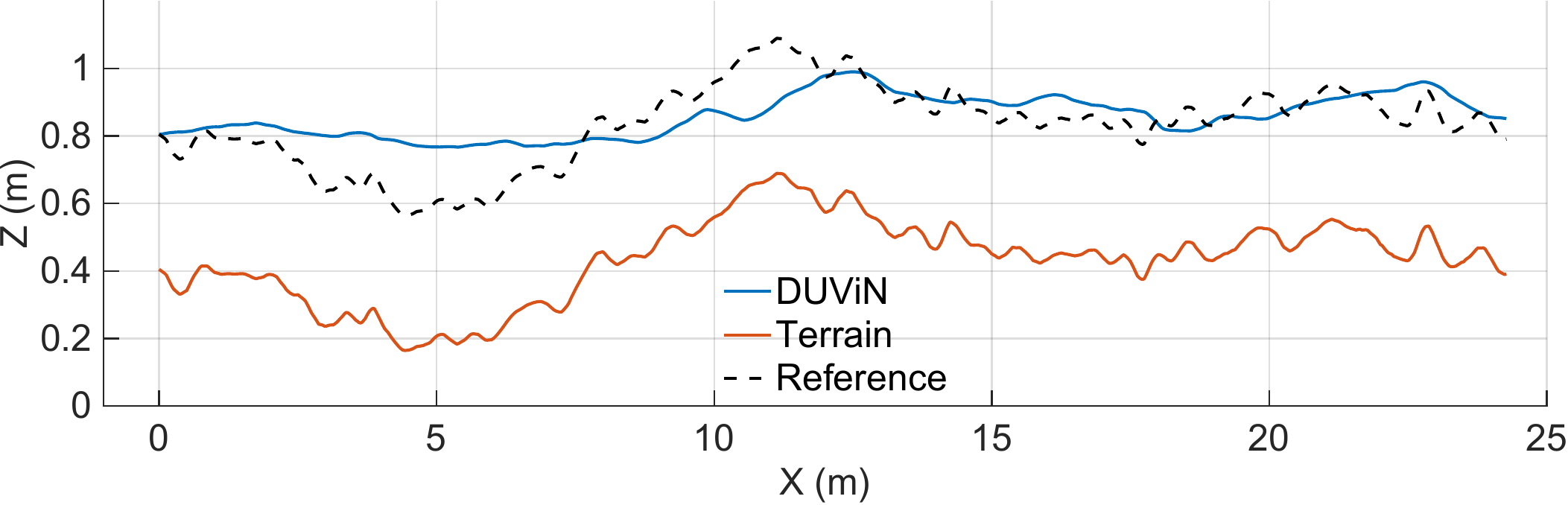}
            \par\nointerlineskip
            \includegraphics[width=0.99\linewidth]{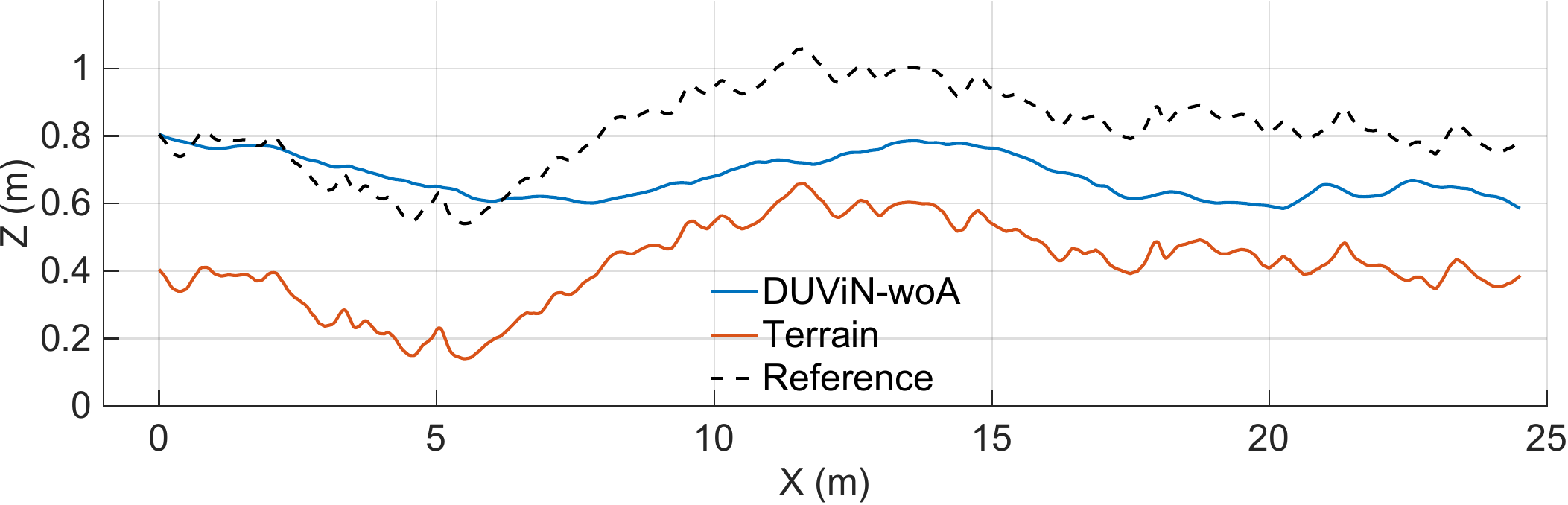}
        }%
    }
    \par
    \subfloat[Disturbance rejection performance\label{fig:terrain_following_b}]{%
        \includegraphics[width=0.99\linewidth]{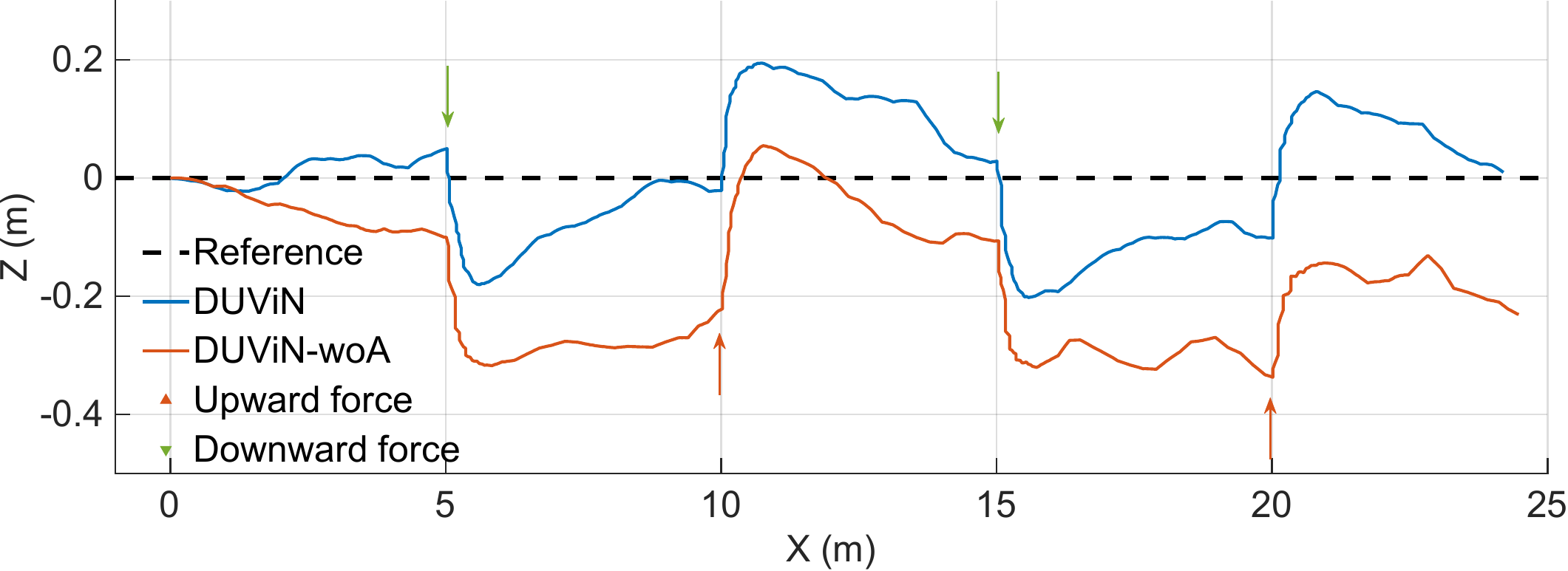}%
    }
    \caption{\textbf{Effectiveness of altitude maintenance.} DUViN, trained with altitude maintenance $Z$-axis planning, can maintain altitude relative to the initial altitude and effectively reject external disturbances. In contrast, DUViN-woA follows terrain variations but drifts over time and fails to maintain the altitude under disturbances.}
    \label{fig:terrain_following}
\end{figure}

An example of terrain-following performance is illustrated in \cref{fig:terrain_following_a}, where two models are evaluated in the \textbf{Pillars} environment. As shown, DUViN successfully navigates the robot along the terrain while maintaining a consistent altitude relative to the initial altitude. In contrast, DUViN-woA, which is trained without replanning on the Z axis, follows the general trend of the terrain, but suffers from altitude drift over time and cannot maintain a stable altitude.

Meanwhile, the altitude maintenance performance under external disturbances is illustrated in \cref{fig:terrain_following_b}. In this scenario, vertical forces are applied to the robot every time it moves forward by 5 meters. The robot equipped with DUViN demonstrates a clear recovery trend after disturbances, whereas the DUViN-woA variant exhibits a step-like response with drift. These results validate the effectiveness of our approach in maintaining altitude by leveraging the reference height, and further highlight the necessity of incorporating Z-axis planning during the training process.

\subsubsection{Navigation Under Wave Disturbances}
To evaluate the effect of wave disturbances on DUViN, we conduct an additional simulation experiment in which a sinusoidal flow disturbance in the horizontal plane, simplified from the model in~\cite{siriya2025toward}, with a peak velocity of 0.2~m/s and a frequency of 0.2~Hz, is introduced into the AUV dynamics. The disturbance cannot be fully rejected by the controller. We conduct three trials with and without the disturbance under the same settings.

\begin{figure}[b]
    \centering
    \includegraphics[width=0.99\linewidth]{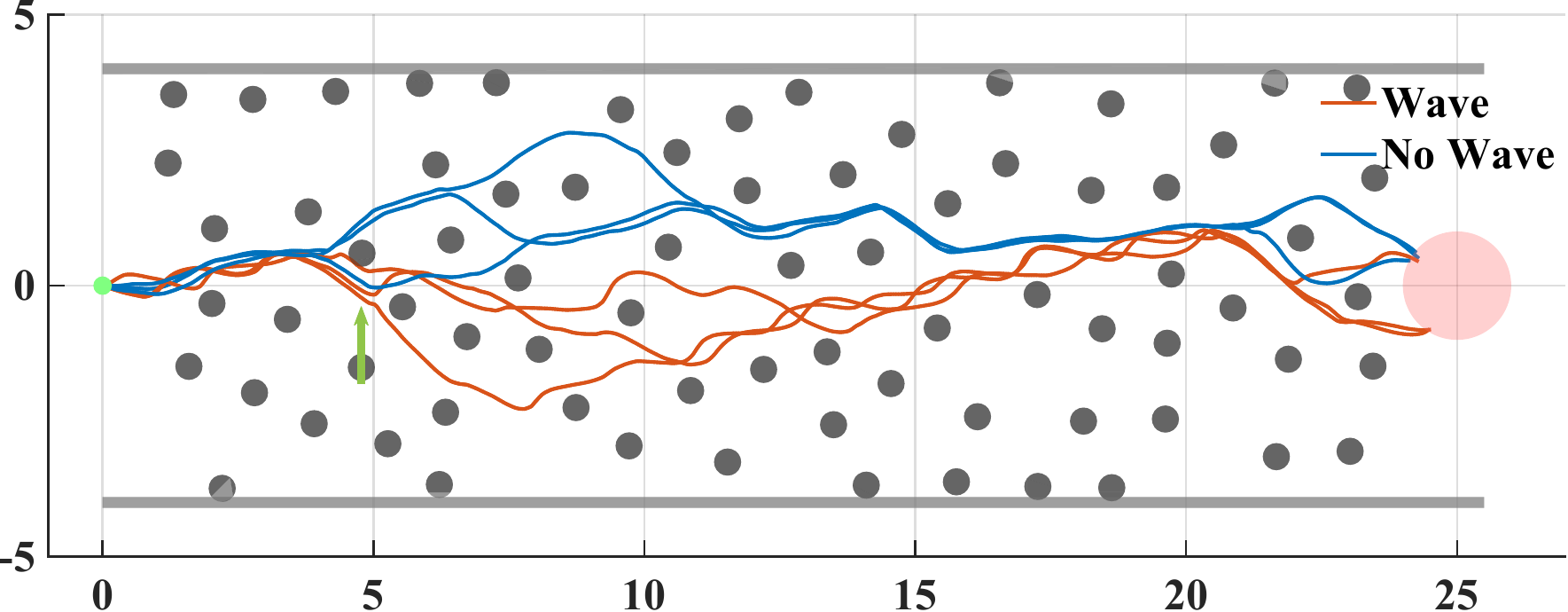}
    \caption{\textbf{DUViN performance with and without wave disturbances.} The red trajectories represent navigation under wave disturbances, whereas the blue trajectories represent navigation without wave disturbances. Under wave disturbances, DUViN produces sharper reactions near obstacles because the AUV is pushed away from its planned motion.}
    \label{fig:wave_disturbance}
\end{figure}

As shown in \cref{fig:wave_disturbance}, DUViN produces smoother trajectories and reaches the destination more directly without wave disturbances. By contrast, under wave disturbances, the AUV can be pushed away from its planned trajectory, and DUViN therefore has to react more sharply and make more frequent corrections based on the updated visual observations, especially near obstacles, as indicated by the green arrow in \cref{fig:wave_disturbance}. Although the disturbance increases the collision risk, the policy can readjust its motion through subsequent visual replanning. Overall, these results demonstrate DUViN's ability to make timely motion adjustments in response to external disturbances.

\subsubsection{Goal-Aware Feature Fusion Analysis}
\begin{figure}[t]
    \centering
    \captionsetup[subfigure]{farskip=0pt}
    \subfloat[Input scene]{%
        \includegraphics[width=0.22\textwidth]{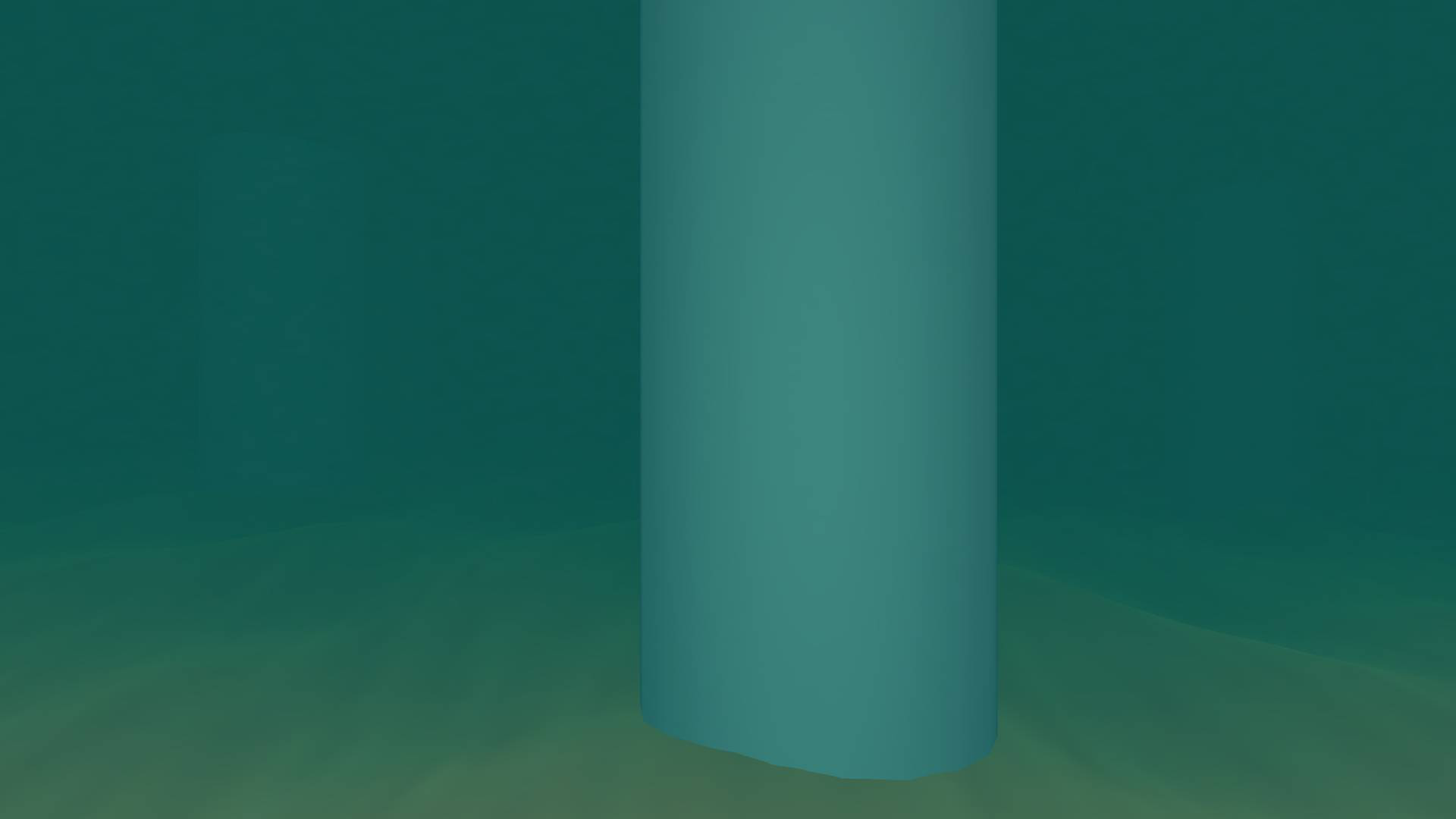}%
    }%
    \hspace{0.8mm}%
    \subfloat[Effect of $d$\label{fig:goal_aware_sim_d}]{%
        \includegraphics[width=0.22\textwidth]{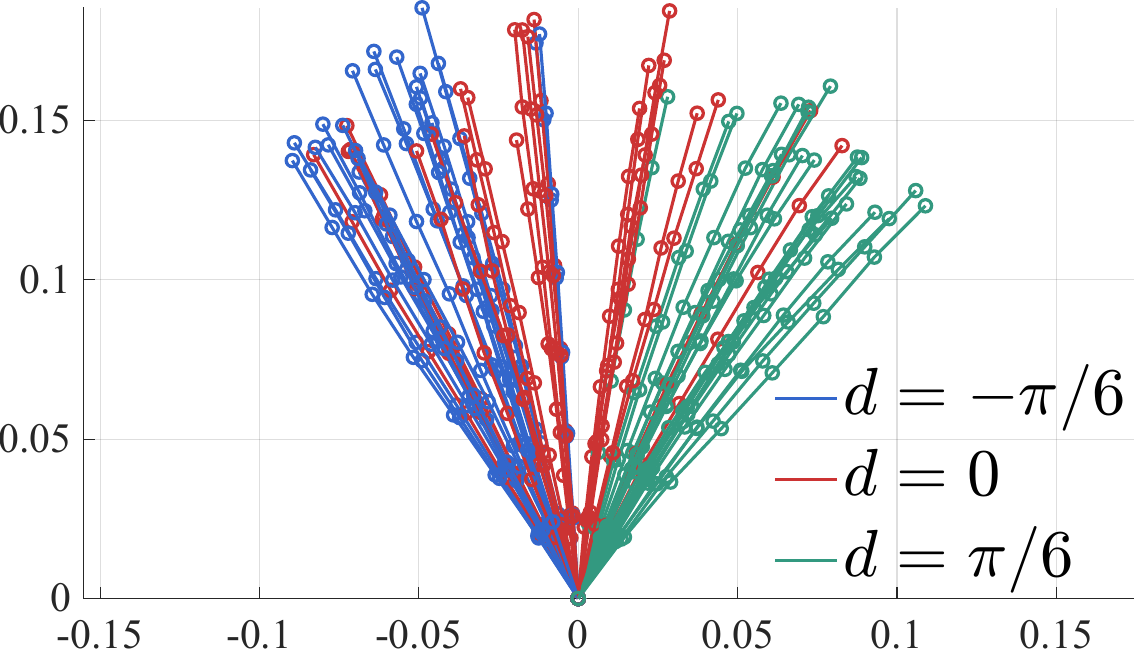}%
    }
    \par
    \subfloat[Effect of $s$\label{fig:goal_aware_sim_c}]{%
        \includegraphics[width=0.48\textwidth]{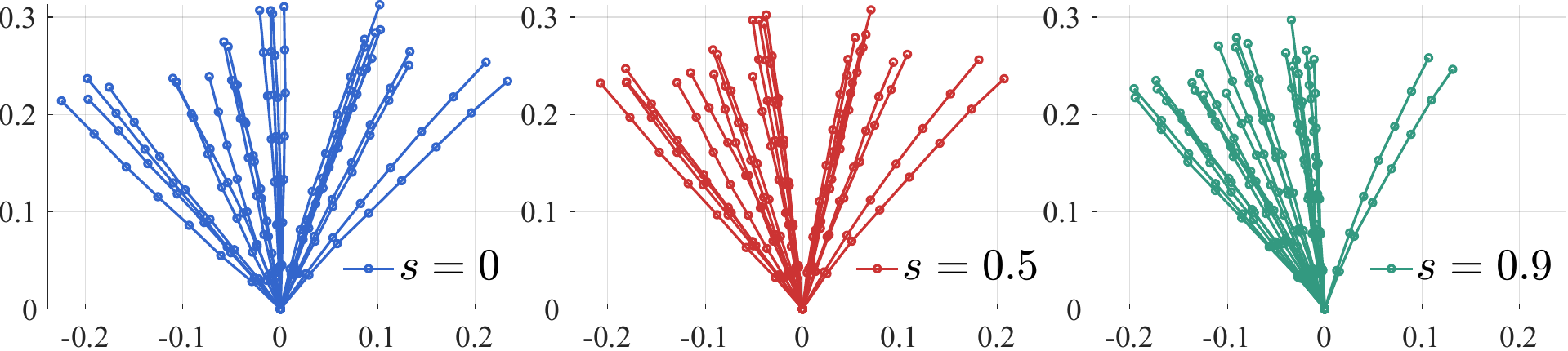}%
    }
    \caption{\textbf{Effectiveness of goal awareness.} Given an input scene with multiple feasible paths, the goal direction $d$ significantly influences the policy’s decision-making, guiding the agent toward the intended direction. Additionally, the distance signal $s$ affects the policy preference, where closer to the goal encourages safer, less exploratory paths.}
    \label{fig:goal_aware_sim}
\end{figure}

To assess the influence of goal conditioning on the diffusion-based navigation policy, we simulate a scene with two symmetric obstacle avoidance paths (left and right) and vary the goal parameters: direction~$d$ and relative distance~$s$. By fixing the random noise input, the resulting denoised actions are depicted in \cref{fig:goal_aware_sim}. In particular, when $d=0^\circ$ and $s=0.5$, the policy yields a multi-modal set of trajectories (see \cref{fig:goal_aware_sim_d}), reflecting unbiased exploration. Introducing a leftward ($d=-30^\circ$) or rightward ($d=+30^\circ$) goal bias skews the generated trajectories toward the specified direction, demonstrating effective goal awareness.

We further analyze the effect of varying $s$, which interpolates the robot’s proximity to the goal (see \cref{fig:goal_aware_sim_c}). At small $s$ (near the start), the trajectory distribution remains broad and multi-modal, favouring exploration. As $s$ increases (approaching the goal), the mode becomes more conservative: actions decisively choose the safer, more direct avoidance route. This progression from exploratory to exploitative behaviour demonstrates that the policy dynamically balances obstacle avoidance and goal approaching as the robot nears its goal.

\subsubsection{Comparison With Direct Underwater-Simulator Training}
Direct comparison with underwater-specific methods such as UIVNav~\cite{lin2024uivnav} or classical sonar-based planners is not straightforward, because they rely on different sensing assumptions, task formulations, and action spaces. Therefore, to provide a more closely matched baseline, we further compare DUViN with variants trained directly in our underwater simulator under the same training pipeline, which further substantiates the effectiveness and necessity of our method.

With the aid of underwater imaging formation models, current underwater simulators can partially reduce the visual gap between clear image simulation and real-world underwater environments. However, a vision-based navigation policy trained directly in underwater simulation still faces challenges in generalization across underwater environments. Underwater environments exhibit large visual variations across water types, including differences in appearance, turbidity levels, and backscattering effects. Therefore, a policy trained in one simulated underwater condition may improve performance in that condition, but still generalizes poorly to different underwater conditions.

To examine this point, we trained additional variants directly in our underwater simulation environments under two different water types, Type 1C and Type 7C, using the same Phase 1 policy training pipeline and the original in-air depth feature extractor, but without Phase 2 feature adaptation. These two variants are denoted as DUViN-1C and DUViN-7C, respectively. However, neither direct underwater-training variant outperformed the proposed two-phase DUViN, as shown in \cref{simulate_inair}.

\begin{table}[b]
\centering
\caption{Obstacle Avoidance Performance of Policies Trained Under Different Simulation Conditions. \textbf{\textcolor{ForestGreen}{Bold green}} indicates best, \textcolor{ForestGreen}{green} indicates second best}
\vspace{0.5em}
\resizebox{\columnwidth}{!}{%
\begin{tabular}{|c|ccc|ccc|}
\hline
\textbf{Water Types}  & \multicolumn{3}{c|}{Type 1C} & \multicolumn{3}{c|}{Type 7C} \\
\hline
\textbf{Metrics} & Succ.\% & CF\% & A.\ C. & Succ.\% & CF\% & A.\ C. \\
\hline
\textbf{DUViN-1C} & \textcolor{ForestGreen}{\textbf{100}} & \textcolor{ForestGreen}{55} & \textcolor{ForestGreen}{0.55} & 90 & 15 & 1.40 \\
\hline
\textbf{DUViN-7C} & \textcolor{ForestGreen}{\textbf{100}} & 40 & 0.90 & \textcolor{ForestGreen}{\textbf{100}} & \textcolor{ForestGreen}{40} & \textcolor{ForestGreen}{0.75} \\
\hline
\textbf{DUViN-air} & \textcolor{ForestGreen}{\textbf{100}} & 50 & 0.60 & \textcolor{ForestGreen}{95} & 25 & 1.20 \\
\hline
\textbf{DUViN} & \textcolor{ForestGreen}{\textbf{100}} & \textcolor{ForestGreen}{\textbf{60}} & \textcolor{ForestGreen}{\textbf{0.45}} & \textcolor{ForestGreen}{\textbf{100}} & \textcolor{ForestGreen}{\textbf{55}} & \textcolor{ForestGreen}{\textbf{0.65}} \\
\hline
\end{tabular}%
}
\label{simulate_inair}
\end{table}

These results show that DUViN-1C and DUViN-7C achieve improved performance in their corresponding training water types, but perform less effectively in the other water type. This is mainly due to the large visual differences between different underwater environments. In contrast, DUViN achieves the best overall performance across both water types, and both DUViN-1C and DUViN-7C remain inferior to DUViN. One possible reason is that the in-air depth estimation model still suffers from degraded and distorted feature extraction in underwater imagery before adaptation, which can negatively affect policy learning when the navigation model is trained directly on such features. By comparison, DUViN benefits from the proposed two-phase design, which preserves effective policy learning in Phase 1 and further improves robustness through physics-informed feature adaptation in Phase 2.

%% file: mainBody/4_2_Experiments.tex
\subsection{Real-World Experiments}
\label{sec:real}
\subsubsection{Real-World Experimental Setup}

\begin{figure}[t]
\centering
\includegraphics[width=0.99\linewidth]{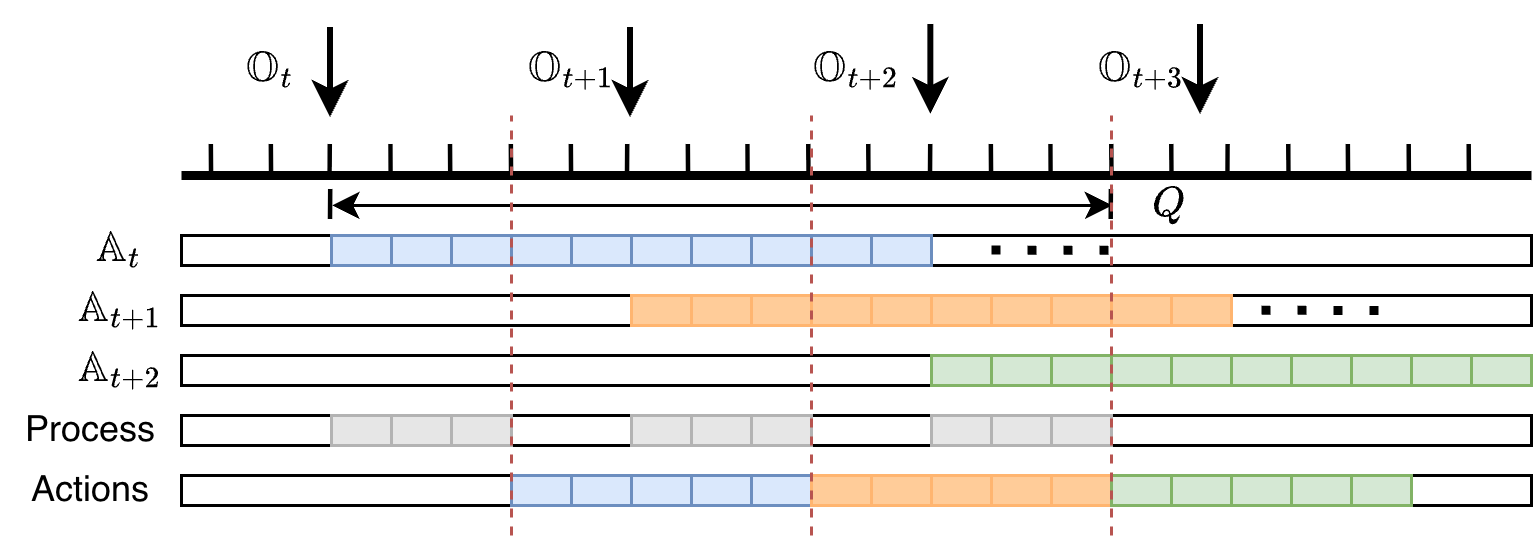}
\caption{\textbf{Asynchronous visual inference and control execution.} Each $\mathbb{A}_t$ is generated after a delay and reused across multiple control cycles. \textit{Actions} denote the actual control signals executed by the AUV.}
\label{fig:control_timeline}
\end{figure}

We evaluate DUViN on a BlueROV2 Heavy Configuration platform within an indoor water tank. The vehicle is equipped with eight T200 thrusters, allowing full actuation across all 6 DoF. For perception, an onboard forward-facing Low-Light HD USB monocular camera provides $1920\times1080$ resolution imagery at 15 FPS, calibrated specifically for underwater operation. The calibrated images are rectified to match a horizontal field of view (FOV) of $86^\circ$, which aligns with the onboard camera specifications in the training data collected in Unity. The camera streams are transmitted in real time to an external computing server equipped with an NVIDIA A40 GPU for policy inference, where DUViN performs visual inference at a rate of 2 Hz. The generated 4-DoF motion commands are then sent back to the robot through MAVLink for onboard execution. The motion commands are executed with actuator voltage magnitudes scaled according to the command. Meanwhile, the onboard control loop operates at a higher frequency of 10~Hz to maintain smooth actuation. To address the delay introduced by image acquisition and inference, which can cause misalignment between the perception stream and control execution, we implement an asynchronous control scheduler. Fig.~\ref{fig:control_timeline} illustrates the asynchronous control scheduling pipeline designed to compensate for delays in image processing and action generation. Each incoming observation $\mathbb{O}_t$ undergoes a ``Process" period before generating the corresponding action sequence $\mathbb{A}_t$, which spans a fixed length $Q$ and is conditioned on the current observation. During this processing delay, the control module continues to operate by executing the most recently available action sequence from the previous observation.

\begin{figure}[tp]
    \centering
    \subfloat[Experimental setup\label{fig:setup}]{%
        \includegraphics[width=0.9\linewidth]{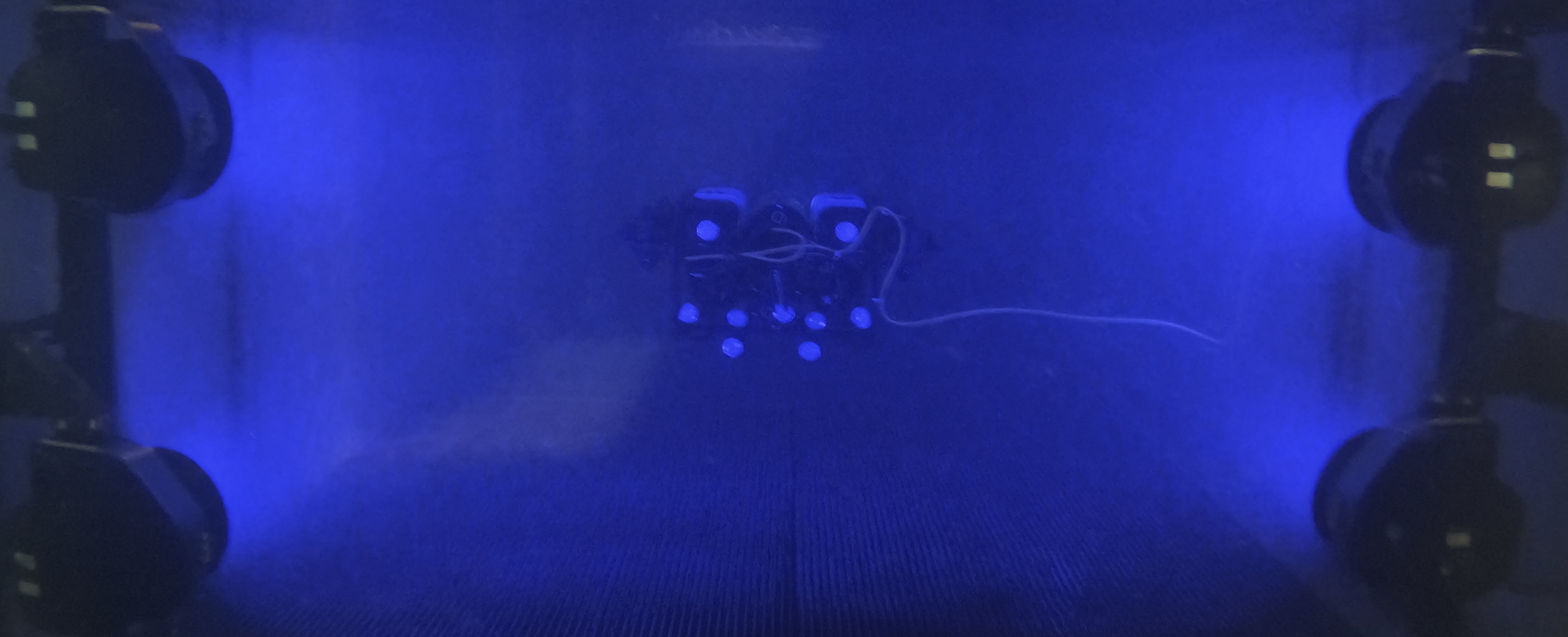}%
    }
    \par\vspace{-3mm}
    \subfloat[DUViN navigating an AUV in the real world\label{fig:title}]{%
        \includegraphics[width=0.9\linewidth]{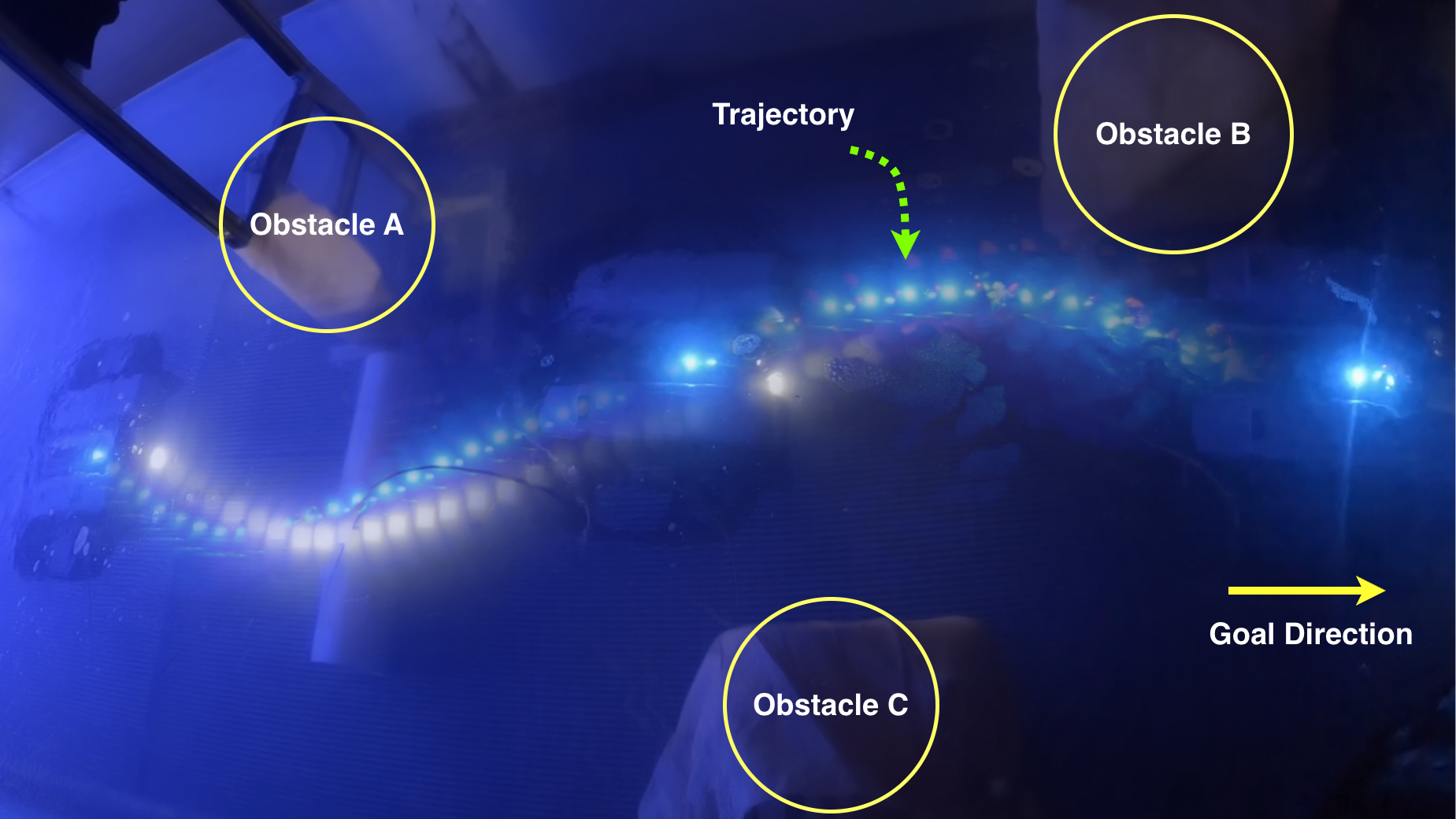}%
    }
    \caption{\textbf{Real-world DUViN deployment and experimental setup.} 
    (a) Four Qualisys tracking cameras are installed in the water tank to capture the tracking markers attached to the AUV for accurate motion tracking. 
    (b) An AUV navigates underwater using DUViN. The white and blue lights on the robot correspond to onboard illumination and system status indicators, respectively. The surrounding blue hue is caused by active lighting from the underwater motion tracking system.}
    \label{fig:realworld_overview}
\end{figure}

Groundtruth pose data is acquired via a Qualisys underwater motion capture system, comprising four optical cameras mounted at the corners of the operational volume on one side. These cameras emit active blue lighting and detect reflected light from tracking markers to accurately record the vehicle's trajectory. The evaluation is conducted in an indoor long water tank facility measuring $1.8$ meters in width, and $1$ meter in depth. However, the effective operational area is constrained to a $3 \times 1.8 \times 1$ meters volume, corresponding to the calibrated workspace of the groundtruth tracking system. The reduced effective workspace is primarily due to the attenuation of optical signals in water, which limits the reliable capture volume of the tracking system. The motion capture system is used only for trajectory recording and quantitative evaluation, and is not part of the online perception or control loop of DUViN.

The real-world environment is designed to approximate real underwater conditions. Both optical backscattering and attenuation effects are present, and water turbidity also appears, reducing visibility. Ambient lighting is minimized to simulate low-light subsea scenarios. Illumination is provided solely by the tracking system’s integrated blue lights and the vehicle’s onboard lights. These conditions provide a realistic yet repeatable environment for validating the algorithm under perceptually degraded conditions. Fig.~\ref{fig:realworld_overview} shows the real-world deployment of DUViN. Specifically, Fig.~\ref{fig:setup} presents the experimental setup, while Fig.~\ref{fig:title} shows DUViN’s real-world navigation performance on the BlueROV2 platform.

To evaluate the performance and effectiveness of the proposed algorithm, three experimental scenarios are designed and conducted: obstacle avoidance, altitude maintenance, and goal awareness.

\begin{figure}[t]
    \centering
    \subfloat[Feature-level analysis of real-world Scene A]{%
        \begin{minipage}[t]{0.48\textwidth}
            \centering
            \begin{minipage}[b]{0.32\linewidth}\centering
                \includegraphics[width=\linewidth]{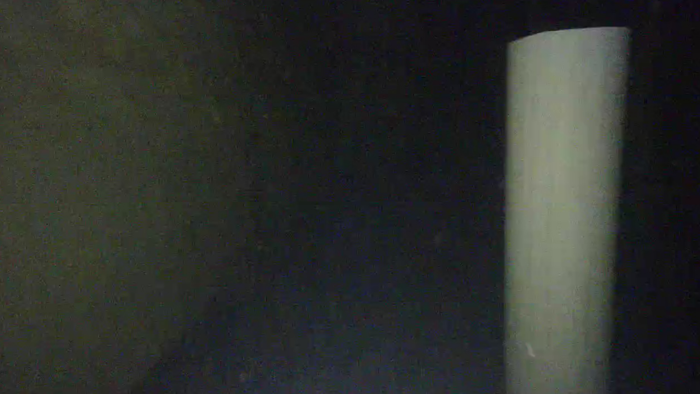}
                \par{\footnotesize Input\par}
            \end{minipage}\hspace{0.8mm}%
            \begin{minipage}[b]{0.32\linewidth}\centering
                \includegraphics[width=\linewidth]{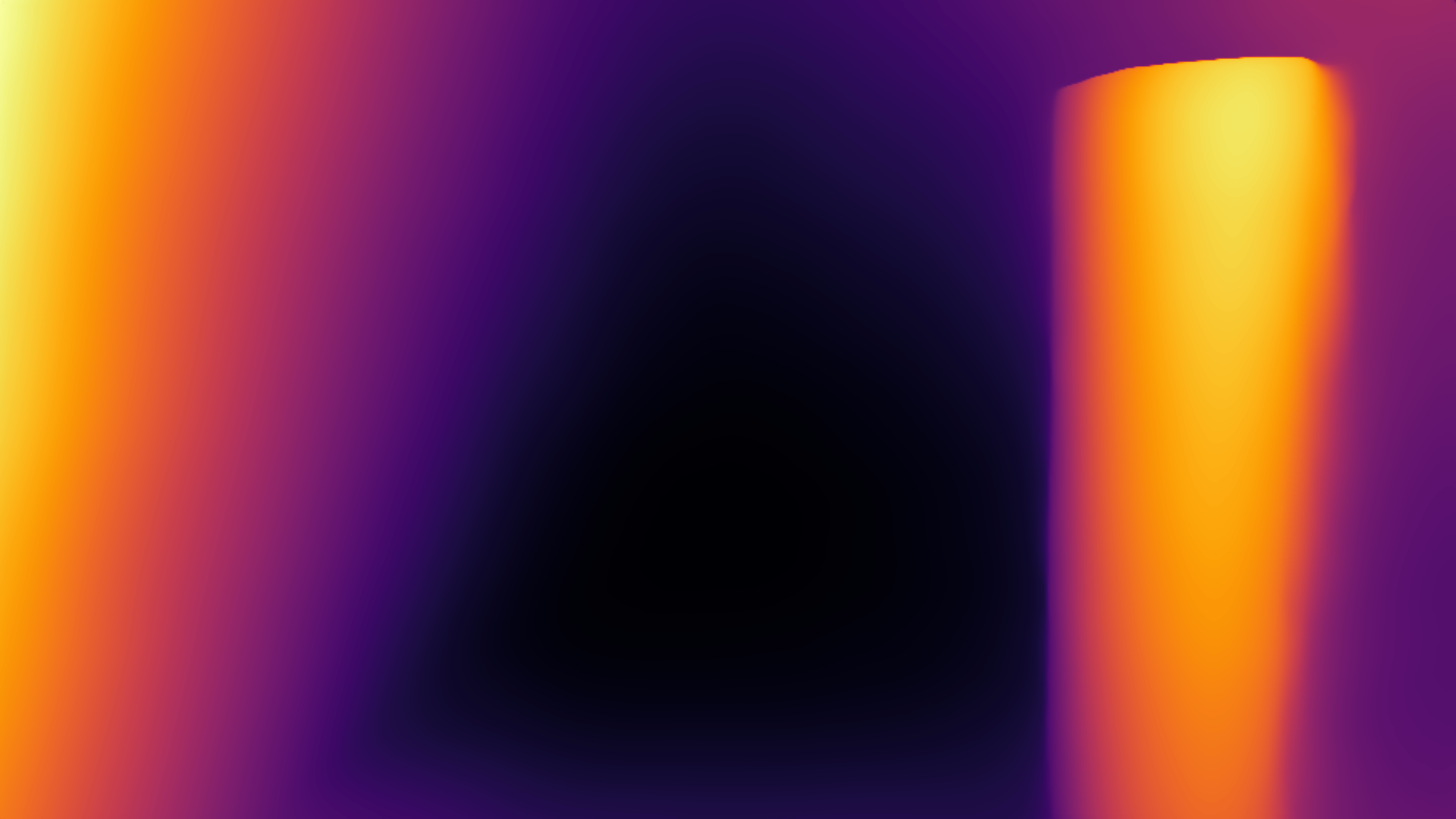}
                \par{\footnotesize DUViN-air\par}
            \end{minipage}\hspace{0.8mm}%
            \begin{minipage}[b]{0.32\linewidth}\centering
                \includegraphics[width=\linewidth]{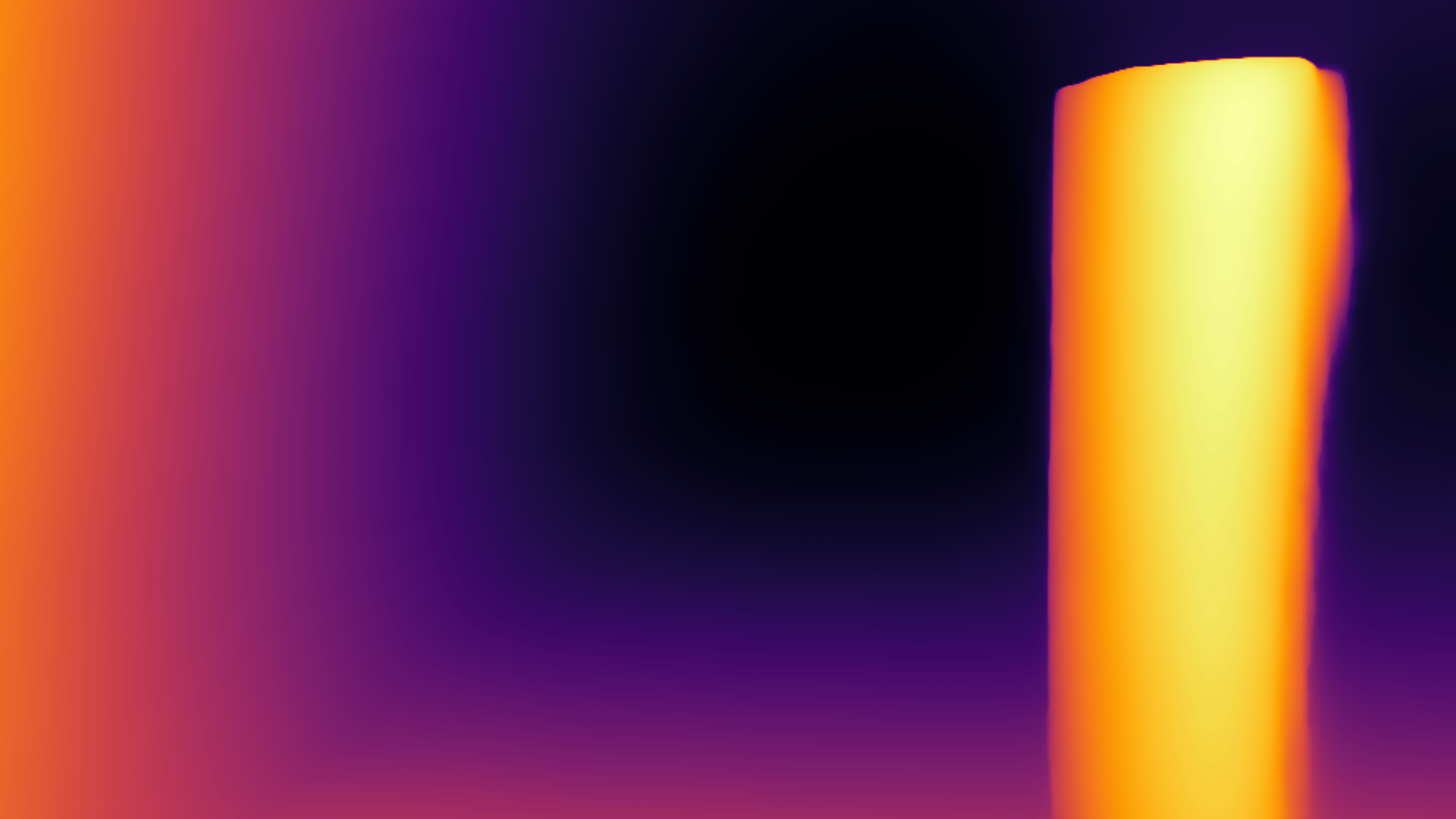}
                \par{\footnotesize DUViN\par}
            \end{minipage}
            \par\vspace{1mm}
            \begin{minipage}[b]{0.19\linewidth}\centering
                \includegraphics[width=\linewidth]{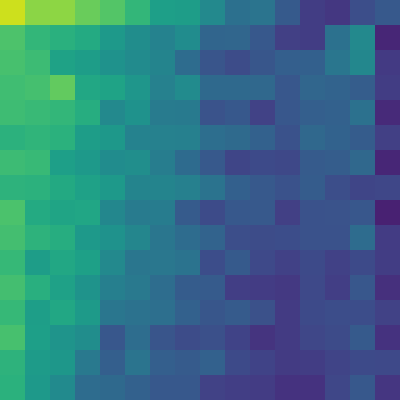}
                \par{\footnotesize DUViN(a) $\mathbb{F}_t^c$\par}
            \end{minipage}\hspace{0.8mm}%
            \begin{minipage}[b]{0.19\linewidth}\centering
                \includegraphics[width=\linewidth]{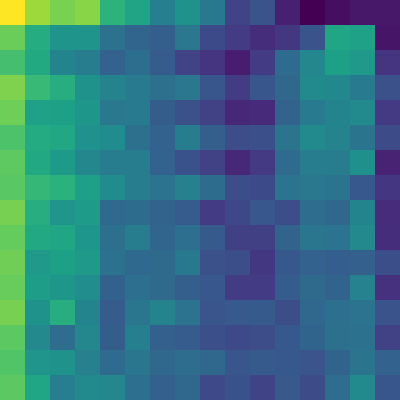}
                \par{\footnotesize DUViN $\mathbb{F}_t^{c}$\par}
            \end{minipage}\hspace{0.8mm}%
            \begin{minipage}[b]{0.19\linewidth}\centering
                \includegraphics[width=\linewidth]{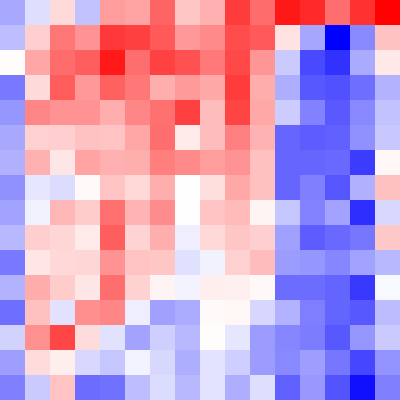}
                \par{\footnotesize Feature diff.\par}
            \end{minipage}\hspace{0.8mm}%
            \begin{minipage}[b]{0.34\linewidth}\centering
                \includegraphics[width=\linewidth]{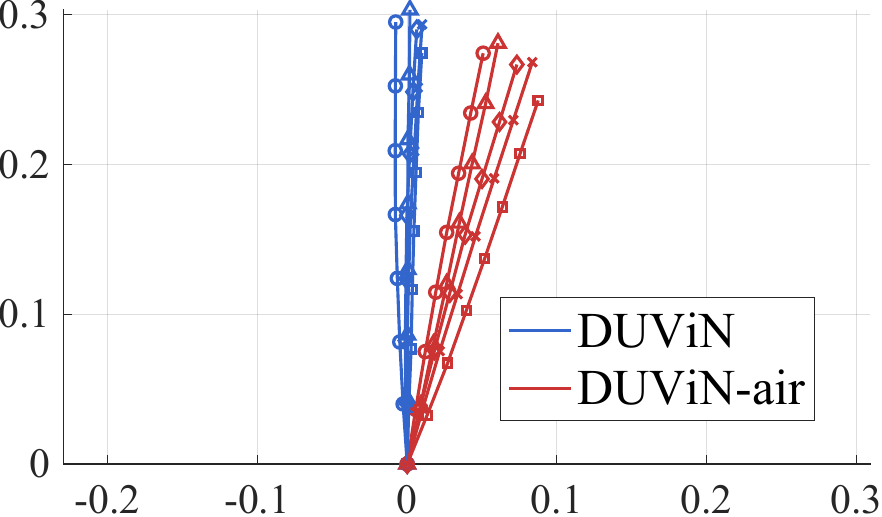}
                \par{\footnotesize Actions\par}
            \end{minipage}
        \end{minipage}%
    }
    \par\vspace{-1mm}
    \subfloat[Feature-level analysis of real-world Scene B]{%
        \begin{minipage}[t]{0.48\textwidth}
            \centering
            \begin{minipage}[b]{0.32\linewidth}\centering
                \includegraphics[width=\linewidth]{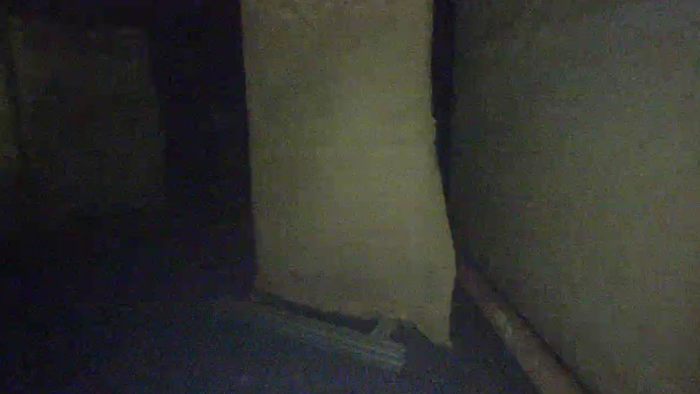}
                \par{\footnotesize Input\par}
            \end{minipage}\hspace{0.8mm}%
            \begin{minipage}[b]{0.32\linewidth}\centering
                \includegraphics[width=\linewidth]{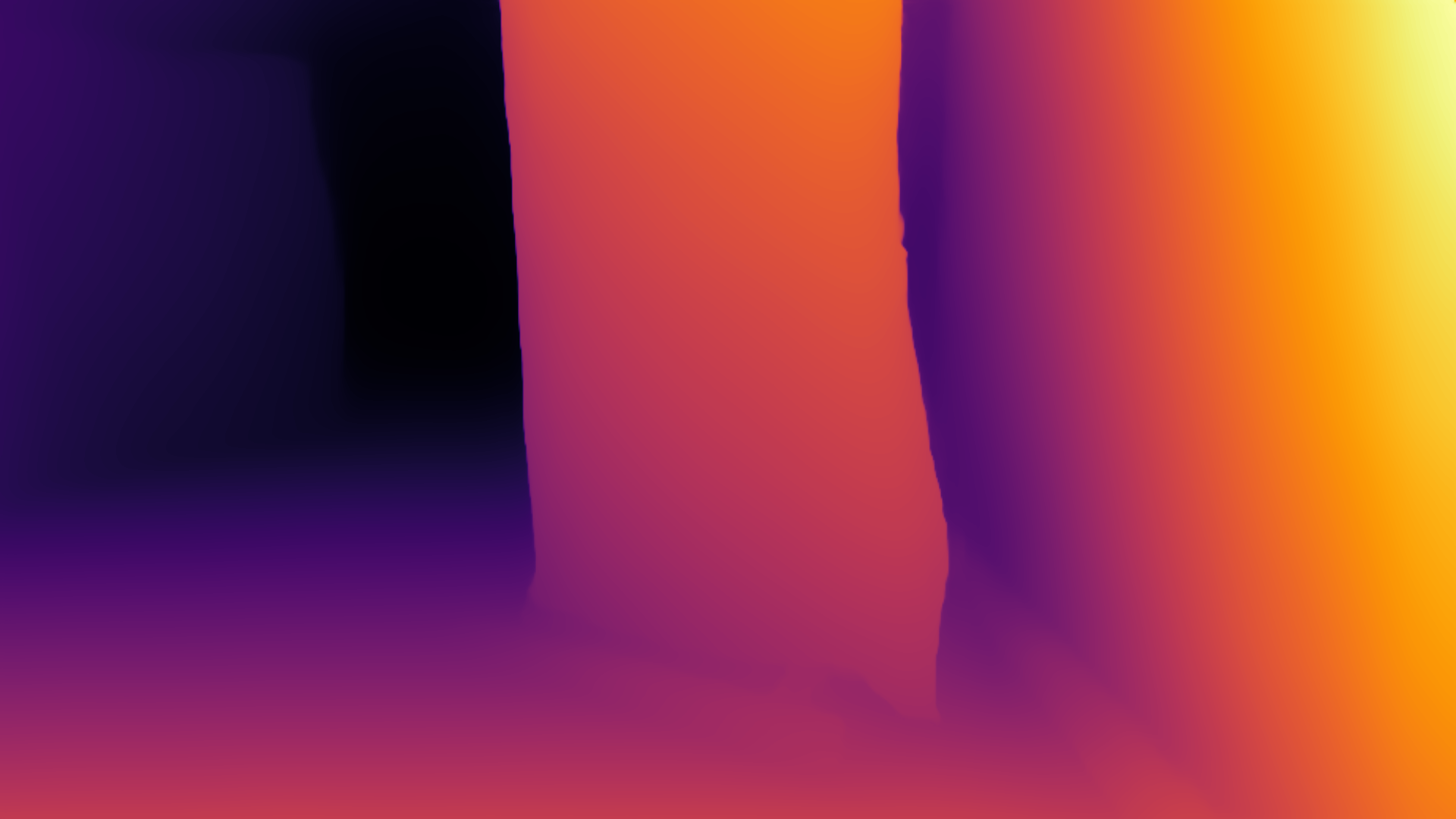}
                \par{\footnotesize DUViN-air\par}
            \end{minipage}\hspace{0.8mm}%
            \begin{minipage}[b]{0.32\linewidth}\centering
                \includegraphics[width=\linewidth]{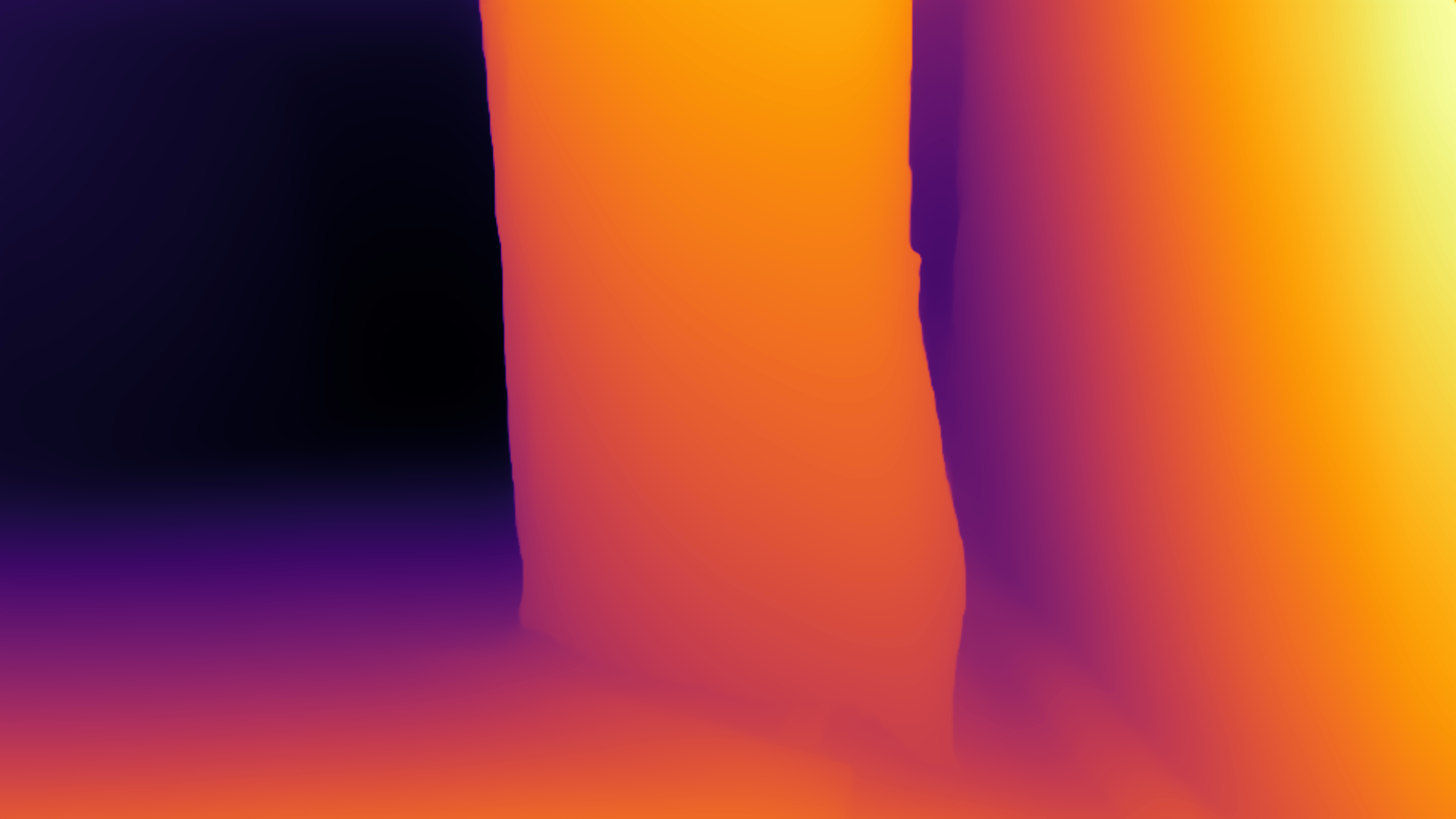}
                \par{\footnotesize DUViN\par}
            \end{minipage}
            \par\vspace{1mm}
            \begin{minipage}[b]{0.19\linewidth}\centering
                \includegraphics[width=\linewidth]{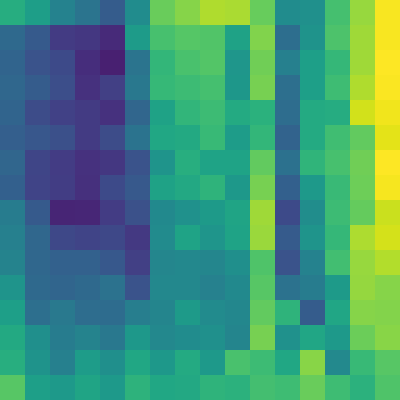}
                \par{\footnotesize DUViN(a) $\mathbb{F}_t^c$\par}
            \end{minipage}\hspace{0.8mm}%
            \begin{minipage}[b]{0.19\linewidth}\centering
                \includegraphics[width=\linewidth]{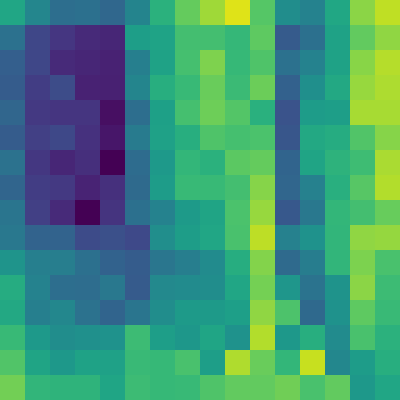}
                \par{\footnotesize DUViN $\mathbb{F}_t^{c}$\par}
            \end{minipage}\hspace{0.8mm}%
            \begin{minipage}[b]{0.19\linewidth}\centering
                \includegraphics[width=\linewidth]{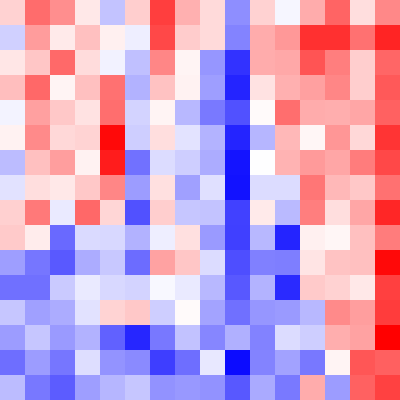}
                \par{\footnotesize Feature diff.\par}
            \end{minipage}\hspace{0.8mm}%
            \begin{minipage}[b]{0.34\linewidth}\centering
                \includegraphics[width=\linewidth]{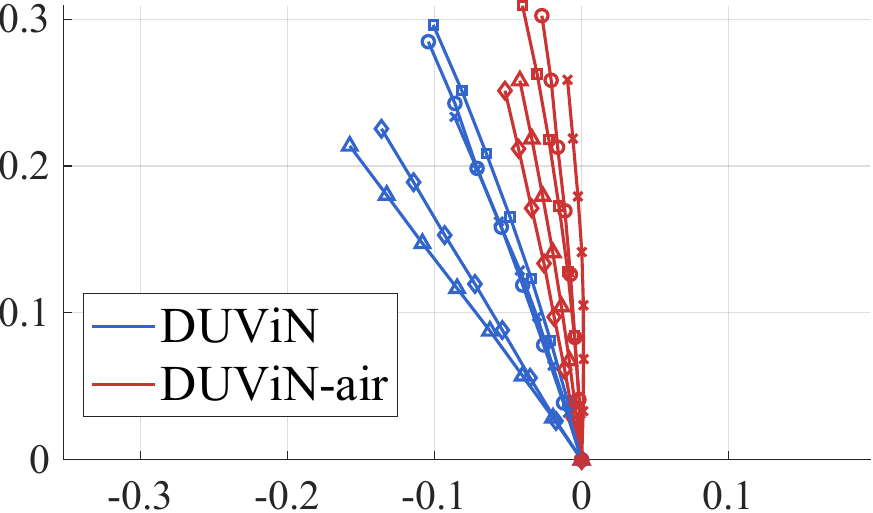}
                \par{\footnotesize Actions\par}
            \end{minipage}
        \end{minipage}%
    }
    \caption{\textbf{Feature-level analysis of DUViN-air failures.} Failure scenes of DUViN-air are extracted for analysis. The inaccurate depth estimation produced by the in-air depth extractor leads to incorrect feature representations $\mathbb{F}_t^c$ for navigation. In contrast, DUViN correctly estimates the obstacle positions as closer (highlighted in blue in the feature difference map), enabling more accurate avoidance behaviour.}
    \label{fig:failure}
\end{figure}

\begin{table}[b]
    \centering
    \caption{Obstacle Avoidance Performance Comparison Between DUViN-air and DUViN in the Real-World Experiment}
    \renewcommand{\arraystretch}{1.5}
    \setlength{\tabcolsep}{12pt} 
    \begin{tabular}{l|c|c}
    \toprule
        \textbf{Method} & \textbf{DUViN-air} & \textbf{DUViN} \\
        \hline
        Succ.\% $\uparrow$ &  40 & \textbf{93} \\
        \hline
        A. C. $\downarrow$ &  0.67 & \textbf{0.36} \\
        \bottomrule
    \end{tabular}
    \label{tab:duvin_navigation_results}
\end{table}

\subsubsection{Obstacle Avoidance Experiments}
To evaluate the obstacle avoidance capability of DUViN in real-world settings, we conduct experiments in which the BlueROV2 navigates through cluttered environments using only vision-based inputs. The ROV is tasked with moving in a forward direction while avoiding randomly placed obstacles, relying entirely on DUViN for motion planning and local collision avoidance. To evaluate the performance of our method under visually challenging conditions, we conducted experiments in a dark underwater environment. In this setup, the BlueROV2 relied solely on its onboard light, and three obstacles were deliberately placed in its path to simulate a cluttered scene. We observe that NoMaD~\cite{sridhar2024nomad} also exhibits roundabout behavior under this visual condition, similar to the results shown in \cref{fig:simulate_nomad}, and is unable to proceed forward. Therefore, we compared the navigation performance between two models: \textbf{DUViN-air}, which uses an in-air trained encoder, and \textbf{DUViN}, which incorporates our physics-informed encoder transferred from the PUDE model.

\begin{figure}[!bh]
    \centering
    \subfloat[\textbf{Navigation in a cluttered environment (Case 1).} Three obstacles are placed in the scene. The AUV, guided by DUViN, successfully navigates around them. Grey shapes represent obstacles, and red outlines denote collision boundaries.\label{fig:trajectory_case1}]{%
        \includegraphics[width=0.99\linewidth, trim=0 60 20 110, clip]{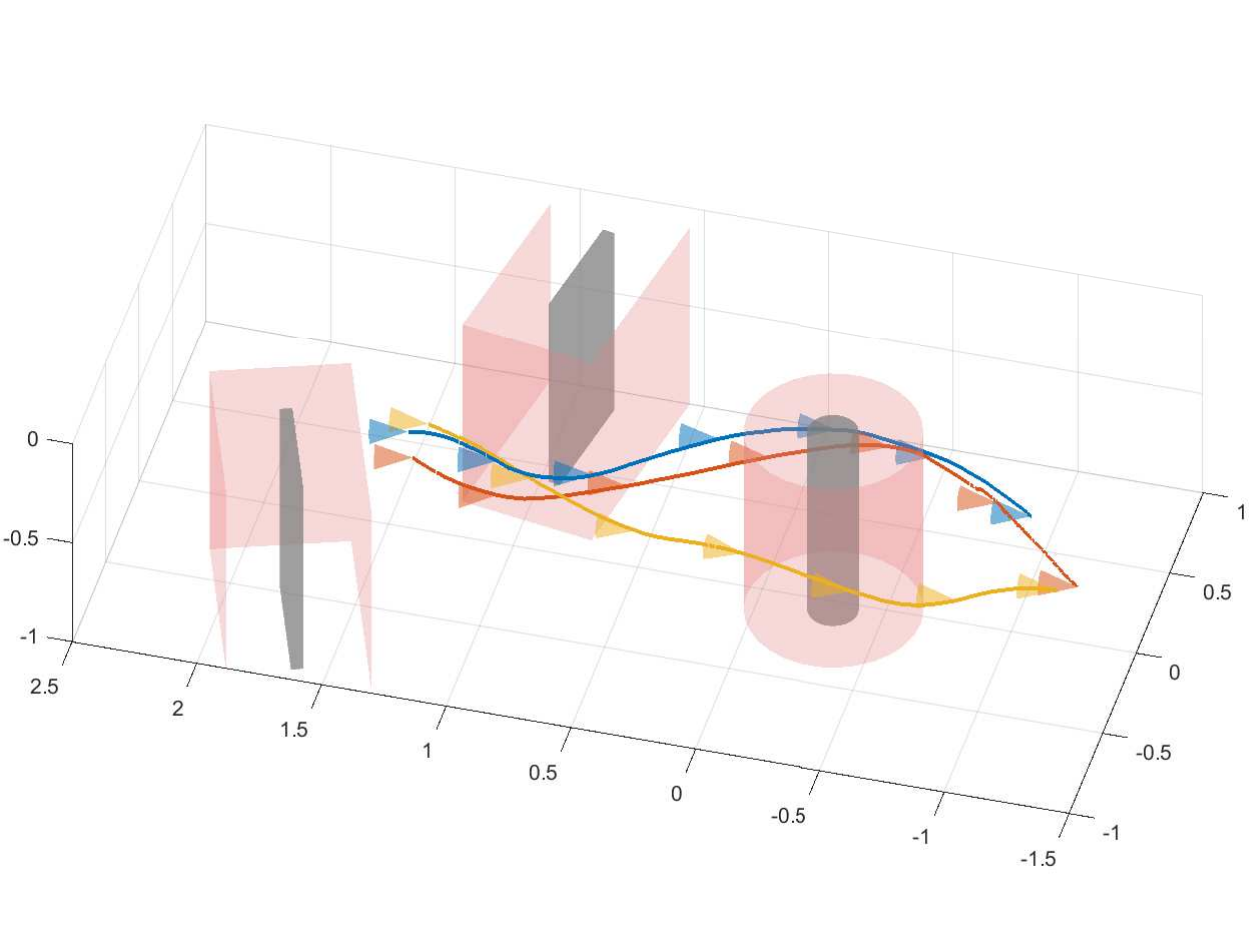}%
    }
    \par\vspace{2mm}
    \subfloat[\textbf{Navigation in a cluttered environment (Case 2).} The AUV exhibits similarly smooth avoidance behaviour under a different layout, including a tube obstacle placed on the ground.\label{fig:trajectory_case2}]{%
        \includegraphics[width=0.99\linewidth]{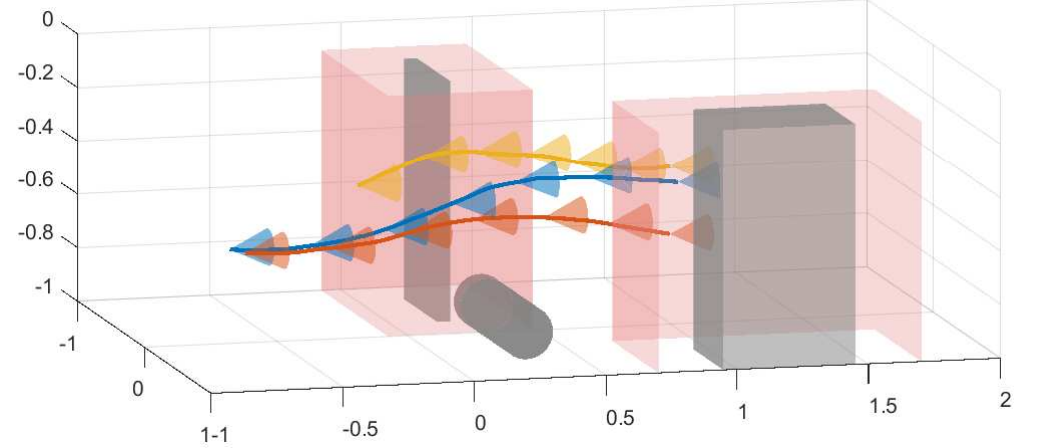}%
    }
    \par\vspace{2mm}
    \subfloat[\textbf{Navigation through a narrow sloped passage (Case 3).} The AUV successfully navigates a narrow central corridor while adjusting its altitude to avoid the inclined slope.\label{fig:trajectory_case3}]{%
        \includegraphics[width=0.99\linewidth, trim=0 15 50 20, clip]{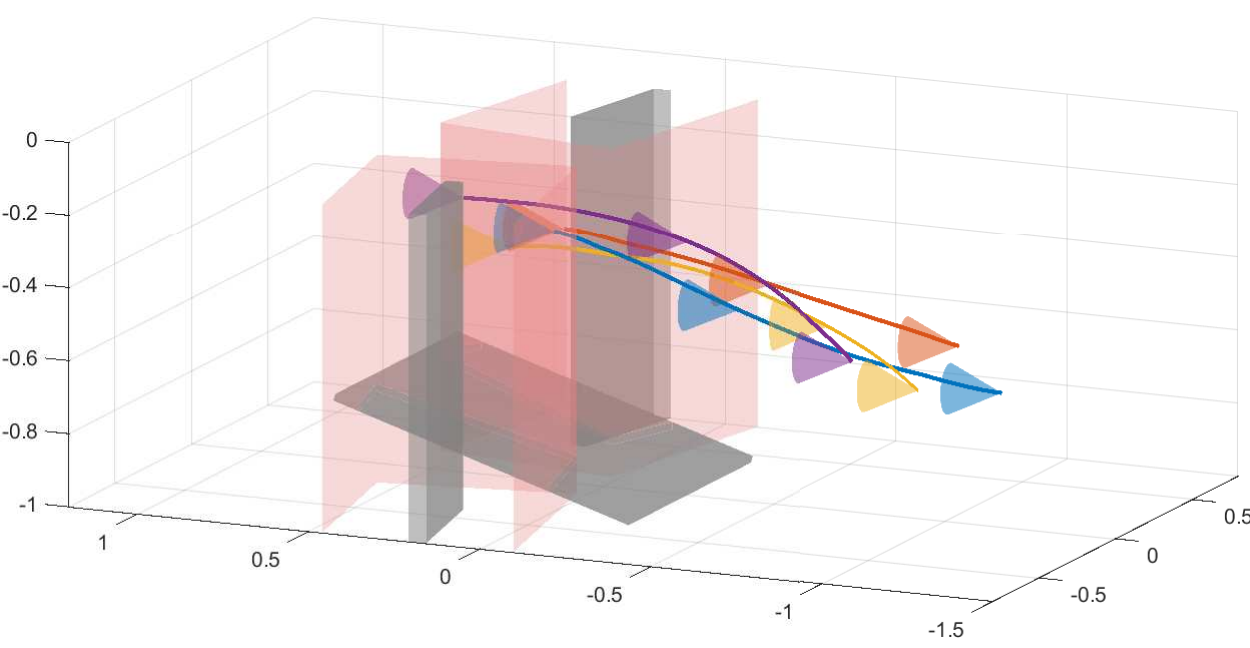}%
    }
    \caption{\textbf{Real-world navigation performance in cluttered environments using DUViN.} DUViN enables the AUV to navigate through densely obstructed scenes under degraded underwater visual conditions, demonstrating robust obstacle avoidance. Different colors represent the trajectories recorded in multiple trials.}
    \label{fig:trajectory_visualization}
\end{figure}

Each model was evaluated over 15 runs. We report two primary metrics: (i) \textit{Success Rate} (Succ.\%): the percentage of runs where the AUV successfully navigated through all obstacles without getting stuck, and (ii) \textit{Average Collision} (A.C.): the mean number of minor collisions or surface contacts recorded during successful trials. The quantitative results are summarized in \cref{tab:duvin_navigation_results}. The findings show that \textbf{DUViN} can successfully navigate through the clustered environment under degraded underwater visual conditions, achieving a success rate of 93\%. This provides proof-of-concept evidence of the effectiveness and sim-to-real transferability of the proposed DUViN method. Meanwhile, \textbf{DUViN} outperformed \textbf{DUViN-air}, achieving a higher success rate and a lower average number of collisions. This improvement demonstrates the effectiveness of the physics-informed encoder in enhancing obstacle avoidance under the degraded underwater environment. We attribute this to improved depth estimation quality, which benefits from underwater depth features learned via the PUDE method. Two failure cases of DUViN-air, along with their feature-level analyses, are presented in \cref{fig:failure}. The results indicate that the failures are primarily due to inaccurate depth estimations produced by the in-air depth feature extractor when applied to underwater environments. Specifically, obstacles are estimated to be further away than their actual positions, leading to delayed avoidance behaviour.

To further illustrate the experimental setup and navigation behaviour, we utilized the underwater tracking system to record trajectories during several successful runs. Recorded trajectories are visualized in \cref{fig:trajectory_visualization}. Interestingly, because the first cylindrical obstacle was placed in the centre of the path, the DUViN policy exhibited multi-modal trajectory generation behaviour, choosing different paths to circumvent the obstacle across trials. Meanwhile, we also validate our model's performance across different environment layouts. For instance, in the layout shown in \cref{fig:title} (case 2), which includes a tube obstacle placed on the ground, the robot adjusts its trajectory along the $Z$-axis to achieve successful avoidance, as illustrated in \cref{fig:trajectory_case2}. Another experiment is conducted in an environment featuring a single narrow path with an inclined slope in the middle (case 3). In this scenario, the AUV is required not only to navigate through the constrained passage but also to adjust its altitude to avoid the slope. The recorded trajectories are illustrated in \cref{fig:trajectory_case3}. The corresponding demonstration videos are available at \url{https://www.youtube.com/watch?v=h_j4T2-p4lM&list=PLqt2s-RyCf1gfXJgFzKjmwIqYhrP4I-7Y&index=2}.

\subsubsection{Altitude Maintenance Experiments}
To validate the altitude maintenance performance of DUViN using reference altitude images, we conduct additional experiments in which the BlueROV2 performs long-term hovering tasks under external disturbances. During these trials, only the altitude control module of DUViN is activated, while the remaining degrees of freedom are manually controlled to move in a small range. This setup isolates the effect of altitude regulation, allowing a focused assessment of DUViN’s robustness in maintaining altitude.

\begin{figure}[t]
    \centering
    \subfloat[Long-term hovering task.\label{fig:duvin_woa}]{%
        \includegraphics[width=0.99\linewidth]{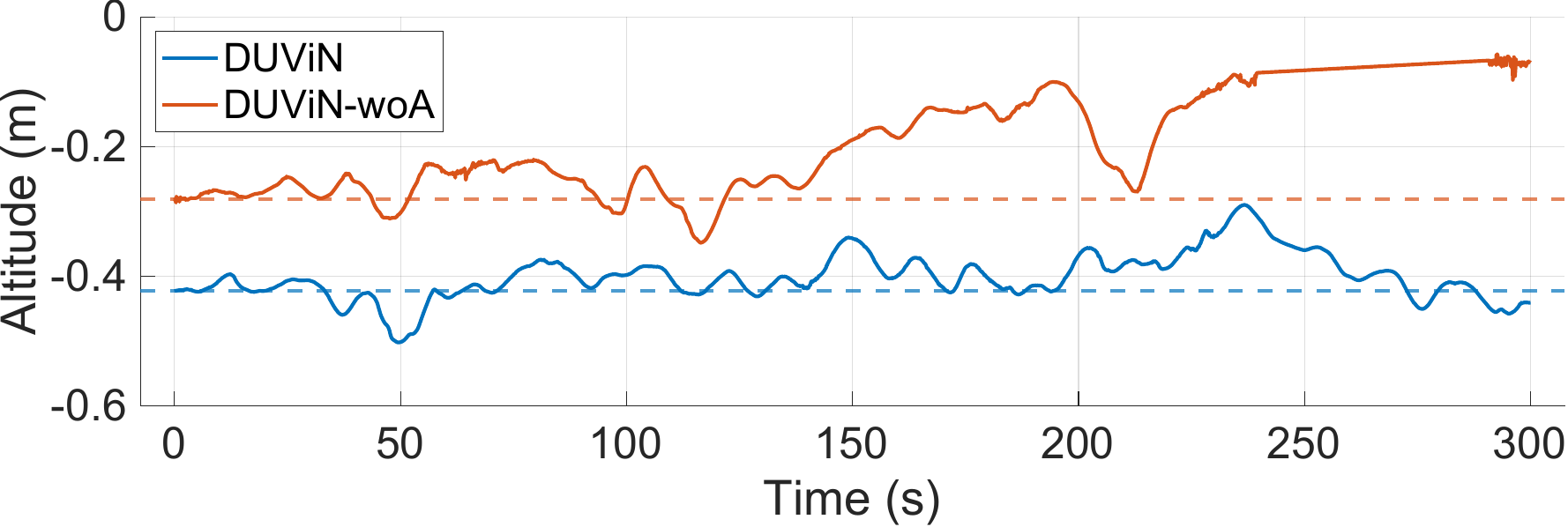}%
    }
    \par
    \subfloat[Disturbance rejection of DUViN.\label{fig:duvin}]{%
        \includegraphics[width=0.99\linewidth]{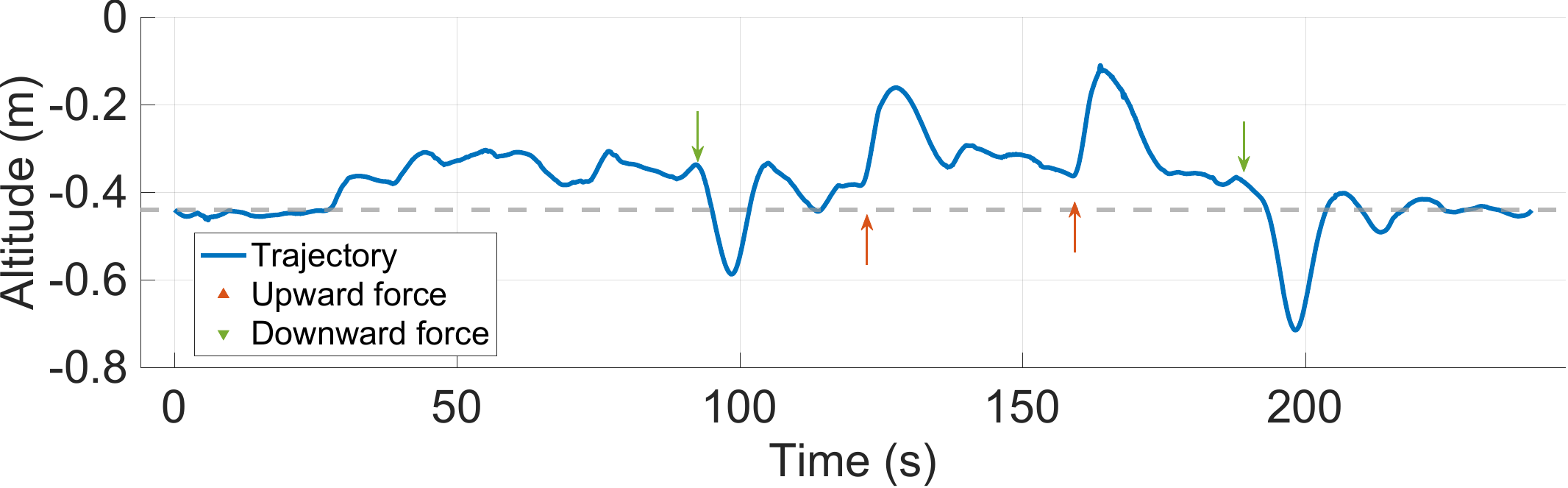}%
    }
    \par
    \subfloat[Disturbance rejection of DUViN-woA.\label{fig:disturbance}]{%
        \includegraphics[width=0.99\linewidth]{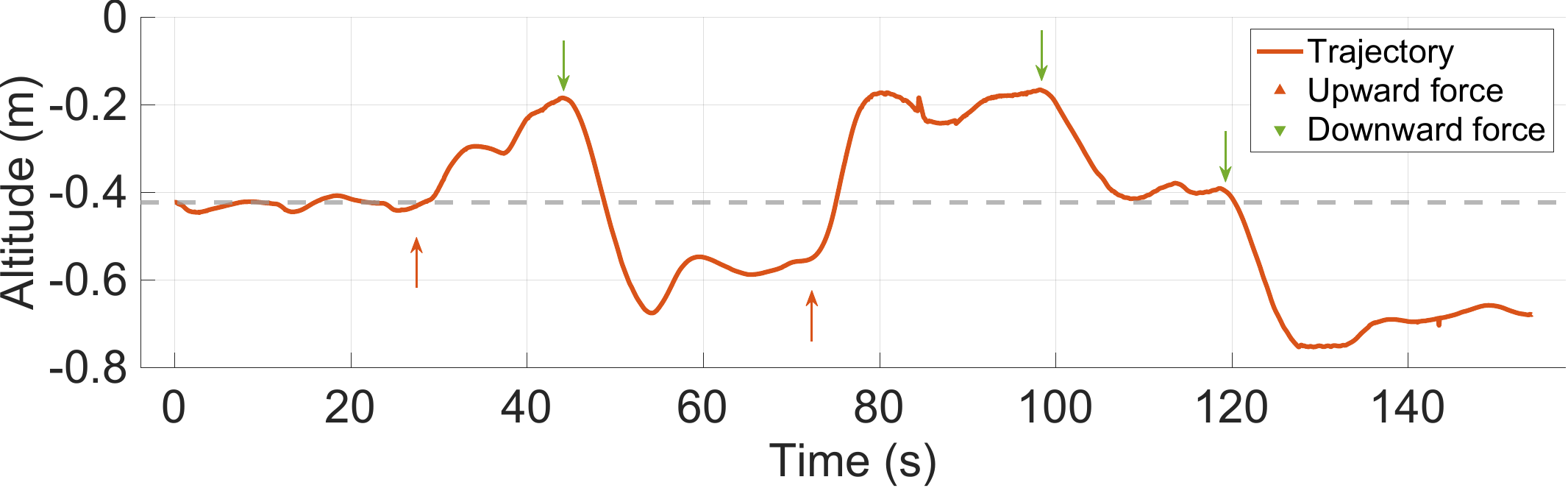}%
    }
    \caption{\textbf{Comparison of altitude control between DUViN and DUViN-woA.} The DUViN model, trained with altitude-maintenance Z-axis planning, is able to maintain a stable altitude relative to the reference and effectively reject external disturbances.}
    \label{fig:altitude_comparison}
\end{figure}

We begin by evaluating the long-term pure hovering task, where no external disturbances are introduced. The robot is subjected only to small-range manual control, inducing mild motion and visual fluctuations along with coupled dynamic disturbances. Each test lasts 5 minutes. We compare two models: \textbf{DUViN}, which uses the transferred underwater encoder; and \textbf{DUViN-woA}, the model trained without the altitude maintenance z-axis planning in \cref{MPC2}. The results are shown in \cref{fig:duvin_woa}, demonstrating that the AUV controlled by DUViN-woA fails to maintain a consistent altitude and gradually ascends due to buoyancy. In contrast, the AUV guided by DUViN, which incorporates the Z-axis replanning process during training, successfully maintains its altitude relative to the original position.

Furthermore, we evaluate the model's ability to reject disturbances. As shown in \cref{fig:duvin}, external disturbances are manually applied to the AUV during the experiment, including both upward and downward forces, as indicated by the colored arrows in the plot. The results show that DUViN successfully guides the AUV back to the reference altitude after each disturbance. In comparison, DUViN-woA exhibits a step-like response to the disturbances, maintaining the perturbed altitude rather than recovering to the original level, as shown in \cref{fig:altitude_comparison}. The corresponding experimental hovering videos are available at \url{https://www.youtube.com/watch?v=dQXiJgGYm8M&list=PLqt2s-RyCf1gfXJgFzKjmwIqYhrP4I-7Y&index=3}.

\subsubsection{Effectiveness of Goal Awareness}
To validate the phenomenon and behavior of goal awareness in real-world scenarios, we conduct an obstacle avoidance task with varying directional guidance. Since direct sensing of the goal position is not available in our experimental setup, we define a virtual goal according to the initial goal direction. During navigation, gyroscope data is integrated to track the accumulated heading changes, enabling continuous estimation of the current heading relative to the goal direction. The task is therefore characterized by the robot navigating under the influence of the goal direction while simultaneously performing obstacle avoidance.

\begin{figure}[t]
    \centering
    \captionsetup[subfigure]{farskip=0pt}
    \subfloat[Goal direction: $0^\circ$]{%
        \includegraphics[width=0.24\textwidth]{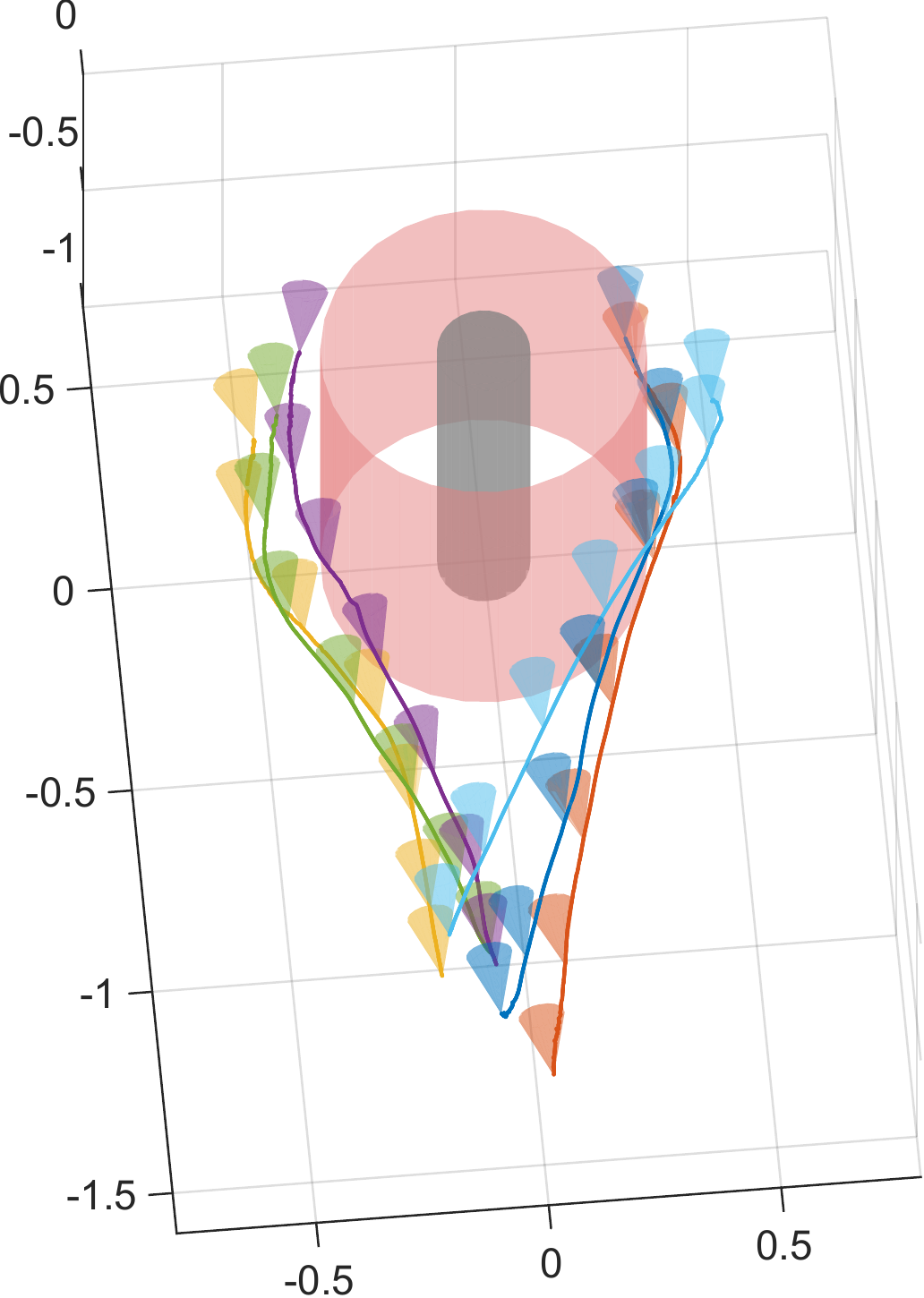}%
    }
    \subfloat[Goal direction: $\pm 30^\circ$]{%
        \includegraphics[width=0.24\textwidth]{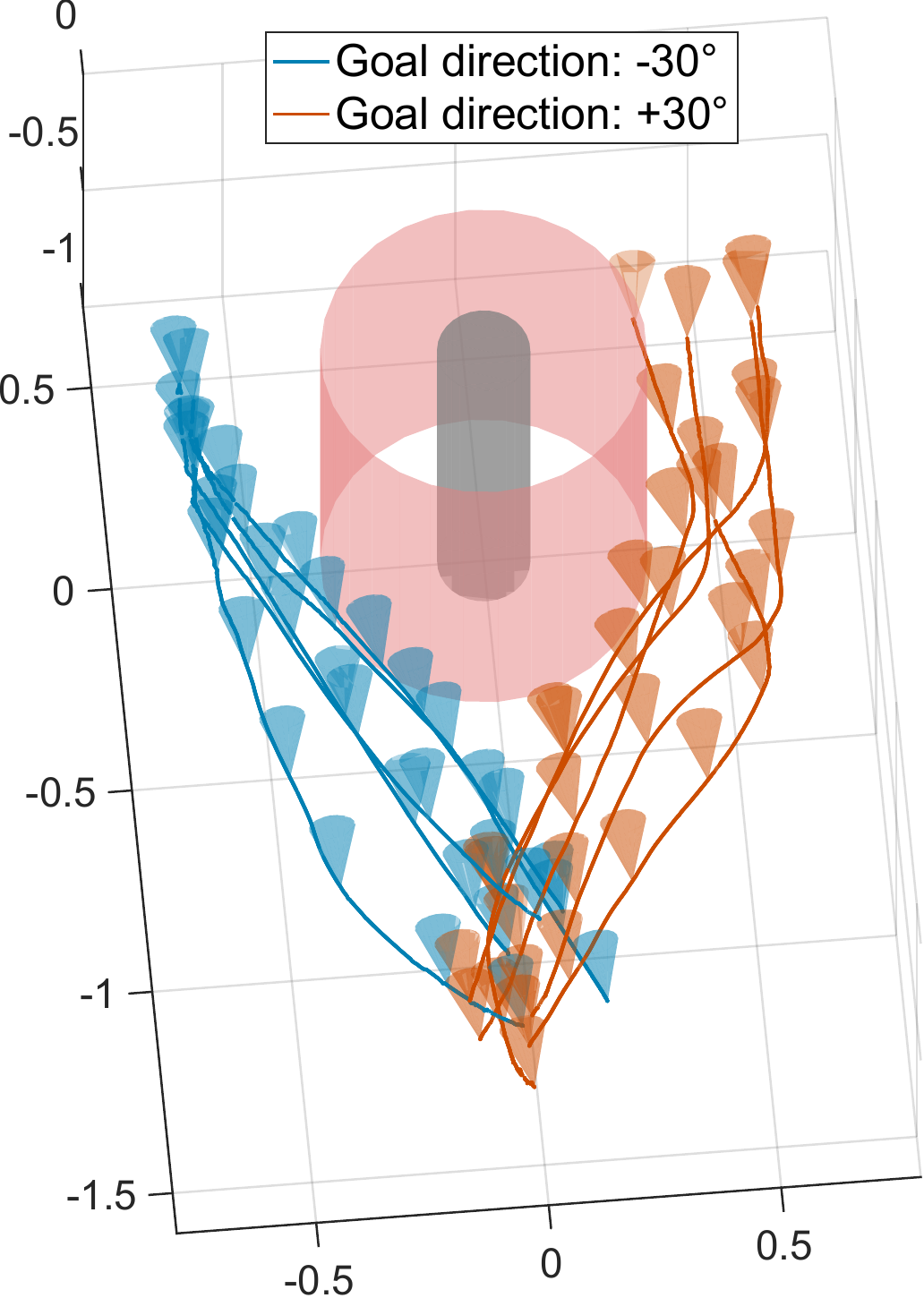}%
    }
    \caption{\textbf{Effectiveness of goal awareness.} A cylindrical obstacle (grey) with a radius of 0.1\,m is placed at the center. The red cylinder represents the collision boundary, considering the 0.25\,m radius of the ROV. The cones indicate the heading directions of the AUV, as measured by the motion tracking system.}
    \label{fig:goal_awareness_visualization}
\end{figure}

With a cylindrical obstacle placed at the center of the environment, we evaluate the goal awareness behavior by assigning initial goal directions of $-30^\circ$, $0^\circ$, and $30^\circ$. We set the distance ratio to a constant value of $0.7$. For each goal direction, multiple trials are conducted and recorded using the tracking system. The corresponding visualization results are shown in \cref{fig:goal_awareness_visualization}.

We observe that goal awareness significantly influences the navigation policy of the AUV. When the goal is oriented forward, the AUV exhibits a detouring behavior characterized by a multi-modal trajectory. In this case, it adjusts its yaw to face the back side of the obstacle and navigates closely along its boundary to reach the goal. In contrast, when the goal direction is set to the left ($-30^\circ$) or right ($+30^\circ$), the policy adopts a more direct strategy: the AUV aligns its movement with the goal direction and makes only minor heading adjustments accordingly. These behaviors observed in real-world experiments are consistent with simulation results, indicating that the policy effectively adapts its strategy based on goal information.

%% file: mainBody/5_Conclusion.tex
\section{Conclusion}
In this work, we propose a diffusion-based navigation policy incorporating a transferred depth feature extractor, tailored for visually degraded underwater environments. The policy enables obstacle avoidance while maintaining altitude relative to the initial condition. Experiments conducted both in simulation and in an indoor water tank demonstrate the effectiveness of our approach, showing its transferability across different layouts and underwater visual conditions.

However, the current simulation experiments are conducted on a single simulation platform, while the real-world validation should be regarded as a proof-of-concept study in an indoor water tank environment. Although the tank experiments were performed under relatively turbid water conditions, different lighting settings, and various obstacle layouts, they do not fully capture the complexity of open-water environments, such as long-range navigation and realistic seabed geometry. The current setup may reach its limit when the deployment environment contains geometric structures or conditions that are significantly beyond the training distribution, such as more complex terrain transitions.

Additionally, the current robustness evaluation is limited to a simulated flow disturbance and does not capture the more complex hydrodynamic disturbances encountered in open-water environments. Practical deployment will also require tighter integration with onboard low-level state estimation for accurate command execution. In particular, the current study does not evaluate errors in the relative goal direction and distance provided by acoustic positioning systems such as USBL, and this remains an important direction for future work.

Future improvements may include expanding policy training to more diverse simulation and real-world scenarios to improve generalization in more complex environments. Additionally, further improving depth estimation consistency, for example, through video-based depth models with stronger temporal consistency, is a promising direction for enhancing the stability of underwater navigation. Another important direction for future work is to validate the proposed framework in open-ocean experiments, where more challenging conditions, such as large-scale scenes and stronger disturbances, can provide a more comprehensive assessment of real-world applicability.

\section*{Acknowledgements}
This work was supported by the Melbourne Research Scholarship from the University of Melbourne, the Australian Research Council via grants DE220101527 and DP250100112, and the Australian Government via grant AUSMURIB000001.